%% file: main.tex
\documentclass[twoside]{article}
\input{2025-AISTATS/header}

\input{2025-AISTATS/definitions}

\usepackage[preprint]{2025-AISTATS/style}

\usepackage{minitoc}

\begin{document}

\twocolumn[

\aistatstitle{Federated UCBVI: Communication-Efficient Federated Regret Minimization with Heterogeneous Agents}

\aistatsauthor{Safwan Labbi 
\And Daniil Tiapkin
\And  Lorenzo Mancini }

\aistatsaddress{
CMAP, École Polytechnique, \\
Palaiseau, France
\And  
CMAP, CNRS, École Polytechnique, \\ Palaiseau, France; \\
Universit{\'e} Paris-Saclay, CNRS, LMO,\\
Orsay, France
\And 
CMAP, École Polytechnique, \\
Palaiseau, France 
}
\aistatsauthor{
Paul Mangold
\And Eric Moulines }

\aistatsaddress{
CMAP, École Polytechnique, \\
Palaiseau, France 
\And
CMAP, École Polytechnique,
Palaiseau, France\\
MBZUAI
}]

\doparttoc %
\faketableofcontents %

\begin{abstract}
In this paper, we present the Federated Upper Confidence Bound Value Iteration algorithm (\algo), a novel extension of the \texttt{UCBVI} algorithm \citep{azar2017minimax} tailored for the federated learning framework. We prove that the regret of \algo\ scales as $\tcO(\sqrt{\sepisode^3\nstates \nactions \nepisode / \nagent})$, with a small additional term due to heterogeneity, where $\nstates$ is the number of states, $\nactions$ is the number of actions, $\sepisode$ is the episode length,  $\nagent$ is the number of agents, and $\nepisode$ is the number of episodes. Notably, in the single-agent setting, this upper bound matches the minimax lower bound up to polylogarithmic factors, while in the multi-agent scenario, \algo\ has linear speed-up. To conduct our analysis, we introduce a new measure of heterogeneity, which may hold independent theoretical interest. Furthermore, we show that, unlike existing federated reinforcement learning approaches, \algo's communication complexity only marginally increases with the number of agents.
\end{abstract}

\section{INTRODUCTION}

\input{2025-AISTATS/main_text/Introduction}

\section{RELATED WORK}
\label{sec:relatedworks}

\input{2025-AISTATS/main_text/relatedworks}

\section{SETTING}
\label{sec:prelimaries}

\input{2025-AISTATS/main_text/preliminaries}

\section{FED-UCBVI ALGORITHM}
\label{sec:feducbvi}

\input{2025-AISTATS/main_text/FedUCBVI}

\section{EXPERIMENTS}
\label{sec:numerical-study}

\input{2025-AISTATS/main_text/Experiments}

\section{CONCLUSION}

\input{2025-AISTATS/main_text/Conclusion}

\section*{ACKNOWLEDGEMENTS}
The work of S. Labbi, L.Mancini and P. Mangold has been supported by Technology Innovation Institute (TII), project Fed2Learn. The work of D.Tiapkin has been supported by the Paris Île-de-France Région in the framework of DIM AI4IDF.
The work of E. Moulines has been partly funded by the European Union (ERC-2022-SYG-OCEAN-101071601). Views and opinions expressed are however those of the author(s) only and do not necessarily reflect those of the European Union or the European Research Council Executive Agency. Neither the European Union nor the granting authority can be held responsible for them.

\bibliography{references}
\clearpage
\newpage

\newpage
\appendix
\onecolumn
\aistatstitle{Supplementary Materials}
\setcounter{section}{0}

\input{2025-AISTATS/appendix}
\input{2025-AISTATS/appendix/notation}

\input{2025-AISTATS/appendix/algo}
\input{2025-AISTATS/appendix/concentration}

\input{2025-AISTATS/appendix/heterogeneity}
\input{2025-AISTATS/appendix/regret_new}
\input{2025-AISTATS/appendix/communication}
\input{2025-AISTATS/appendix/technical}

\end{document}

%% file: 2025-AISTATS/header.tex
\usepackage{color}
\usepackage{graphicx}
\usepackage{subcaption}
\usepackage{booktabs}
\usepackage{tabularx}
\usepackage[usenames,dvipsnames,svgnames]{xcolor}

\usepackage[colorlinks=true, allcolors=blue]{hyperref}
\usepackage{mdframed}
\usepackage{colortbl} 

\usepackage{amsfonts,amsmath,amssymb}
\usepackage{amsthm}
\usepackage{srcltx}
\usepackage{etoolbox}
\usepackage{nameref}
\usepackage{comment}
\usepackage{mathrsfs}
\usepackage{enumerate}
\usepackage[shortlabels]{enumitem}
\usepackage{amsfonts} %
\usepackage{nicefrac} %
\usepackage{mathtools}
\usepackage{dsfont,mathrsfs}
\usepackage[capitalize,noabbrev]{cleveref}
\usepackage{xargs}
\usepackage{multirow}
\usepackage[textsize=tiny]{todonotes}
\usepackage{ifthen}

\usepackage[ruled,vlined]{algorithm2e}
\usepackage{etoolbox}

\mdfdefinestyle{myboxstyle}{
  linecolor=lightgray!20,
  linewidth=1pt,
  roundcorner=5pt,
  backgroundcolor=lightgray!50,
  innertopmargin=5pt,
  innerbottommargin=5pt,
  innerleftmargin=5pt,
  innerrightmargin=5pt,
  skipabove=0pt,
  skipbelow=0pt,
}

\usepackage{bbm}
\usepackage{thmtools,thm-restate}

\usepackage{upgreek}

\usepackage{xfrac}

\usepackage{parskip}
\usepackage{amssymb}%
\usepackage{pifont}%
\usepackage[letterpaper,top=2cm,bottom=2cm,left=2cm,right=2cm,marginparwidth=1.25cm]{geometry}
\usepackage[round]{natbib}

\usepackage{float}
\usepackage{placeins}

\newtheorem{theorem}{Theorem}[section]

\newtheorem{lemma}{Lemma}[section]
\newtheorem{corollary}{Corollary}[section]

\usepackage[flushleft]{threeparttable}
\usepackage{array}

  {\par\noindent\textit{Sketch of the proof.}\hspace{0.3em}} %
  {\hfill$\square$\par} %

%% file: 2025-AISTATS/definitions.tex
\newcommand{\algname}[2]{\hyperref[#2]{\textbf{#1}}\label{#2}}

\definecolor{ao}{rgb}{0.0, 0.5, 0.0}

\def\rset{\mathbb{R}}

\def\nset{\ensuremath{\mathbb{N}}}
\def\nsets{\ensuremath{\mathbb{N}^*}}

\newcommand{\cA}{\mathcal{A}}

\newcommand{\cE}{\mathcal{E}}

\newcommand{\cF}{\mathcal{F}}
\newcommand{\cG}{\mathcal{G}}

\newcommand{\cM}{\mathcal{M}}

\newcommand{\cO}{\mathcal{O}}
\newcommand{\tcO}{\widetilde{\cO}}

\newcommand{\cS}{\mathcal{S}}

\renewcommand{\phi}{\varphi}

\DeclareMathOperator*{\argmax}{arg\,max}
\newcommand{\Ind}{\mathsf{1}}

\newcommand{\Bern}{\mathrm{Ber}}

\newtheorem{assumLHG}{\textbf{A}-\hspace{-3pt}}
\Crefname{assumLHG}{\textbf{A}-\hspace{-3pt}}{\textbf{A}-\hspace{-3pt}}
\crefname{assumLHG}{\textbf{A}}{\textbf{A}\hspace{-2pt}}

\newcommandx\sequence[4][2=,3=,4=]
{\ifthenelse{\equal{#3}{}}{\ensuremath{\{ #1_{#2 #4}\}}}{\ensuremath{\{ #1_{#2 #4} \}_{#2 \in #3}}}}
\newcommandx\sequenceD[2][2=]
{\ifthenelse{\equal{#2}{}}{\ensuremath{\{ #1\}}}{\ensuremath{\{ #1\!~:\!~#2  \}}}}
\newcommandx\sequenceDouble[4][3=,4=]
{\ifthenelse{\equal{#3}{}}{\ensuremath{\{ (#1_{#3},#2_{#3})\}}}{\ensuremath{\{ (#1_{#3},#2_{#3})\}_{#3 \in #4}}}}
\newcommandx{\sequencen}[2][2=n\in\nset]{\ensuremath{\{ #1, \eqsp #2 \}}}
\newcommandx\sequencens[2][2=n]
{\ensuremath{\{ #1_{#2} \!~:\!~#2\in\nsets\}}}
\newcommandx\sequencet[4]
{\ensuremath{\{ #1{#2_{#3}} \, : \, \eqsp #3 \in #4 \}}}

\def\PE{\mathbb{E}}
\def\P{\mathbb{P}}
\newcommandx{\CPP}[3][1=]
{\ifthenelse{\equal{#1}{}}{{\mathbb P}\left(\left. #2 \, \right| #3 \right)}{{\mathbb P}_{#1}\left(\left. #2 \, \right | #3 \right)}}
\def\Var{\operatorname{Var}}
\newcommand{\empvar}[2][1=,2=]{\Var_{\widehat{\kergen}_{#1}}(#2)}
\newcommand{\comempvar}[2][1=,2=]{\Var_{\widehat{\kergen}_{#1}^{\hspace{0.07em} c}}(#2)}

\newcommand{\agentvar}[3][1=,2=,3=]{\Var_{\kergen_{#1}^{#2}}(#3)}

\def\PE{\mathbb{E}}
\newcommandx{\CPE}[3][1=]{{\mathbb E}_{#1}\left[#2 \middle\vert #3 \right]}

\newcommandx{\CPEs}[3][1=,2=,3=]{{\mathbb{E}}_{#2}[#1|#3]} 
\def\PE{\mathbb{E}}

\newcommand{\abs}[1]{\vert #1\vert}
\newcommandx{\norm}[2][2=]{\Vert#1 \Vert_{{#2}}}
\newcommandx{\normop}[2][2=]{\Vert{#1}\Vert_{{#2}}}

\newcommandx{\as}[1][1=\PP]{\ensuremath{#1\, -\mathrm{a.s.}}}

\newcommand{\ie}{i.e.}

\newcommand{\eg}{e.g.}

\newcommand{\eqsp}{\;}

\newcommandx{\kerMRP}[2][1=,2=]{\kergen_{#1}^{\hspace{0.07em}#2}}

\newcommand{\policy}{\pi}
\newcommandx{\apolicy}[1][1=]{\policy_{#1}}
\newcommandx{\optpolicy}[2][1=,2=]{\policy_{#1}^{#2}}
\newcommandx{\gloptpolicy}[1][1=]{\policy_{#1}^{\text{col}, \star}}
\newcommand{\kergen}{\mathsf{P}}
\newcommand{\common}{\mathsf{c}}
\newcommand{\indiv}{\mathsf{ind}}
\newcommandx{\kerMDP}[2][1=,2=]{\kergen_{#1}^{\hspace{0.07em}#2}}
\newcommandx{\akerMDP}[2][1=,2=]{\overline{\kergen}_{#1}^{\hspace{0.07em}#2}}
\newcommandx{\comkerMDP}[2][1=,2=]{
    \kergen_{#1}^{\hspace{0.07em}\common \ifthenelse{\equal{#2}{}}{}{,}#2}
}
\newcommandx{\ecomkerMDP}[2][1=,2=]{
\widehat{\kergen}_{#1}^{\hspace{0.07em}\common \ifthenelse{\equal{#2}{}}{}{,}#2}
}
\newcommandx{\perkerMDP}[2][1=,2=]{\kergen_{#1}^{\hspace{0.07em}\indiv \ifthenelse{\equal{#2}{}}{}{,}#2}
}
\newcommandx{\ekerMDP}[2][1=, 2=]{\widehat{\kergen}_{#1}^{\hspace{0.07em}#2}}
\def\r{\mathsf{r}}
\newcommandx{\rewardMDP}[2][1=,2=]{\r_{#1}^{\hspace{0.07em}#2}}
\newcommandx{\erewardMDP}[1][1=]{\widehat{\r}_{#1}}
\newcommandx{\arewardMDP}[1][1=]{\tilde{\r}_{#1}}
\newcommandx{\comrewardMDP}[1][1=]{\r_{#1}^{\mathsf{c}}}
\newcommandx{\qstar}[1][1=]{\qfuncgen_{#1}^{ \star}}
\newcommandx{\vstar}[1][1=]{\valuefuncgen_{#1 }^{ \star}}
\def\valuefuncgen{\mathcal{V}}
\def\qfuncgen{\mathcal{Q}}
\newcommandx{\valuefunc}[2][1=,2=]{\valuefuncgen_{#1}^{#2}}
\newcommandx{\comvaluefunc}[2][1=,2=]{
    \valuefuncgen_{#1}^{\hspace{0.07em}\common \ifthenelse{\equal{#2}{}}{}{,}#2}
}
\newcommandx{\comstarvaluefunc}[1][1=]{\valuefuncgen_{#1}^{\hspace{0.07em}\common,\star}}
\newcommandx{\avaluefunc}[2][1=,2=]{\overline{\valuefuncgen}_{#1}^{#2}}
\newcommandx{\gloptvaluefunc}[1][1=]{\valuefuncgen_{#1}^{\text{col}, \star}}
\newcommandx{\qfunc}[2][1=,2=]{\qfuncgen_{#1}^{#2}}
\newcommandx{\comqfunc}[2][1=,2=]{
    \qfuncgen_{#1}^{\hspace{0.07em}\common \ifthenelse{\equal{#2}{}}{}{,}#2}
}
\newcommandx{\comstarqfunc}[1][1=]{\qfuncgen_{#1}^{\hspace{0.07em}\common,\star}}
\newcommandx{\aqfunc}[2][1=,2=]{\overline{\qfuncgen}_{#1}^{#2}}
\newcommandx{\occmeasure}[2][1=, 2=]{d_{#1}^{#2}}
\newcommandx{\optoccmeasure}[1][1=]{d_{#1}^{\star}}

\newcommand{\statdist}[2][1=,2=]{d_{#1}^{\hspace{0.07em} #2}}
\def\gengap{\Delta}
\def\genegap{\delta}
\newcommandx{\gap}[2][1=,2=]{\gengap_{#1}^{#2}}
\newcommandx{\ugap}[2][1=,2=]{\widetilde{\gengap}_{#1}^{#2}}
\newcommandx{\egap}[2][1=,2=]{\genegap_{#1}^{#2}}
\newcommandx{\eugap}[2][1=,2=]{\widetilde{\genegap}_{#1}^{#2}}

\def\counter{N}
\newcommand{\parcounter}{\widetilde{N}}
\def\bonus{b}

\newcommandx{\evaluefunc}[2][1=,2=]{\hat\valuefuncgen_{#1}}

\newcommandx{\eqfunc}[2][1=,2=]{\hat\qfuncgen_{#1}^{#2}}

\newcommandx{\sevaluefunc}[2][1=,2=]{\bar\valuefuncgen_{#1}}

\newcommandx{\seqfunc}[2][1=,2=]{\bar\qfuncgen_{#1}^{#2}}

\newcommand{\tildeo}{\tilde{\mathcal{O}}}

\def\A{\mathcal{A}}
\def\S{\mathcal{S}}
\def\nstates{|\mathcal{S}|}
\def\nactions{|\mathcal{A}|}

\def\nagent{M}
\def\sepisode{H}
\def\nepisode{T}

\def\troundsmax{R_{\max}}

\def\hgkernel{\varepsilon_{\mathsf{p}}}
\def\hgreward{\varepsilon_{\r}}

\newcommand{\cmark}{\ding{51}}%
\newcommand{\xmark}{\ding{55}}%
\newcommand{\yes}{\cmark}
\newcommand{\no}{\xmark}

\newcommand{\KL}{\mathrm{KL}}
\newcommand{\rme}{\mathrm{e}}
\newcommand{\regret}{\mathfrak{R}}

\newcommand{\term}[1]{\mathbf{(#1)}}
\newcommand{\algo}{\hyperref[algo:FEDUCBVI]{\texttt{Fed-UCBVI}}}

\newcolumntype{C}{>{\centering\arraybackslash}X}

\newcommand{\aggw}{\omega}

%% file: 2025-AISTATS/main_text/Introduction.tex
\begin{table*}
  \caption{Comparison with related algorithms in the online setting}\label{ta:upper_bound}
  \centering
\resizebox{\textwidth}{!}{
  \begin{threeparttable}
\renewcommand{\arraystretch}{1.2}
    \label{tab:comparaison_table}
    \begin{tabular}{>{\centering\arraybackslash}c>{\centering\arraybackslash}c>{\centering\arraybackslash}c>{\centering\arraybackslash}c>{\centering\arraybackslash}c} 
     \toprule
     \multicolumn{1}{c}{Type}  & \multicolumn{1}{c}{Algorithm}  &
     \multicolumn{1}{c}{Heterogeneity}&
     \multicolumn{1}{c}{Communication complexity}
     & \multicolumn{1}{c}{Regret}
     \\ \midrule 
    \multirow{3}{*}{Model-based} &  \texttt{Concurrent UCBVI}~(\citealt{azar2017minimax}) & \no & $\cO ({T})$ & $\tildeo (\sqrt{\sepisode^3 \nstates \nactions \nepisode / \nagent })$  \\
    & \texttt{Byzan-UCBVI}~(\citealt{chen2023byzantine}) & \no & $\cO ({\nagent} \cdot \nstates \nactions  \sepisode \cdot {\log(\nepisode)})$ & $\tildeo (\sqrt{\sepisode^4 \nstates^2 \nactions \nepisode / \nagent })$  \\
    \rowcolor{lightgray!50}  & \algo~(our work) & \yes & $\cO \left( \nstates  \nactions \sepisode \cdot  {\log(\nepisode)}\right)$ & $\tildeo (\sqrt{\sepisode^3 \nstates \nactions \nepisode / \nagent})$ \\
     \midrule
         \multirow{3}{*}{Model-free } & \texttt{Concurrent UCB-Advantage}~(\citealt{zhang2020almost})& \no & $\cO({\nepisode})$& $\tildeo(\sqrt{\sepisode^3 \nstates \nactions \nepisode / \nagent }) $  \\
         & \texttt{FedQ-Bernstein}~(\citealt{zheng2023federated}) & \no & $\cO ({\nagent} \cdot  \nstates \nactions  \sepisode^3 \cdot  {\log(\nepisode)})$  &  $\tildeo(\sqrt{\sepisode^4 \nstates \nactions \nepisode / \nagent}) $ \\
         & \texttt{FedQ-Advantage}~(\citealt{zheng2024federated}) & \no & $\cO ({\nagent} \cdot  \nstates \nactions \sepisode^2 \log(\sepisode) \cdot {\log(\nepisode)})$  &  $\tildeo(\sqrt{\sepisode^3 \nstates \nactions \nepisode / \nagent}) $
         \\
     \midrule
         \multirow{1}{*}{ } & \texttt{Lower Bound}~(\citealt{jin2018q,domingues2021episodic})& \no & ? & $\tildeo(\sqrt{\sepisode^3 \nstates \nactions \nepisode / \nagent }) $  \\
         \bottomrule
    \end{tabular}
\begin{tablenotes}
   \item[*] The results are derived in a homogeneous setting. For all bounds, only the leading term with respect to the dependence on $T$ is shown. $\sepisode$: number of steps per episode; $\nepisode$: total episodes collected per agent; $\nstates$: number of states; $\nactions$; number of actions;  $\nagent$: number of agents. 
  \end{tablenotes}
  \end{threeparttable}}
  \end{table*}

Federated reinforcement learning (FRL, \citealp{zhuo2019federated,qi2021federated}) adapts the principles of federated learning (FL, \citealp{mcmahan2017communication}) to the domain of reinforcement learning (RL, \citealp{sutton2018}).
It enables multiple agents, evolving in independent environments, to learn a policy collaboratively without directly exchanging their states/actions.
To learn together, agents communicate under the supervision of a central server (CS), aiming to maximize the expected rewards averaged across all agents.
Consequently, agents participating in FRL may learn better policies with fewer interactions with the environment. FRL appears to be a promising solution for reducing the cost of training.
However, the efficient implementation of FRL faces significant challenges.
Similarly to FL, agents typically evolve in different environments and often have limited computational power and communication bandwidth. Furthermore, the traditional challenges of RL, such as balancing exploration and exploitation, remain.
Thus, there is a growing demand for methods tailored for FRL, aiming to reduce communication complexity (\ie, the number of communications) while maintaining efficient exploration and learning. 

FRL has attracted considerable attention in recent years, with a strong focus put on federated versions of Q-Learning. 
This research often relies on one of two following assumptions: either (1) all agents operate in \emph{identical environments} \citep{chen2023byzantine,zheng2023federated, zheng2024federated}, or (2) a generative model is available, allowing access to sampling from any state-action pair without exploration \citep{jin2022federated, wang2024convergence}. Another notable category of methods, called distributed reinforcement learning \citep{bai2019provably,zhang2020almost}, enables agents to address RL problems collaboratively.
However, these methods require centralizing observational data on a single server, which may not be feasible in real applications.

Unfortunately, the aforementioned approaches do not address the exploration-exploitation trade-off in heterogeneous environments. Furthermore, their high communication complexity poses a major challenge for their use, even with homogeneous agents.

In this paper, we introduce the algorithm \algo\ for tabular episodic FRL and we analyze its \emph{federated regret}, i.e., the regret averaged across all agents, in the presence of environmental heterogeneity.  The tabular FRL problem involves $\nagent$ agents, each interacting with its own environment, modeled as a finite-horizon Markov Decision Process (MDP). For an agent $i \in [\nagent]$, a finite-horizon MDP is defined by a tuple $\cM^i := (\S, \A, \sepisode, \{\kerMDP[h][i]\}_{h \in [\sepisode]}, \{\rewardMDP[h][i]\}_{h \in [\sepisode]})$, where $\S$ is the finite state space, $\A$ is the finite action space,  $\sepisode$ is the number of steps in one episode (also referred to as a planning horizon), $\kerMDP[h][i](s'|s,a)$ denotes the probability of transitioning from a state $s \in \S$ to the next state $s' \in \S$ after taking action $a \in \A$ at step $h$ for agent $i$, and $\rewardMDP[h][i](s,a)$, is a bounded deterministic reward function that satisfies $\rewardMDP[h][i](s,a) \in [0,1]$ for all $(s,a,h,i) \in \S \times \A \times [\sepisode] \times [\nagent]$. Note that both the transition probabilities (kernel) and the reward function can vary depending on the decision-making step $h \in [\sepisode]$. The learning process is divided into $\nepisode$ episodes, each of length $\sepisode$. Both the transition kernel and the reward function are assumed to be \textit{unknown} to all agents and the central server (CS).

\algo\ is a model-based approach where each agent independently estimates its local state-action transition kernel. These local estimates are then used to compute state-action value functions, which are aggregated by a CS using an adaptive scheme that accounts for each agent's level of uncertainty. Communication complexity is managed through an adaptive communication strategy triggered by the optimization process's progress and ensures efficient coordination. Overall, our contributions are:
\begin{itemize}[itemsep=-5pt,leftmargin=8pt]
    \item We propose \algo, an FRL algorithm designed to aggregate the local estimators of each agent. We prove that the federated regret of \algo\ scales as $\cO(\sqrt{\sepisode^3\nstates \nactions \nepisode / \nagent})$, up to a heterogeneity term which scales proportionally to our heterogeneity measure.
 This shows that \algo\ achieves a linear speedup and effectively accelerates training compared to single-agent RL.
 To our knowledge, \algo\ is the first provably efficient algorithm for regret minimization in heterogeneous environments.
    \item To analyze \algo, we introduce a new measure of heterogeneity that quantifies the divergence of each agent's state-transition kernel from a baseline kernel, which may be of independent interest.
    \item We develop a novel method for reducing the communication cost. We prove that the communication complexity of \algo\ is $\cO(\nagent\log\log\nepisode + \log \nepisode)$. This is a significant improvement over existing methods (\eg, \citealp{zheng2023federated}), that require $\cO(\nagent \log \nepisode)$ communication rounds.
    \item We validate our theoretical results through numerical experiments on FRL problems, demonstrating that our algorithm outperforms existing FRL baselines with theoretical guarantees. In particular, our simulations show a significant improvement in regret compared to Fed-Q-learning  \citep{zheng2023federated} for different degrees of heterogeneity.
\end{itemize}

The paper is organized as follows: we review the related work in \Cref{sec:relatedworks}, and introduce the necessary mathematical background in \Cref{sec:prelimaries}. In \Cref{sec:feducbvi}, we introduce and analyze the \algo\ algorithm. Then, we present numerical experiments in \Cref{sec:numerical-study}.

%% file: 2025-AISTATS/main_text/relatedworks.tex
\paragraph{Reinforcement Learning.} 
Two main approaches have been proposed for regret minimization in the single-agent, finite-horizon tabular setting: (i) model-based algorithms \citep{azar2017minimax, dann2017unifying,zanette2019tighter,zhang2024settling}, and (ii) model-free algorithms \citep{jin2018q, zhang2020almost,li2021breaking}.

Both approaches offer algorithms that achieve the minimax optimal lower bound up to poly-logarithmic factors, specifically $\Omega(\sqrt{\sepisode^3 \nstates \nactions \nepisode})$ \citep{jin2018q,domingues2021episodic}. Among these, \texttt{UCBVI} \citep{azar2017minimax}, which is based on the principle of optimism in the face of uncertainty, was the first algorithm to achieve the minimax bound.

\paragraph{Federated Reinforcement Learning.}
The FRL method most closely related to ours is the Byzantine robust distributed \texttt{UCBVI} algorithm \citep{chen2023byzantine}, which assumes homogeneous agents. This algorithm achieves a regression bound of $\tildeo(\sqrt{\sepisode^4 \nstates^2 \nactions \nepisode / \nagent})$ and a communication complexity that scales logarithmically with the number of episodes $\nepisode$.

In contrast, our method achieves a regret of $\tildeo(\sqrt{\sepisode^3 \nstates \nactions \nepisode / \nagent})$, which is optimal in single-agent environments. Moreover, we also provide guarantees in heterogeneous environments.

Other FRL approaches are based on model-free methods. 
\citet{zhang2020almost} proposed a federated variant of Q-learning, achieving a regret of $\tildeo(\sqrt{\sepisode^3 \nstates \nactions \nepisode / \nagent})$, with a communication complexity linear in $ \nepisode$. \citet{zheng2023federated} later reduced the communication cost to $O(\nagent \log \nepisode)$, but introduced an additional factor of $\sepisode$ in the regret bound.
More recently, \cite{zheng2024federated} improved both regret and communication cost. However, their method still requires homogeneous agents, and the communication complexity remains $O(\nagent \log \nepisode)$.

%% file: 2025-AISTATS/main_text/preliminaries.tex
\subsection{Federated Reinforcement Learning}

\paragraph{Policy and Value Functions.} 
A deterministic policy $\pi$ is a set of functions $\pi_h \colon \S \to \A$ where $\policy_{h}(s) \in \A$, $h \in [\sepisode]$. The value function $\valuefunc[h][\,i,\policy]$, is defined as:
\begin{align}
\label{def:value_function_for_agent}\textstyle
\valuefunc[h][\,i,\policy](s) = \CPE[\policy]{\sum_{h'=h}^{\sepisode}\rewardMDP[h'][i](s_{h'}^{i},a_{h'}^{i})}{s_{h}^{i}=s}\eqsp,
\end{align}
where for all $h \leq h' \leq \sepisode$, $a_{h'}^{i} \sim \policy_{h'}^{i}(. | s_{h'}^{i}) $ and for all $h \leq h' \leq \sepisode -1, s_{h'+1}^{i} \sim \kerMDP[h'][i](. | s_{h'}^{i}, a_{h'}^{i})$. Similarly, the Q-function of a policy $ \policy$ for agent $i$ at step $h$ is
\begin{align}
\nonumber\textstyle
\qfunc[h][i, \policy](s,a) :=\CPE[\policy]{\sum_{h'=h}^{\sepisode} \expandafter{\rewardMDP[h'][i]}(s_{h'}^{i},a_{h'}^{i})}{s_{h}^{i}=s, a_{h}^{i}=a} \eqsp,
\end{align}
and satisfies the Bellman equations
\begin{align}\label{eq:bellman_equations}  
    \begin{split}
    \qfunc[h][i,\policy](s,a) &=  \rewardMDP[h][i](s,a) + \kerMDP[h][i]\valuefunc[h+1][\,i,\policy](s,a)\eqsp,\\ 
    \valuefunc[h+1][\,i,\policy](s) &= \qfunc[h][i,\policy](s, \policy_h(s))\eqsp.   \end{split}
\end{align}
Additionally, the optimal Q-value satisfies the optimal Bellman equations
\begin{equation}\label{eq:optimal_bellman_equations}
\begin{split}
    \qfunc[h][i,\star](s,a) &=  \rewardMDP[h][i](s,a) + \kerMDP[h][i]\valuefunc[h+1][\,i,\star](s,a)\eqsp,\\
    \valuefunc[h][\,i,\star](s) &= \max_{a \in \cA} \qfunc[h][i,\star](s,a)\eqsp.
\end{split}
\end{equation}

\paragraph{Learning Protocol.}
 At the beginning of each episode $t \in [\nepisode]$, all agents select a common policy $\pi_t$, which is computed based on the information \textit{exchanged} prior to episode $t$. Subsequently, each agent generates an independent trajectory of length $\sepisode$. At each step $h$, an agent observes its state $s^i_{t,h} \in \S$ and takes an action $a^i_{t,h} = \pi_{t,h}(s^i_{t,h}) \in \A$. The agent then observes the next state $s^i_{t,h+1}$ according to the transition probabilities $\kerMDP[h][i](\cdot | s^i_{t,h}, a^i_{t,h})$ and receives a deterministic reward $\r^i_{t,h} = \r^i_{h}(s^i_{t,h}, a^i_{t,h})$. After generating these trajectories, agents \textit{may} exchange information through the central server.

\paragraph{Federated Regret.}
The performance of the learning algorithm is evaluated using the \textit{federated regret}, defined as
\begin{align}\label{def:regret}        
    \regret(T) := \max_{\pi}\frac{1}{\nagent} \sum_{i=1}^\nagent \sum_{t=1}^\nepisode \valuefunc[1][\,i,\pi](s^i_{t,1}) - \valuefunc[1][\,i, \policy_t](s^i_{t,1}) \eqsp. 
\end{align} 
This regret measures the cumulative difference, in expectation, between the average value of the optimal collaborative policy and the policies used throughout the training procedure.

\paragraph{Communication Complexity and Cost.}

The \textit{communication complexity}, denoted by $\mathfrak{C}(T)$, is defined as the number of episodes where communication between the CS and the agents occurs. The \textit{communication cost} refer to the total number of bits exchanged between the central server and the agents during the learning process. The objective of the FRL algorithm is to simultaneously minimize both the regret $\regret(T)$ and the communication complexity $\mathfrak{C}(T)$.

\subsection{Environmental Heterogeneity}
\label{subsec:environemnthetereogenity}
The environments in which agents evolve may differ from one to another.
However, since agents aim to learn a shared policy, environmental heterogeneity must be small. To measure this, we introduce a new notion of heterogeneity, decomposing each agent's state-action transition kernel into a common part, shared by all agents, and an individual part that reflects unique environmental characteristics. Formally, this is captured by the following assumption.
\begin{assumLHG}
\label{assum:lhg1}
There exists a non-homogeneous transition kernel $\{\comkerMDP[h]\}_{h \in [\sepisode]}$, $\nagent$ individual non-homogeneous transition kernels $\{\perkerMDP[h][i]\}_{h \in [\sepisode]}$ for any $i \in [M]$, and a constant $ \hgkernel \in [0, 1) $, such that for any $i \in [\nagent]$ and $ (s,a,s',h)\in\S \times \A \times \S \times  [\sepisode]$, 
\begin{align}
\nonumber \textstyle
\kerMDP[h][i](s'|s,a) = (1 - \hgkernel )\comkerMDP[h](s'|s,a) + \hgkernel \perkerMDP[h][i](s'|s,a) \eqsp.
\end{align}
\end{assumLHG}
Likewise, we assume that agents receive comparable rewards for a given state-action pair.
\begin{assumLHG}
\label{assum:lhg2} 
There exists a constant $ \hgreward \in [0,1)$ such that for all $(i,j) \in [\nagent],$ and for all $ h\in [\sepisode]$ it holds that 
\begin{align*}
 \norm{\rewardMDP[h][i] -\rewardMDP[h][j]}[\infty] \leq \hgreward \eqsp .
\end{align*}
\end{assumLHG}
Note that \Cref{assum:lhg1} implies the following bound on the difference between the common transition kernel and each agent's transition kernel, measured in $L_1$-norm,
\begin{align}\label{eq:tv_assumption}
\max_{(s,a,h) \in \S \times \A \times [\sepisode]}\norm{\comkerMDP[h](\cdot | s,a) - \kerMDP[h][i](\cdot | s,a)}[1] \leq \hgkernel \eqsp.
\end{align}
We prove this inequality in \Cref{sec:technicallemmas}.
Consequently, \eqref{assum:lhg1} is slightly stronger than \Cref{eq:tv_assumption}, which is the typical assumption in other FRL settings, such as FedSARSA \citep{zhang2024finite} or policy optimization with access to a simulator \citep{jin2022federated,wang2024convergence}. The motivation for using \Cref{assum:lhg1} over \eqref{eq:tv_assumption} lies in the need to control how \textit{samples from $\kerMDP[h][i]$} relate to \emph{samples from $\comkerMDP[h]$}. This connection is crucial in RL without a generative model, as the data generation process is not independent and identically distributed, forcing agents to exploit all the samples they have. In \Cref{sec:feducbvi}, we discuss in detail the necessity of this assumption for our analysis.

%% file: 2025-AISTATS/main_text/FedUCBVI.tex
{\tiny \begin{algorithm*}[t!]
\caption{\algname{\texttt{Fed-UCBVI}}{algo:FEDUCBVI}}
{

    \textbf{Initialization: }$t = 1$; $r = 1$; $\erewardMDP[h]^{\eqsp i}(s,a) = 0$; $N_{(1),h}(s,a) = 0$; $n_{(1,0),h}^{i}(s,a) = 0$; $\eqfunc[(1),h](s,a) =\evaluefunc[(1),h](s) = \sepisode$ for all $(s,a,h,i) \in \S \times \A \times [\sepisode] \times [\nagent]$; $\policy_{(1)} = \{\policy_{(1),h}\}_{h}$ for some policy $\policy_{(1)}$; and $\nu(\delta, \nepisode)$ set as in \eqref{def:sdeltat}.   

    \While{$t \leq \nepisode$}{

        \begin{minipage}{0.93\textwidth}
        \begin{mdframed}[style=myboxstyle]
        \For{each agent $i = 1$ \textbf{to} $\nagent$ \textbf{in parallel}}{
            Set $l = 1$; $n_{(r),h}^{i}(s,a) = n_{(r,0),h}^{i}(s,a) $; and $\hat{N}^{\,i}_{(r,0),h}(s,a) = N_{(r),h}(s,a)$
            
            \While{no synchronization signal}{
                Collect $(s_{t,h}^i, a_{t,h}^i, \r_{t,h}^i, s_{t,h+1}^i)_{1 \leq h \leq \sepisode} $ using $\policy_{(r)}$

                \For{$h = \sepisode$ \textbf{to} $1$}{                
                Set $n_{(r,\ell),h}^{i}(s,a) = n_{(r,\ell-1),h}^{i}(s,a)+ \Ind_{(s,a)}(s_{t,h}^i, a_{t,h}^i)$ for $(s,a) \in \S \times \A$ and
                
                $n_{(r,\ell),h}^{i}(s,a,s') = n_{(r,\ell-1),h}^{i}(s,a,s') + \Ind_{(s,a,s')}(s_{t,h}^i, a_{t,h}^i, s_{t,h+1}^i)$ for $(s,a,s') \in \S \times \A\times \S$ 

                Set $\hat{N}^{\,i}_{(r,\ell),h}(s,a) = \hat{N}^{\,i}_{(r,\ell -1),h}(s,a) + \nagent \Ind_{(s,a)}(s_{t,h}^i, a_{t,h}^i)\}$ for $(s,a) \in \S \times \A$

                Set $\erewardMDP[h]^{\eqsp i}(s_{t,h}^i, a_{t,h}^i) = \r_{t,h}^i$  and 
                $\ell = \ell+1$
                }
                
                \If{$\big( N_{(r),h}(s_{t,h}^i, a_{t,h}^i) \leq \nu(\delta, \nepisode)
                \textnormal{\textbf{ and } }
                n_{(r,\ell),h}^{i}(s_{t,h}^i, a_{t,h}^i) > 2 n_{(r),h}^{i}(s_{t,h}^i, a_{t,h}^i)
                \big)$
                \\
                \text{\hspace{0.5em} \textnormal{\textbf{or}}}
                $\big( N_{(r),h}(s_{t,h}^i, a_{t,h}^i) > \nu(\delta, \nepisode)
                \textnormal{ \textbf{and} }
                \hat{N}^{\,i}_{(r,\ell),h}(s_{t,h}^i, a_{t,h}^i) >  N_{(r),h}(s_{t,h}^i, a_{t,h}^i) \big)$
                 }{
                Send synchronization signal}

            }
        
        Set $t = t+\ell$; $n_{(r+1),h}^{i}(s,a) = n_{(r, \ell),h}^{i}(s,a)$  and update the transition kernels using \eqref{eq:update_of_tr_kernel}
        
        }
        \end{mdframed}
        \end{minipage}
        Set $\evaluefunc[(r+1),\sepisode+1](s) = 0$ for all $s \in \S$ and broadcast it to all the clients  
        
        \For{$h = \sepisode$ \textbf{to} $1$}{
            \begin{minipage}{0.90\textwidth}
            \begin{mdframed}[style=myboxstyle]
             \For{agent $i = 1$ \textbf{to} $\nagent$ \textbf{in parallel}}{
            Compute $\eqfunc[(r+1),h][i]$ using \eqref{eq:local_q_function}
             
            Send $n_{(r+1),h}^{i}$,$ \ekerMDP[(r+1),h][i]\evaluefunc[(r+1),h+1]$, $ \ekerMDP[(r+1),h][i]\evaluefunc[(r+1),h+1]^{\, 2}$, and $\eqfunc[(r+1),h][i]$ to the central server 
            }
            \end{mdframed}
            \end{minipage}
            Compute $N_{(r+1),h}$, $\eqfunc[(r+1),h] $, $\evaluefunc[(r+1),h](s) $, and
            $ \policy_{(r+1),h}$ using \eqref{def:globalcounter}, \eqref{eq:q_agregation},
            \eqref{eq:v_aggregation}, and \eqref{eq:policy_aggregation}
            and broadcast them to all the clients
        }
        Set $r = r + 1$
    }
}
\end{algorithm*}
}
In this section, we present the \algo\ algorithm, which extends the \texttt{UCBVI} algorithm proposed by \cite{azar2017minimax} to the federated learning framework. The process involves multiple communication rounds with a CS. The number of episodes in each communication round (or epoch) \(r\) is random, and each epoch is decomposed into three phases:
\begin{itemize}[itemsep=-5pt]
    \item[(i)] \textit{Data collection}: During this phase, each agent interacts with its environment using the policy \(\pi_{(r)}\) provided by the CS, gathering trajectory data.
    \item[(ii)] \textit{Synchronization}: Once any agent meets the synchronization conditions, it sends a synchronization signal to the central server, which then broadcasts this information to all other agents.
    \item[(iii)] \textit{Policy update}: In this phase, all agents engage in \(\sepisode\) sequential communications with the CS. At each step $h = \sepisode$ to $1$, agents send their local estimates of the \(Q\)-values and other related information related to step \(h\) to the CS. In return, they receive a global estimate of the \(V\)-values, along with an updated policy and related information for that step.
\end{itemize}

The following sections provide a detailed overview of each of these stages.

\paragraph{Data Collection.} At the beginning of round $r$, each agent $i \in [\nagent] $ follows the policy $\pi_{(r)}$ to collect new trajectories. For $\ell \in \nset$, denote by $n^i_{(r,\ell),h}(s,a)$ and $n^i_{(r,\ell),h}(s,a,s')$  the number of visits to a state-action pair $(s,a)$ and the number of transitions from $(s,a)$ to $s'$ at step $h$ after $\ell$ episodes in the round $r$.

\paragraph{Synchronization.}  
At the start of epoch $r$, all agents receive the current global counters
\begin{equation}\label{def:globalcounter}\textstyle
N_{(r),h}(s,a) := \sum_{i=1}^\nagent n_{(r),h}^{i}(s,a),
\end{equation}
where $n_{(r),h}^i(s,a) := n_{(r,0),h}^i(s,a)$ is the number of visits of a state-action pair by agent $i$ prior to round $r$.

During epoch $r$, after $\ell$ episodes, agent $i$ sends a synchronization signal if a newly visited state-action-step triplet $(s,a,h)$ is identified and one of two synchronization conditions is met. These conditions depend on whether the total number of visits $N_{(r),h}(s,a)$ exceeds a threshold $\nu(\delta, \nepisode) = \tcO(\hgkernel T H \nagent + \nagent)$ (see \Cref{def:sdeltat} in \Cref{app:communicationanalysis} for the full expression).

1) \textit{Local Doubling Condition.} If $N_{(r),h}(s,a) \leq \nu(\delta, \nepisode)$, an agent $i$ sends the synchronization signal if
\begin{equation}\label{eq:local_doubling_cond}
    n^i_{(r,\ell),h}(s,a) > 2n^i_{(r),h}(s,a)\eqsp.
\end{equation}

2) \textit{Globally Estimated Doubling Condition.} If $N_{(r),h}(s,a) > \nu(\delta, \nepisode)$, agent $i$ sends the synchronization signal if
\begin{equation}\label{eq:global_est_doubling_cond}
    \hat{N}^{\,i}_{(r,\ell),h}(s,a) > 2 N_{(r),h}(s,a)\eqsp,
\end{equation}
where $\hat{N}^{\,i}_{(r,\ell),h}(s,a)$ is an estimate of $\sum_{i=1}^N n^i_{(r,\ell),h}(s,a)$ based on the information available to agent $i$.

\paragraph{Policy Update.} Upon receiving the synchronization signal, each agent computes its local estimates of transition probabilities as
\begin{equation}
\label{eq:update_of_tr_kernel} 
\textstyle
\ekerMDP[(r+1),h][i](s'|s,a) := \frac{n_{(r+1),h}^{i}(s,a,s')}{n_{(r+1),h}^{i}(s,a)}
\end{equation}
if $n_{(r+1),h}^{i}(s,a)\!> \! 0$,  otherwise $\ekerMDP[(r+1),h][i](s'|s,a) \!:= 1/\nstates$.

Next, the agents and the central server exchange their $Q$- and $V$-value estimates. For $h=H,\ldots,1$, each agent computes the local $Q$-value estimate
\begin{align}
\label{eq:local_q_function}
\textstyle
\! \! \eqfunc[(r+1),h][i](s,a) := \left[\erewardMDP[h]^{\eqsp i} + \ekerMDP[(r+1),h][i] \evaluefunc[(r+1),h+1]\right](s,a),
\end{align}
using the global value estimate $\evaluefunc[(r+1),h+1]$ previously received from the CS; note that for $h=H$, this value is set to zero and does not require communication.

Then, the CS collects the local $Q$-value estimates from all agents, along with additional information necessary to compute a Bernstein-like bonus function $\bonus_{(r+1),h}(s,a)$ (see \eqref{eq:bonus_function} in Appendix for an exact expression). The aggregated $Q$-value is computed as
\begin{equation}
\label{eq:q_agregation}
\textstyle
\!\!\!\eqfunc[(r+1),h](s,a) 
\!\!:=\! \min
\!\big(  [\mathcal{T}_{(r+1),h}^\aggw \!+ \bonus_{(r+1),h}](s,a), \!
\sepisode \big)
~,
\end{equation}
with $\mathcal{T}_{(r+1),h}^\aggw (s , a) = \sum_{i=1}^\nagent \! \aggw^i_{(r+1),h}(s,a) \eqfunc[(r+1),h][i](s,a)$, and
\begin{equation}\label{eq:aggergating_weights}\textstyle
    \aggw^i_{(r+1),h}(s,a) := \frac{n_{(r+1),h}^{i}(s,a)}{N_{(r+1),h}(s,a)}
    \eqsp.
\end{equation}
Finally, the central server updates the value function and policy according to the equations
\begin{align}
\label{eq:v_aggregation}
\textstyle
\evaluefunc[(r+1),h](s) & := \max_{a \in \A} \eqfunc[(r+1),h](s,a) \eqsp, \\
\label{eq:policy_aggregation}
\textstyle 
\policy_{(r+1),h}(s) & := \argmax_{a \in \A} \eqfunc[(r+1),h](s,a) \eqsp.
\end{align}

These updated values are distributed to all agents, and the process continues for all $h = H,\ldots,1$. Once $h=1$ is reached, the new epoch $r+1$ begins.

\paragraph{Communication Complexity.} Our algorithmic design shares similarities with previous work on reinforcement learning with low switching cost \citep{bai2019provably,zhang2020almost,qiao2022sample}. In particular, the number of times the local data collection policy changes—known as the switching cost—directly corresponds to the number of communication rounds in our framework, which we define as the communication complexity.
In its simplest form, the doubling condition in this context can be expressed as: \begin{equation}\label{eq:doubling_condition_bai2019}
 \exists (s,a,h) : N_{(r, \ell),h}(s,a) > 2 N_{(r),h}(s,a),, \end{equation} where $N_{(r,\ell),h}(s,a) := \sum_{i=1}^{\nagent} n^i_{(r,\ell),h}(s,a)$ represents the cumulative count across agents. 
 
However, this condition cannot be directly verified in a federated learning setting, as the value of $N_{(r,\ell),h}(s,a)$ is not accessible to any individual agent.
One potential solution is to use a weaker local doubling condition, as defined in \eqref{eq:local_doubling_cond}. However, this approach results in communication complexity scaling linearly with the number of agents $\nagent$, which is impractical for large-scale federated learning environments. Instead, we propose to construct an estimate of the global counter $\hat{N}^i_{(r,\ell),h}(s,a)$ to serve as a plug-in estimate on the left-hand side of \eqref{eq:doubling_condition_bai2019}. This is the core idea behind the condition in \eqref{eq:global_est_doubling_cond}. While such estimates may be inaccurate during the initial stages of training, they become reliable once the number of visits exceeds a threshold $\nu(\delta, \nepisode)  = \tcO(\hgkernel T H \nagent + \nagent)$, defined in \eqref{def:sdeltat}. At that point, $\hat{N}^i_{(r,\ell),h}(s,a)$ can be effectively used as a plug-in estimate. Using this approach, we establish a bound on the communication complexity of \algo.

\begin{restatable}[Communication Complexity]{lemma}{comcomplexity}
\label{lem:communication}
With probability at least $1- \delta$, the number of communication rounds of \algo\ is bounded by
\begin{align*}
\mathfrak{C}(T) \leq \cO \big( \nstates  \nactions \sepisode &\log\nepisode  + \nagent \nstates  \nactions \sepisode \log \log \nepisode \\ 
& + \nagent \nstates  \nactions \sepisode\log(1 + \hgkernel \nepisode) \big) \eqsp,
\end{align*}
where logarithmic dependence in $\nstates, \nactions$, $\sepisode$, $1/\delta$ and $\nagent$ is ignored.
\end{restatable}
\paragraph{\emph{Sketch of the proof:}} 
To prove the result, we consider a fixed triplet $(s,a,h)$ and count how many synchronizations this triplet can trigger. Let $k_{s,a,h}^{\min}$ represent the index of the last round where $\counter_{(r),h}(s,a) \leq \nu(\delta, \nepisode)$. To bound the number of synchronizations that occur between the first round and round $k_{s,a,h}^{\min}$, note that agents send an abort signal only when their local visit count of $(s,a)$ at time $h$ has doubled. This can happen at most $\log_2(\nu(\delta, \nepisode))$ times for an individual agent, and for all agents combined, the total is upper bounded by $\nagent\log_2(\nu(\delta, \nepisode))$.

Next, we bound the number of synchronizations between round $k_{s,a,h}^{\min}$ and the final round. By applying a Bernstein-type concentration inequality, we can show that the synchronization rule \eqref{eq:global_est_doubling_cond} implies the equivalent of \eqref{eq:doubling_condition_bai2019}, although with a coefficient of $8/7$ instead of $2$ on the right-hand side. Using a similar argument as above, we obtain $\cO(\log(\nagent))$ synchronizations triggered by a single state-action-step triplet. 
We complete the proof by summing these bounds over all $(s,a,h)$ and using the expression of $\nu(\delta, \nepisode)$. \hfill$\square$\par

A complete proof of \Cref{lem:communication} is provided in \Cref{app:communicationanalysis}. Importantly, we observe that, in the homogeneous setting, the linear dependence on $\nagent$ vanishes. %
Moreover, we can estimate the communication cost, i.e., the number of bits exchanged, by noting that in each communication round, each agent transmits objects of size at most $\nstates \nactions \sepisode$.

\paragraph{Computational and Space Complexity.}
First, we remark that, at all times, agents store objects of size $\cO(\nstates^2 \nactions \sepisode)$. %
At every episode, agents perform $\cO(1)$ operations, while they perform $\cO(\nstates^2 \nactions\sepisode)$ operations at communication times. By \Cref{lem:communication}, we deduce that the computation complexity of this algorithm is $\cO(\nepisode +  \nstates^3  \nactions^2 \sepisode^2 \log\nepisode  + \nagent \nstates  \nactions \sepisode \log \log \nepisode 
 + \nagent \nstates  \nactions \sepisode\log(1 + \hgkernel \nepisode) )$ for all $T$ episodes. 
 
 \textbf{Regret Bound.} We now state our main result, which bounds the federated regret of \algo.
\begin{theorem}
\label{thm:regret}
With probability at least $1- \delta$, the following bound on the regret of \algo\ holds
\begin{align*}
\begin{split}
\regret(T) & = \tcO\left(  \sqrt{\sepisode^3 \nstates \nactions \nepisode / \nagent} +  \sepisode^3 \nstates^2 \nactions\right)\\
&+ \tcO\left( \nepisode \sepisode (\sepisode \hgkernel +   \hgreward)\right) \eqsp.
\end{split}
\end{align*}
\end{theorem}
We give a sketch of the proof below, and postpone the detailed proof to \Cref{sec:regretanalysis}.

In the homogeneous setting, where $\hgkernel = \hgreward = 0$, we recover the expected linear speedup in number of agents and achieve a minimax optimal regret bound up to logarithmic factors (see \Cref{tab:comparaison_table} for comparisons). In contrast, in the heterogeneous setting, an additional term, that scales linearly with the degree of heterogeneity, emerges. We show in \Cref{lem:performance-difference-lb} in Appendix that this is expected, and comes from the fact that, in some cases, a policy optimal for one agent is sub-optimal by at least $\hgkernel \sepisode^2$ for another agent. This illustrates the trade-off involved in cooperation between heterogeneous agents: if the degree of heterogeneity is too large, cooperation can become counterproductive. %

\paragraph{\emph{Sketch of the proof:}}
As a first step of the proof, we reduce the problem of minimizing the federated regret \eqref{def:regret} to the problem of minimizing a \textit{common} regret. We introduce the common MDP $\cM^{\common}$ as follows
\begin{align}
\label{eq:comMDP}\textstyle
\cM^{\common} := (\S, \A, \sepisode, \{\rewardMDP[h][\common] := \frac{1}{\nagent}\sum_{i=1}^\nagent \rewardMDP[h][i]\}_{h } , \{\comkerMDP[h]\}_{h} )
\eqsp,
\end{align}
where $\{\comkerMDP[h]\}_{h}$ is defined in \Cref{assum:lhg1}.
We set $\comvaluefunc[h][\pi]$ and $\comstarvaluefunc[h]$ the value-function of a policy $\pi$ and optimal value-function in  $\cM^{\common}$. The common regret is defined as
\begin{equation}\label{def:common_regret}
\regret^{\common}(T) := \frac{1}{\nagent} \sum_{t=1}^T \sum_{i=1}^\nagent \comstarvaluefunc[1](s_{t,1}^i) - \comvaluefunc[1][\pi_{t}](s_{t,1}^i)
\eqsp.
\end{equation}
Adapting the performance-difference lemma of \cite{russo2019worst} under \Cref{assum:lhg1}, it may be shown that
\begin{align*}
\regret(T) &=  \max_{\pi} \frac{1}{\nagent} \sum_{t=1}^T \sum_{i=1}^{\nagent} \valuefunc[1][i,\pi](s_{t,1}^i) - \valuefunc[1][i, \pi_{t}](s_{t,1}^i) \\
&\leq \regret^{\common}(\nepisode)  + 2 \nepisode \varepsilon_{\kerMDP} \sepisode^2 + 2 \nepisode\varepsilon_{\r} \sepisode
\eqsp.
\end{align*}
As shown in \Cref{lem:performance-difference-lb}, the scaling $\cO(\nepisode( \hgkernel \sepisode^2 + \hgreward \sepisode))$ with $\sepisode^2 $ is unavoidable.

The remainder of the proof involves three key steps, outlined below. The first step focuses on estimating the common transition kernel and introduces the primary technical innovations of this work. It also provides justification of \Cref{assum:lhg2}.

\paragraph{Step 1: Estimation of the common transition kernel.}

First, we prove that the weighted average kernel, 
\[
    \textstyle \ekerMDP[(r),h](s'|s,a) := \sum_{i=1}^\nagent \aggw^i_{(r),h}(s,a) \cdot \ekerMDP[(r),h][i](s'|s,a)\eqsp,
\]
where weights are defined in \eqref{eq:aggergating_weights}, forms a well-defined (biased) estimator of the common transition kernel \(\comkerMDP[h]\) using data from all agents. Importantly, neither the agents nor the CS have direct access to this quantity.

The analysis of \(\ekerMDP[(r),h]\), under \Cref{assum:lhg2} poses significant challenges compared to both the generative model setting and the case involving homogeneous agents. To illustrate, the kernel can be reformulated as follows, incorporating all samples from the agents:
\[
\textstyle
\ekerMDP[(r),h](s'|s,a) = \frac{1}{N_{(r),h}(s,a)}\sum_{i=1}^M n^i_{(r),h}(s,a,s')\eqsp.
\]
In the homogeneous scenario, where \(\hgkernel = 0\), as explored in prior work \citep{zheng2023federated,zheng2024federated}, the estimate is derived from an i.i.d. sequence of categorical random variable samples from \(\comkerMDP[h](\cdot|s,a)\), simplifying the analysis. Moreover, within the generative model framework, such as in \citep{jin2022federated,wang2024convergence}, we can ensure an equal sample count from each agent’s transition kernel \(\kerMDP[h][i]\), resulting in \(\ekerMDP[(r),h]\) as a simple mean of independent biased estimates of the common kernel.

However, in our setting, the estimator \(\ekerMDP[(r),h]\) incorporates a random and non-stationary number of samples from each agent, making standard techniques of conditioning on a total sample size $N_{(r),h}(s,a)$ inapplicable. Using union-bound arguments to account for the variability in sample sizes across agents $\{n^i_{(r),h}(s,a)\}_{i \in [\nagent]}$ results in an exponential number of configurations with respect to \(\nagent\), constraining any possibility of linear speed-up.

 Using \Cref{assum:lhg2}, every kernel $\kerMDP[h][i]$ is a mixture of $\comkerMDP[h]$ and $\perkerMDP[h][i]$. The samples obtained by agent $i$ as a mixture of samples coming from the two latter kernels: sample $s^i_{t,h}$ is with probability $1- \hgkernel$ generated from $\comkerMDP[h]$, and with probability $\hgkernel$ from $\perkerMDP[h][i]$. We define a \textit{virtual estimate of the common kernel}, $\ecomkerMDP[(r),h]$, for each communication round $r$, representing the estimate we would have obtained if all samples were drawn solely from $\comkerMDP[h]$. This estimate is subject to a bias resulting from the heterogeneity. 
\begin{align}
\label{eq:proof_sketch_concentration}
\textstyle
\! \! \! \left\Vert (\ekerMDP[(r),h] -  \ecomkerMDP[(r),h])(\cdot | s,a)\right\Vert_1 = \tcO\left(\varepsilon_{\kerMDP} + \frac{1}{N}\right) \eqsp,
\end{align}
where $N = \counter_{(r),h}(s,a)$, that holds for any $(r,s,a,h)\in [\mathfrak{C}(T)] \times \S \times \A \times \S$.

\paragraph{Step 2: Optimism.} 
In our setting, our estimates are not optimistic due to the presence of heterogeneity; however, we can show the analog of the required properties $   \evaluefunc[(r),h](s) \geq \comstarvaluefunc[h](s) - (2\hgreward + 3\hgkernel \sepisode) (\sepisode+ 1-h)), \text{ for any } r \text{ and } (s,h) \in \S \times [\sepisode]$. The key ingredients are concentration inequalities, an inequality \eqref{eq:proof_sketch_concentration} and Lemma~14 of \citep{zhang2021reinforcement}; see also \Cref{lem:monotonicity} in Appendix. The proof is carried out by induction on $h$. Applying the update rule \eqref{eq:update_rule}, combined with a simple rearranging of the terms, yields
\begin{align*}
\small
&\eqfunc[(r),h](s,a) \ge \comstarqfunc[h](s,a)  +  \underbrace{( \ecomkerMDP[(r),h] -\comkerMDP[h])\comstarvaluefunc[h+1](s,a)}_{\term{IV}: \eqsp \textbf{concentration error}}
\\
&\quad   + \underbrace{\sum_{i=1}^\nagent \aggw^i_{(r),h}(s,a) \erewardMDP[h]^{\,i}(s,a) - \frac{1}{\nagent} \sum_{i=1}^{\nagent} \rewardMDP[h][i](s,a)}_{\term{I}: \eqsp \textbf{reward heterogeneity error}} \\
&\quad  + \underbrace{\ekerMDP[(r),h]  (\evaluefunc[(r),h+1](s,a) - \comstarvaluefunc[h+1](s,a))}_{\term{II}: \eqsp \textbf{correction error}}\\
&\quad + \underbrace{(\ekerMDP[(r),h]-\ecomkerMDP[(r),h])\comstarvaluefunc[h+1](s,a)}_{\term{III}: \eqsp \textbf{transition heterogeneity error}} +\, \bonus_{(r),h}(s,a) \eqsp.
\end{align*}
Terms $\term{II}$ and $\term{IV}$, which represent the correction and the concentration errors, are standard and are controlled using respectively induction hypothesis, \Cref{lem:monotonicity} and standard deviation inequalities. We control $\term{I}$ by applying \Cref{assum:lhg2} and noticing that the convex combination of $\erewardMDP[h]^{\,i}(s,a)$ is also a convex combination of the true rewards $\rewardMDP[h][i](s,a)$. Finally, to control $\term{III}$ we combine Holder's inequality and inequality \eqref{eq:proof_sketch_concentration}. An appropriate choice of the exploration bonus concludes the statement. 

\textbf{Step 3: Bounding the regret.} For each quantity indexed by the number of communication rounds $r$ (e.g. $\evaluefunc[(r),h]$), we introduce a corresponding quantity indexed by the episode number $t$ (e.g. $\evaluefunc[t,h]$), defined as the value of the former at the last communication round before $t$ (see \eqref{def:changing_timescale} in Appendix for formal definitions). Next, following the approach of \cite{azar2017minimax}, we define $\delta^i_{t,h} = \evaluefunc[t,h](s_{t,h}^i) - \comvaluefunc [1][\pi_{t}](s_{t,h}^i)$ and analyze this term independently
\begin{align*}
\small \delta^i_{t,h} &\leq  \delta^i_{t,h+1} 
 + \underbrace{[\ekerMDP[t,h] - \ecomkerMDP[t,h]] \evaluefunc[t,h+1](s^i_{t,h}, a^i_{t,h})}_{\term{A}: \eqsp \textbf{heterogeneity error}} +  \zeta^i_{t,h}  \\
&+ \underbrace{[\ecomkerMDP[t,h] - \comkerMDP[h]]\left[\evaluefunc[t,h+1] - \comstarvaluefunc[h+1]\right](s^i_{t,h}, a^i_{t,h})}_{\term{B}: \eqsp \textbf{correction error}} + 2\hgreward \\
&+ \underbrace{[\ecomkerMDP[t,h] - \comkerMDP[h]]\comstarvaluefunc[h+1](s^i_{t,h}, a^i_{t,h})}_{\term{C}: \eqsp \textbf{concentration error}}  + b_{t,h}(s^i_{t,h},a^i_{t,h})\\
&+ \underbrace{[\comkerMDP[h] - \kerMDP[h][i]]\left[ \evaluefunc[t,h+1] -  \comvaluefunc[h+1][\pi_t]\right](s^i_{t,h}, a^i_{t,h})}_{\term{D}: \eqsp \textbf{heterogeneity error}} 
\eqsp,
\end{align*}
where $\zeta^i_{t,h} $ is a martingale term defined in \eqref{def:zeta_t_h}. The analysis of $\term{C}$ and $\zeta^i_{t,h} $ is standard in the literature. To bound $\term{A}$ we employ a combination of \eqref{eq:proof_sketch_concentration} and Holder's inequality.  The bound on $\term{D}$ also combines Holder's inequality and \Cref{lem:tvbound}. The standard recursion argument concludes the proof.\hfill$\square$\par

%% file: 2025-AISTATS/main_text/Experiments.tex
\begin{figure}[t]
    \centering
    
     \begin{subfigure}[b]{0.5\linewidth}
         \centering
         \includegraphics[width=\textwidth]{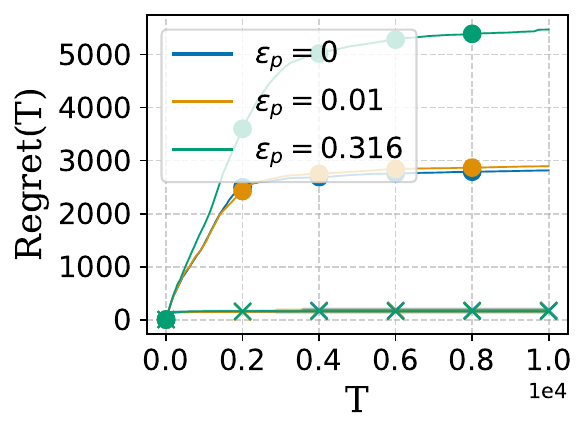}
         \caption{GridWorld}
     \end{subfigure}%
     \begin{subfigure}[b]{0.5\linewidth}
         \centering
         \includegraphics[width=\textwidth]{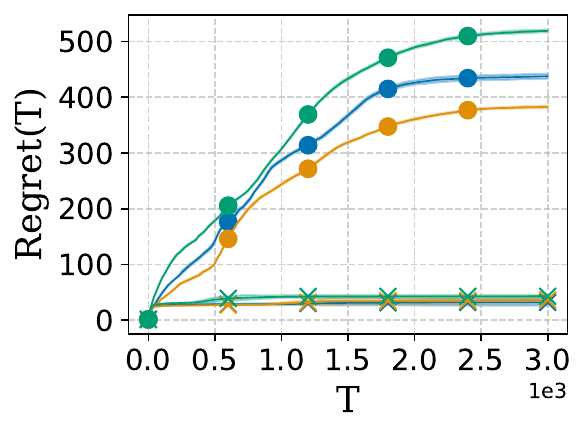}
         \caption{Synthetic}
     \end{subfigure}
     
    \caption{Common regret  \textit{(lower is better)} for $\nagent = 20$ agents as a function of $T$ for different $\hgkernel$: crosses represent \algo, and circles \texttt{FedQ-Bernstein}.}
    \vspace{-1.em}
\label{fig:regretvsepsilon}
\end{figure}

\begin{figure}[t]
    \centering
    
     \begin{subfigure}[b]{0.5\linewidth}
         \centering
         \includegraphics[width=\textwidth]{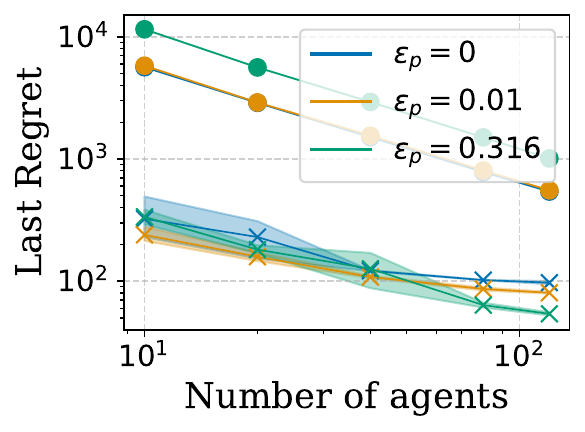}
         \caption{GridWorld}
     \end{subfigure}%
     \begin{subfigure}[b]{0.5\linewidth}
         \centering
         \includegraphics[width=\textwidth]{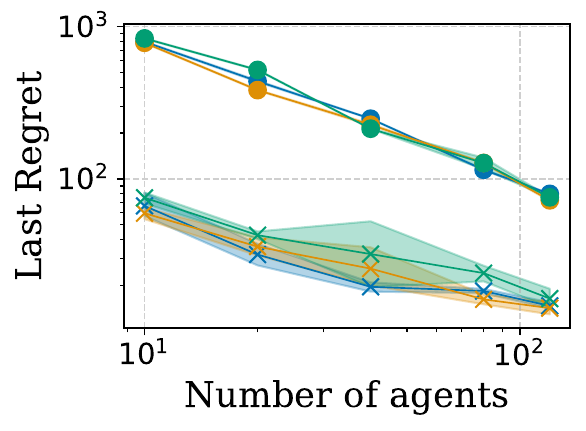}
         \caption{Synthetic}
     \end{subfigure}
        \caption{Common regret \textit{(lower is better)}, at $\nepisode= 3 \cdot 10^4$ for \texttt{GridWorld}, and $\nepisode=3 \cdot 10^3$ for \texttt{synthetic} as a function of $\nagent$ for different $\hgkernel$ in a log-log scale: crosses represent \algo, and circles represent \texttt{FedQ-Bernstein}.}
        \vspace{-1.em}
    \label{fig:linearspeedup}
\end{figure}

In this section, we study the empirical performance \footnote{Our code is available online on GitHub: \url{https://github.com/Labbi-Safwan/Fed-UCBVI}} of \algo, and compare it with the \texttt{FedQ-Bernstein} algorithm \citep{zheng2023federated} on two environments.

\paragraph{Environments.}
We consider two environments specifically designed to satisfy \Cref{assum:lhg1} and \Cref{assum:lhg2}.
In both environments, transitions are defined using two distinct kernels: with probability $1 - \hgkernel$, the agent follow the global kernel, and with probability $\hgkernel$,  it follows an individualized kernel.
The first environment is based on \texttt{GridWorld} \citep{rlberry}, where the agent navigates a grid to reach a target. Upon reaching the target, the agent receives a reward of +1; otherwise, the reward is 0. At each step, the agent selects one of four possible directions (up, down, left, or right). Under the global transition kernel, the agent moves to the intended square with a probability of 0.8, and to a random neighboring square with the remaining probability. In the individual transition kernels, the agent's movement to neighboring squares follows a probability distribution unique to each agent.
We use a \(3 \times 3\) grid with a wall located at coordinate \((1, 1)\), resulting in \(\nstates = 8\) possible states. The planning horizon is set to \(\sepisode = 10\), with the agent starting at coordinate \((0, 0)\) and aiming to reach the target at \((2, 2)\).

The second environment is a \texttt{synthetic} setting, modeled after \citet{zheng2023federated}, with \(\nstates = 5\), \(\nactions = 5\), and \(\sepisode = 5\). All agents share the same reward function \(\rewardMDP[h]{(s, a)}\), with rewards drawn uniformly from \([0, 1]\) for each \((s, a, h) \in \S \times \A \times [\sepisode]\). For each \(s, a, h\), the common and individual transition kernels are drawn uniformly at random from the \(\nstates\)-dimensional simplex.

In all results, we report the common regret instead of the federated regret to simplify computations. Experiments were conducted on a computer with an Intel Xeon 6534 and 196GB RAM. We report the average over $5$ runs and the standard deviation in all the plots. 
The code is provided in the supplementary material.

\paragraph{\hypersetup{allcolors=black}Impact of Heterogeneity.}
In \Cref{fig:regretvsepsilon}, we present the regret of \algo\, for various values of \(\hgkernel\).
\algo's regret is significantly lower than that of \texttt{FedQ-Bernstein}, reflecting similar performance gaps as observed in the single-agent setting.
Moreover, as predicted by our theoretical analysis, increasing \(\hgkernel\) only incurs a slight increase in \algo's regret, due to the additional term scaling linearly with \(\nepisode\).

\paragraph{\hypersetup{allcolors=black}\algo\, has linear speed-up.}
In \Cref{fig:linearspeedup}, we evaluate the regret after \(\nepisode\) iterations of training with varying numbers of agents \(\nagent\) across different levels of heterogeneity. As shown in \Cref{thm:regret}, the regret decreases as \(\nagent\) increases. Notably, this trend persists even in high-heterogeneity settings, highlighting the robust empirical performance of our approach.

\begin{figure}[t]
    \centering
    
     \begin{subfigure}[b]{0.5\linewidth}
         \centering
         \includegraphics[width=\textwidth]{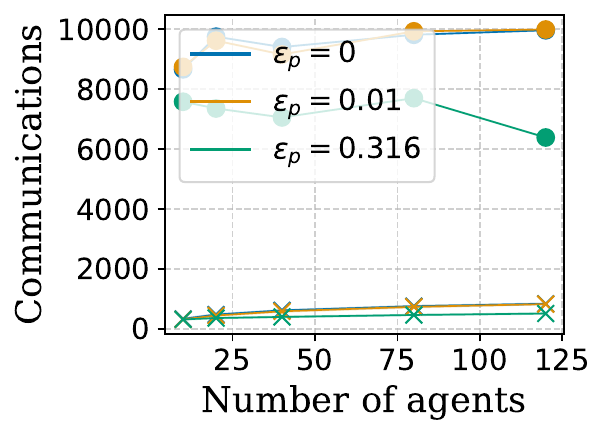}
         \caption{GridWorld}
     \end{subfigure}%
     \begin{subfigure}[b]{0.5\linewidth}
         \centering
         \includegraphics[width=\textwidth]{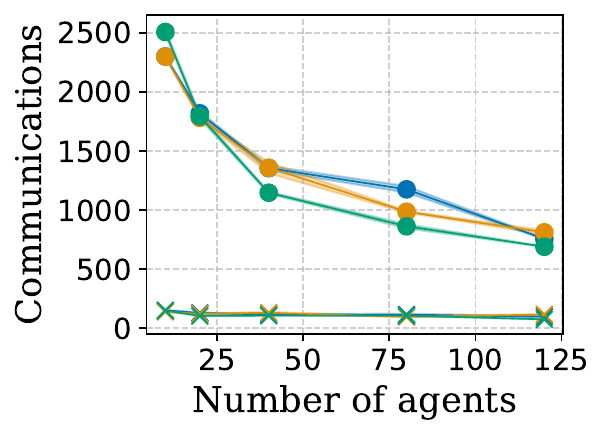}
         \caption{Synthetic}
     \end{subfigure}
     
    \caption{Number of communication \textit{(lower is better)} as a function of $\nagent$ for different $\hgkernel$ and $\nepisode=3 \cdot 10^4$ for \texttt{GridWorld}, $\nepisode=3 \cdot 10^3$ for \texttt{synthetic}: crosses represent \algo, and circles represent \texttt{FedQ-Bernstein}.}
    \vspace{-1.em}
\label{fig:communications}

\end{figure}

\paragraph{\hypersetup{allcolors=black}\algo's communication complexity is small.}
In \Cref{fig:communications}, we observe that the communication complexity of \algo\, is significantly lower than the number of iterations \(\nepisode\) and increases only marginally with the number of agents \(\nagent\). This aligns with the results of \Cref{lem:communication}.  In contrast, \texttt{FedQ-Bernstein} exhibits consistently high communication complexity. The reduced communication in \algo\ results from our novel method for triggering communication rounds based on local estimates of global counters, validating the effectiveness of this approach.

%% file: 2025-AISTATS/main_text/Conclusion.tex
In this paper, we presented \algo, a federated reinforcement learning method based on a new aggregation strategy that reduces communication cost and handles heterogeneous agents. 
We introduced a novel measure of heterogeneity, under which we provide a formal analysis of \algo's regret, showing that it nearly matches minimax optimal regret bounds. To our knowledge, this is the first federated regret analysis with guarantees in heterogeneous environments.
Furthermore, our method provably removes the linear dependence of the communication complexity on $\nagent \log \nepisode$. 
A promising direction for future work is to reduce the communication cost further, by developing new methods that correct for heterogeneity.

%% file: 2025-AISTATS/appendix/notation.tex
\allowdisplaybreaks

\section{NOTATION}
For clarity, we summarize here the notations that we use
\begin{table}[h]
    \centering
    \begin{tabular}{c c c}
    \toprule
         Symbols & Meaning & Definition \\
    \midrule
        $\mathfrak{C}(T)$ & Number of communication rounds performed in average & \Cref{sec:prelimaries}\\
         $\regret(T)$ & Federated regret of the algorithm &   \Cref{def:regret}\\
    \midrule
        $\S$ & State space  & \Cref{sec:prelimaries}\\
         $\A$ & Action space &   \Cref{sec:prelimaries}\\
        $\nagent$ & Number of agents &  \Cref{sec:prelimaries}\\
        $\nepisode$ & Total number of collected episodes per agent & \Cref{sec:prelimaries} \\
        $\sepisode$ & Length of an episode  &  \Cref{sec:prelimaries}\\
        $ \troundsmax $ & Maximum number of communication rounds & \Cref{def:roundsmax} \\ 
    \midrule
        $\kerMDP[h][i]$ & Transition kernel at step $h$ of agent $i$ & \Cref{sec:prelimaries}\\
        $\comkerMDP[h]$ & Common transition kernel at step $h$ & \Cref{assum:lhg1}\\
        $\perkerMDP[h][i]$ & Individual transition kernel at step $h$ & \Cref{assum:lhg1}\\
        $\hgkernel$ & Degree of heterogeneity on the transition kernels & \Cref{assum:lhg1}\\
        $\hgreward$ & Degree of heterogeneity on the rewards & \Cref{assum:lhg2}\\
        $\rewardMDP[h][i]$ & Reward at step $h$ of agent $i$ & \Cref{sec:prelimaries}\\
         $\rewardMDP[h][\common] $ & Reward function of the common MDP & \Cref{eq:comMDP} \\
        $\qfunc[h][i,\policy]$ & Q-function of a policy $\policy $ at step $h$ of agent $i$ & \Cref{eq:bellman_equations}\\
        $\valuefunc[h][i,\policy]$ &  Value function of a policy $\policy $ at step $h$ in the $i$-th MDP & \Cref{eq:bellman_equations} \\
        $    \qfunc[h][i,\star]$ & Optimal Q-function at step $h$ of agent $i$ in the $i$-th environment & \Cref{eq:optimal_bellman_equations}\\
        $\valuefunc[h][i,\star]$ &  Optimal value function at step $h$ of agent $i$ in the $i$-th environment & \Cref{eq:optimal_bellman_equations}\\
    \midrule
        $\nu(\delta, \nepisode)$ & Threshold for defining the condition on initiating  the aggregation signal& \Cref{def:sdeltat} \\
        $\ekerMDP[(r),h][i]$ & Estimated transition kernel during the round $r$ by agent $i$ at step $ h$ & \Cref{eq:update_of_tr_kernel} \vspace{0.1em}\\
        $\ecomkerMDP[(r),h]$ & Virtual estimate of common transition kernel by agent $i$ at step $ h$ & \Cref{eq:virtial_estimate_common_kernel} \vspace{0.1em}\\
        $\erewardMDP[h]^{\eqsp i}$ & Estimated reward at step $h$ of agent $i$ & \algo \vspace{0.1em}\\
        $n_{(r,\ell),h}^{i}$ & Local counter of the cumulative number of visits at the level of agent $i$  & \algo  \\
        $N_{(r),h}$ & Global counter of the cumulative number of visits over all the agents & \Cref{def:globalcounter} \\
        $\hat{N}_{(r,\ell),h}^i$ & Local estimator of agent $i$ of the \textit{true} cumulative number of visits & \algo \\
        $\bonus_{(r),h}$ & Bonus function used in round $r$ and step $\sepisode$ & 
        \Cref{eq:bonus_function} \vspace{0.1em}\\
        $\eqfunc[(r),h][i](s,a)$ & Estimator of the Q-function at the level of agent $i$ & \Cref{eq:local_q_function} \vspace{0.1em}\\
        $\ekerMDP[(r),h]$ & Weighted average of $\{\ekerMDP[(r),h][i]\}_{i}$ during the round $r$ at step $ h$ & \Cref{eq:virtial_estimate_common_kernel} \vspace{0.1em} \\
        $\eqfunc[(r),h](s,a) $ &  Global estimator of the Q-function  & \Cref{eq:q_agregation}  \\
        $ \evaluefunc[(r),h](s)$  & Global estimator of the value function  & \Cref{eq:v_aggregation} \\ 
        $\empvar[(r),h]{f}(s,a)$ & Variance of a function $f$ with respect to $\ekerMDP[(r),h](\cdot |s,a)$ & \Cref{def:expectation_variance} \\
        $\comempvar[(r),h]{f}(s,a)$ & Variance of a function $f$ with respect to $\ecomkerMDP[(r),h](\cdot |s,a)$ & \Cref{def:expectation_variance}\\
    \bottomrule
    \end{tabular}
    \caption{Summary of the notations.}
    \label{tab:summary_notations}
\end{table}

\newpage
Let $(X, \mathcal{X})$ be a measurable space. For any probability measures $ \mathsf{P}$ and $\mathsf{Q}$ on $(X, \mathcal{X})$, and for any $f: X \rightarrow \rset$ we define 
\begin{align}
\label{def:expectation_variance}
\mathsf{P} f:= \PE_{s \sim \mathsf{P}}[f(s)] \eqsp,
\quad \Var_{\mathsf{P}}{f} := \PE_{s \sim \mathsf{P}}[(f(s) - \mathsf{P} f)^2] \eqsp.
\end{align}
For any probability measures $ \mathsf{P}$ and $\mathsf{Q}$ on $(X, \mathcal{X})$, the Kullback-Leibler~divergence $\KL(\mathsf{P} \Vert \mathsf{Q})$ is given by
$$
\KL\left(\mathsf{P} \Big\Vert \mathsf{Q} \right)  := \begin{cases}\mathbb{E}_{\mathsf{P}}\left[\log \frac{\mathrm{d} \mathsf{P}}{\mathrm{~d} \mathsf{Q}}\right], & \mathsf{P} \ll \mathsf{Q}  \eqsp, \\ +\infty, & \text { otherwise }  \eqsp. \end{cases}
$$

Let $A$ be an element of the $\sigma$-algebra $\mathcal{X}$.  We define the indicator function of $A$ as 
\begin{equation}
\begin{split}
\nonumber
\Ind_{A}(\cdot) \colon X
& \longrightarrow \{0,1\}
\\
x
& \longmapsto
\begin{cases}1 & \text{ if } x\in A  \eqsp, \\ 0, & \text { otherwise }  \eqsp. \end{cases}
\end{split}
\end{equation}

We define the indicator function of an element $x\in X$ as 
\begin{equation}
\begin{split}
\nonumber
\Ind_{x}(\cdot) \colon X
& \longrightarrow \{0,1\}
\\
y
& \longmapsto
\begin{cases}1 & \text{ if } x=y  \eqsp, \\ 0, & \text { otherwise }  \eqsp. \end{cases}
\end{split}
\end{equation}

We write $f(\nstates, \nactions, \sepisode, \nepisode, \nagent)   = \cO(g(\nstates, \nactions, \sepisode, \nepisode, \nagent, \delta))$ if there exists $S_0, A_0, \sepisode_0, \nepisode_0, \delta_0$ and a constant $C$ such that for any $\nstates \geq S_0, \nactions \geq A_0, \sepisode \geq \sepisode_0, \nepisode \geq \nepisode_0, $  and $\delta \leq \delta_0$, we have $f(\nstates, \nactions, \sepisode, \nepisode, \nagent) \leq C \cdot g(\nstates, \nactions, \sepisode, \nepisode, \nagent, \delta)$. We say that $f(\nstates, \nactions, \sepisode, \nepisode, \nagent) = \tcO(g(\nstates, \nactions, \sepisode, \nepisode, \nagent, \delta))$ if in the previous bound $C$ is a poly-logarihmic function with respect to the variables $\nstates, \nactions, \sepisode, \nepisode, \nagent, \delta$.

For $ a \in \mathbb{N} $, define $[a] $ as the set of all natural numbers from 1 to $ a $:
\[
[a] := \{ k \in \nset \mid 1 \leq k \leq a \}.
\]
Additionally, for \( (a, b) \in \nset \times \bar{\nset} \), where $\bar{\nset} = \nset \cup \{ + \infty\}$, such that \( a \leq b \), define the set \( [\![a,b]\!] \) as the set of all natural numbers between \( a \) and \( b \), inclusive:
\[
[\![a,b]\!] := \{ k \in \mathbb{N} \mid a \leq k \leq b \}.
\]

%% file: 2025-AISTATS/appendix/algo.tex
\section{PSEUDO CODE}

For clarity of exposition, we provide the complete pseudo-code of the server-side and client-side algorithms in Algorithm \ref{algo:server-UCBVI} and Algorithm \ref{algo:UCB-Q-values}.

\begin{algorithm}[t]
\caption{\texttt{Fed-UCBVI (Central Server)}}
\label{algo:server-UCBVI}
\textbf{Initialize:} $t = 1$, $r = 1$, $N_{(1),h}(s,a) = 0$, $\eqfunc[(1),h](s,a) = \evaluefunc[(1),h](s) = \sepisode$ for all $(s,a,h) \in \S \times \A \times [\sepisode]$, $\apolicy[(1)] = \{ \apolicy[(1),h] \}_{1 \leq h \leq \sepisode}$ an arbitrary deterministic policy

\While{$t \leq \nepisode$}{
    Broadcast $\apolicy[(r)] = \{ \apolicy[(r),h]\}_{1 \leq h \leq \sepisode}$, $\{N_{(r),h} \}_{1 \leq h \leq \sepisode}$, $r$, and $t$ to all clients;
    
    Wait until receiving the synchronization signal and an updated episode number $t$ and forward the abortion signal to all clients;

    Set $\evaluefunc[(r+1),\sepisode + 1](s) = 0$ for all $s \in \S$ and send it to all clients;
    
    \For{$h = \sepisode$ \KwTo $1$}{
        Receive $\{ \eqfunc[(r+1),h][i] \}_{i}$, $\{ n_{(r+1),h}^{i} \}_{i}$,$\{ \ekerMDP[(r+1),h][i]\evaluefunc[(r+1),h+1](s,a) \}_{s,a,i}$, and $\{ \ekerMDP[(r+1),h][i]\evaluefunc[(r+1),h+1]^{\, 2}(s,a)\}_{s,a,i}$ from the different clients;
        
        \For{$(s,a) \in \S \times \A$}{
            Compute $N = \counter_{(r+1),h}(s,a) = \sum_{i=1}^\nagent n_{(r+1),h}^{i}(s,a)$;

            Set $n^i = n_{(r+1),h}^{i}(s,a) $ for $i \in [\nagent]$;
            \begin{equation*}
            \begin{aligned}
            \! \!\! \! \!\! \! \!\! \! \!\!\! \!\!\! \!\!\! \!\!\! \!\!\! \!\!\! \!\!\! \!\!\! \!\!\! \!\!\! \!\!\! \!\!\! \!\!\! \!\!\! \!\!\! \!\!\! \!\!\! \!\!\! \!\!\! \! 
            \text{Set } V:=  \empvar[(r)]{\expandafter{\evaluefunc[(r),h+1]}}(s,a) &= \frac{1}{N}\sum_{i=1}^\nagent n^i \ekerMDP[(r+1),h][i]\evaluefunc[(r+1),h+1]^{\, 2}(s,a)  \\
            \textstyle
            &-   \left(\frac{1}{N}\sum_{i=1}^\nagent n^i \ekerMDP[(r+1),h][i]\evaluefunc[(r+1),h+1](s,a) \right)^2     
            \end{aligned}
            \end{equation*}
            
            \begin{equation}
            \textstyle \nonumber
            \label{eq:bonus}            \!\!\!\!\!\!\!\!\!\!\!\!\!\!\!\!\!\!\!\!\!\!\!
            \!\!\!\!\!\!\!\!\!\!\!\!\!\!\!\!\!\!\!\!\!\!\!\!\!\!\!\!\!\!\!\!\!\!\!\!\!\!\!\!\!\!\!\!\!\!\!\!\!\!
            \text{Compute }\bonus_{(r),h}(s,a) 
            = \begin{cases}
            \frac{28\beta^\star(\delta)\sepisode + 11\beta^{\common}(\delta, N)}{N} + \sqrt{ \frac{8 \beta^\star(\delta)}{N}\cdot V}\eqsp, & N \geq 2 \,, \\ 
                H\,, & N  \leq 1\,;
            \end{cases}
            \end{equation}
            \begin{equation}
            \textstyle \nonumber
            \label{eq:update_rule}
            \!\!\!\!\!\!\!\!\!\!\!\!\!\!\!\!\!\!\!\!\!\!\!\!\!\!
            \!\!\!\!\!\!\!
            \text{Set }\eqfunc[(r+1),h](s,a) =
            \begin{cases}
                \min \big(\sum_{i=1}^\nagent \frac{n^i}{N} \eqfunc[(r+1),h][i](s,a) + \bonus_{(r+1),h}(s,a), \sepisode\big) & \text{if }N> 0\,,\\
                \sepisode & \text{otherwise};
            \end{cases}
            \end{equation}
        }
        \For{$s \in \S$}{
            Compute $\evaluefunc[(r+1),h](s) = \max_{a \in \A} \eqfunc[(r+1),h](s,a)$;
            
            Compute $\apolicy[(r+1),h](s) = \argmax_{a \in \A} \eqfunc[(r+1),h](s,a)$;
        }
        Broadcast $\evaluefunc[(r+1),h]$ to all clients;
    }
    Set $r = r + 1$;
}
Send a signal to inform the clients of the end of training.
\end{algorithm}

\begin{algorithm}[t]
\caption{\texttt{Fed-UCBVI (i-th Client Side)}}
\label{algo:UCB-Q-values}
\textbf{Initialize:}  $n_{(1,0),h}^{i}(s,a) = 0$, $\erewardMDP[h]^{\eqsp i}(s,a) = 0$, $n_{(1,0),h}^{i}(s,a,s') = 0$ for all $(s,a,s',h) \in \S \times \A \times \S \times [\sepisode]$;

Compute $\nu(\delta, \nepisode) =  14 \hgkernel \nepisode \sepisode \nagent + 182 \nagent \beta^{\common}(\delta, \nepisode) $;

\While{signal of end of training not received}{
    Receive $\{\apolicy[(r),h]\}_{1 \leq h \leq \sepisode}$, $(N_{(r),h})_{1 \leq h \leq \sepisode}$, $r$, and $t$ from the central server;
    
    Set $\ell = 1$;
    
    Set $\hat{N}^{\,i}_{(r,0),h}(s,a) = N_{(r),h}(s,a)$ for all $(s,a,h) \in \S \times \A  \times [\sepisode]$;
    
    Set $n_{(r,0),h}^{i}(s,a) = n_{(r),h}^{i}(s,a)$ for all $(s,a,s',h) \in \S \times \A \times \S \times [\sepisode]$;
    
    \While{no synchronization signal from central server and $t \leq \nepisode$}{
        synchronize = \texttt{False};
        
        \While{synchronize = \texttt{False}}{
            Collect a new trajectory $(s_{t,h}^{i}, a_{t,h}^{i}, \r_{t,h}^{i})_{1 \leq h \leq \sepisode}$ using the policy $\apolicy[(r)]$;
            
            \For{$h = 1$ \KwTo $\sepisode$}{
                Set $\erewardMDP[h]^{\eqsp i}(s_{t,h}^{i},a_{t,h}^{i}) = \r_{t,h}^{i}$;
                
                Set $n_{(r,\ell),h}^{i}(s,a) = n_{(r,\ell-1),h}^{i}(s,a)+ \Ind_{(s,a)}(s_{t,h}^i, a_{t,h}^i)$ for $(s,a) \in \S \times \A$ ;
                
                $n_{(r,\ell),h}^{i}(s,a,s') = n_{(r,\ell-1),h}^{i}(s,a,s') + \Ind_{(s,a,s')}(s_{t,h}^i, a_{t,h}^i, s_{t,h+1}^i)$ for $(s,a,s') \in \S \times \A\times \S$ ;

                Set $\hat{N}^{\,i}_{(r,\ell),h}(s,a) = \hat{N}^{\,i}_{(r,\ell -1),h}(s,a) + \nagent \Ind_{(s,a)}(s_{t,h}^i, a_{t,h}^i)$ for $(s,a) \in \S \times \A$;
                
                \If{ $N_{(r),h}(s_{t,h}^{i},a_{t,h}^{i}) < \nu(\delta, \nepisode)$ and $n_{(r,l),h}^{i}(s_{t,h}^{i},a_{t,h}^{i}) \ge 2 n_{(r),h}^{i}(s_{t,h}^{i},a_{t,h}^{i})  $}{
                synchronize = \texttt{True};
                }
                
                \ElseIf{ $N_{(r),h}(s_{t,h}^{i},a_{t,h}^{i}) \ge \nu(\delta, \nepisode)$ and $\hat{N}^{\,i}_{(r,l),h}(s_{t,h}^{i},a_{t,h}^{i}) \ge 2 N_{(r),h}(s_{t,h}^{i},a_{t,h}^{i})  $}{
                synchronize = \texttt{True};
                }
            }
            Set $\ell = \ell + 1$ and $t = t + 1$\;;
        }
        Send an abortion signal and an episode number $t$ to the central server;
    }
    Set $n_{(r+1),h}^{i}(s,a) = n_{(r,\ell),h}^{i}(s,a)$ and $n_{(r+1),h}^{i}(s,a,s') = n_{(r,\ell),h}^{i}(s,a,s')$ for all $(s,a,s') \in \S \times \A \times \S$;
    
    Set $\ekerMDP[(r+1),h][i](s'|s,a) = \begin{cases}
    \frac{n_{(r+1),h}^{i}(s,a,s')}{n_{(r+1),h}^{i}(s,a)} & \text{if }\,, n_{(r+1),h}^{i}(s,a) > 0\\
    \frac{1}{\nstates} & \text{else};
    \end{cases}$
    
    \For{$h = \sepisode$ \KwTo $1$}{
        Receive $\evaluefunc[(r+1),h+1]$ from the central server;
        
        Compute $\eqfunc[(r+1),h][i](s,a) = \erewardMDP[h]^{i}(s,a) + \ekerMDP[(r+1),h][i] \evaluefunc[(r+1),h+1](s,a)$ for all $(s,a,s') \in \S\times \A \times \S$;
        
        Send $\eqfunc[(r+1),h][i]$,  $n_{(r+1),h}^{i}$,$\{ \ekerMDP[(r+1),h][i]\evaluefunc[(r+1),h+1](s,a) \}_{s,a}$, and $\{ \ekerMDP[(r+1),h][i]\evaluefunc[(r+1),h+1]^{\, 2}(s,a)\}_{s,a}$ to the central server.
    }
}
\end{algorithm}

%% file: 2025-AISTATS/appendix/concentration.tex
\section{CONCENTRATION EVENTS}

Before we proceed, let us define several essential quantities. 

\paragraph{Change of epoch notation}
We notice that the set of all regular episodes $t \in [T]$ is separated into a sequence of different \textit{random} epochs $\mathrm{E}_{1},\mathrm{E}_{2},\ldots$. To define them properly, let us define the epoch-changing timestamps as follows
\begin{equation}\label{eq:def_epoch_changing_timestamps}
    T_1 := 0, \quad T_{r+1} := \min\{t > T_{r} \mid \mathrm{Sync}_{r}(t) = \mathtt{True}\}.
\end{equation}
where the epoch-switching predicate is defined as
\begin{equation}\label{eq:sync_condition}
    \mathrm{Sync}_{r}(t) = \begin{cases}
        \exists i \in [\nagent]: n^i_{(r,\ell),h}(s^i_{t,h}, a^{i}_{t,h}) \geq 2 n^i_{(r),h}(s^i_{t,h}, a^{i}_{t,h}) & \text{if }N_{(r),h}(s_{t,h}^{i},a_{t,h}^{i}) < \nu(\delta, \nepisode) \\
        \exists i\in [\nagent]: \hat{N}^{\,i}_{(r,\ell),h}(s_{t,h}^{i},a_{t,h}^{i}) \ge 2 N_{(r),h}(s_{t,h}^{i},a_{t,h}^{i})  & \text{if } N_{(r),h}(s_{t,h}^{i},a_{t,h}^{i}) \geq \nu(\delta, \nepisode)
    \end{cases}\eqsp,
\end{equation}
for $\ell = t - T_{r} - 1$ and $\nu(\delta, \nepisode)$ is defined in \eqref{def:sdeltat}. In particular, this condition exactly corresponds to the synchronization condition used by \algo. Then, the epoch $\mathrm{E}_{r}$ is defined as $\mathrm{E}_{r} := [\![T_{r}+1; T_{r+1}]\!]$. In particular, for any $t \in [T]$, we define $r_t$ as a unique index $r$ such that $t \in \mathrm{E}_r$:
\begin{equation}\label{eq:def_rt}
    r_t = \min\{ r \geq 1 \mid t > T_r \}.
\end{equation}

\paragraph{Definitions} First of all, let us recall that by \Cref{assum:lhg1} the transition kernel $\kerMDP[h][i]$ for the agent $i$ is a mixture of common kernel $\comkerMDP[h]$ and individual kernel $\perkerMDP[h][i]$, thus any sample $s_{t,h+1}^i \sim \kerMDP[h][i](s^i_{t,h},a^i_{t,h})$ for $(t,h,i) \in [T] \times [\sepisode] \times [\nagent]$ can be represented via the following experiment
\begin{equation}\label{eq:mixture_state_generation}
    s^i_{t,h+1}  =\begin{cases}
        s^{\common,i}_{t,h+1} \sim \comkerMDP[h](s^i_{t,h},a^i_{t,h})\,, & \xi^i_{t,h} = 0\,, \\
        s^{\indiv, i}_{t,h+1} \sim \perkerMDP[h][i](s^i_{t,h},a^i_{t,h})\,, & \xi^i_{t,h} = 1\,,
    \end{cases}
\end{equation}
where $\xi^i_{t,h} \sim \Bern(\hgkernel)$ is a choice of component of the mixture. Using this representation, we can define a \textit{virtual estimate of the common kernel} for a step $t$ as follows
\begin{equation}\label{eq:virtial_estimate_common_kernel}
    \ecomkerMDP[(r),h](s'|s,a) := \frac{1}{\counter_{(r),h}(s,a)} \sum_{i=1}^\nagent \sum_{t=1}^{T_{r}} \Ind_{(s,a,s')}(s^i_{t,h},a^i_{t,h},s^{\common,i}_{t,h+1})\eqsp,
\end{equation}
where $T_r$ is defined in \eqref{eq:def_epoch_changing_timestamps}.
We emphasize that $\ecomkerMDP[(r),h]$ is never computed explicitly by the algorithm since the values of $\xi^i_{t,h}$ are never observed, however we are very interested in the analysis of it. 

Additionally, let us define the weighted average kernel
\begin{equation}\label{eq:weighted_avg_emp_kernel}
    \ekerMDP[(r),h](s'|s,a) := \sum_{i=1}^N \frac{n^i_{(r),h}(s,a)}{\counter_{(r),h}(s,a)} \ekerMDP[(r),h][i](s'|s,a) = \frac{\counter_{(r),h}(s,a,s')}{\counter_{(r),h}(s,a)}\,,
\end{equation}
where $N_{(r),h}(s,a) = \sum_{i=1}^\nagent n_{(r),h}^{i}(s,a)$ was defined in \eqref{def:globalcounter}, and
$\ekerMDP[(r),h][i]$ was defined in \eqref{eq:update_of_tr_kernel} as
\begin{align}
\ekerMDP[(r),h][i](s'|s,a) = \begin{cases}
    \frac{n^{i}_{(r),h}(s,a,s')}{n^{i}_{(r),h}(s,a)} & \text{if } n^{i}_{(r),h}(s,a) > 0\\
    \frac{1}{\nstates} & \text{else}
    \end{cases}
    \eqsp.
\end{align}
Notably, the kernel $\ekerMDP[(r),h]$ is never revealed to any agent or to a central server, but it is very useful in the analysis. Also, for any time $t$ we define $r_t$ as an index of the previous epoch. 
For convenience and ease of reading, we introduce the transition kernels and counters in the \textit{regular timescale}
\begin{align}
\label{def:changing_timescale}
\ekerMDP[t,h][i] := \ekerMDP[(r_t),h][i] \eqsp, \quad  \ekerMDP[t,h] := \ekerMDP[(r_t),h] \eqsp \quad n_{t,h}^{i} = n_{(r_t),h}^{i} \eqsp, \text{ and } N_{t,h}^{i} = N_{(r_t),h}^{i} \eqsp,
\end{align}
where $r_t$ is defined in \eqref{eq:def_rt}.

Let $\beta^{\KL}, \beta^{\common}, \beta^{\Var} \colon (0,1) \times \nset \to \rset_{+}$ and $ \beta^\star,  \beta, \beta^{\text{max}} \colon (0,1) \to \rset_{+}$ be some functions defined later on in Lemma \ref{lem:proba_master_event}, and $\troundsmax$ be the maximal number of communications defined \eqref{def:roundsmax}. We define the following favorable events
\begin{align*}
\cE^{\KL}(\delta) &:= \Bigg\{ \forall r \in \nset, \forall h \in [H], \forall (s,a) \in \cS\times\cA: \quad \KL\left(\ecomkerMDP[(r),h](s,a) \Big\Vert \comkerMDP[h](s,a)\right) \leq \frac{ \nstates\beta^{\KL}(\delta, N_{(r),h}(s,a))}{N_{(r),h}(s,a)} \Bigg\}\,,\\
\cE^{\common}(\delta) & := \Bigg\{ \forall r \in [\troundsmax], \forall h \in [H], \forall (s,a) \in \cS\times\cA: \left\Vert \ekerMDP[(r),h](s,a) -  \ecomkerMDP[(r),h](s,a)\right\Vert_1 \leq \frac{9}{8}\hgkernel + \frac{11\beta^{\common}(\delta, \counter_{(r),h}(s,a))}{\counter_{(r),h}(s,a)} \Bigg\}\,,\\
\cE^\star(\delta) &:= \Bigg\{\forall r \in [\troundsmax], \forall h \in [H], \forall (s,a)\in\cS\times\cA: \quad \left| [\ecomkerMDP[(r),h] - \comkerMDP[h]] \comstarvaluefunc[h+1](s,a)   \right| \\
&\qquad \leq \Ind_{[\![2;+\infty]\!]}( \counter_{(r),h}(s,a) )\bigg(\sqrt{\frac{2\comempvar[(r),h]{\expandafter{\comstarvaluefunc[h+1]}}(s,a) \beta^\star(\delta)}{{\counter_{(r),h}(s,a)}-1}} + \frac{7\beta^\star(\delta)}{{\counter_{(r),h}(s,a)}-1} \bigg) + \sepisode \Ind_{[\![0;2]\!]}( \counter_{(r),h}(s,a))  \Bigg\}\,,\end{align*}\begin{align*}
\mathcal{E}^{\operatorname{Var}}(\delta) &:= \left\{\forall t \in [\nepisode]: \quad \sum_{(t'\geq1,h\geq 1,i \geq 1)}^{(t,\sepisode,\nagent)} \agentvar[h]{i}{\valuefunc[h+1][i,\pi_{t'}]}\left(s^i_{t',h}, a^i_{t',h}\right) \leq \sqrt{2\sepisode^5 \nagent t \beta^{\Var}(\delta, t)}
+ 3\sepisode^3  \beta^{\Var}(\delta, t)
+ \sepisode^2 \nagent t\right\} \,,\\
\mathcal{E}^{\operatorname{count}}(\delta) &:= \bigg\{\forall t \in [\nepisode], \forall h \in [\sepisode], \forall (s,a) \in \S \times \A, \forall i \in [\nagent]: \quad |\parcounter^{\,\nagent}_{t,h}(s,a) - \hat{N}^{\,i}_{t,h}(s,a)|\\
&\qquad\qquad\qquad\qquad \qquad\qquad\qquad \qquad\qquad \qquad\qquad \quad\leq  \frac{2}{7} \hat{N}^{\,i}_{t,h}(s,a) + 2 \hgkernel \nepisode \sepisode \nagent +26 \nagent \beta^{\common}(\delta, \nepisode)\bigg\} \,,\\
\cE(\delta) & := \Bigg\{ \forall h \in [\sepisode] : \sum_{(t\geq1,h'\geq h,i \geq 1)}^{(T,\sepisode,\nagent)} 
 \gamma_{h'-1} \left( \kerMDP[h'][i]\left[ \evaluefunc[t,h'+1] -  \comvaluefunc[h'+1][\pi_t]\right](s^i_{t,h'}, a^i_{t,h'}) -  \left[\evaluefunc[t,h'+1] -  \comvaluefunc[h'+1][\pi_t]\right](s^i_{t,h'+1})\right) \\
&\qquad\qquad\qquad\qquad\leq \sqrt{8\rme^2 H^2 \cdot T\sepisode\nagent \cdot \beta(\delta) }, \quad \gamma_h := \left(1 + \frac{1}{H}\right)^{H-h}\eqsp, \text{ and } \\
 & \sum_{(t\geq1,h'\geq h,i \geq 1)}^{(T,\sepisode,\nagent)} 
\left( \kerMDP[h'][i]\left[ \evaluefunc[t,h'+1] -  \comvaluefunc[h'+1][\pi_t]\right](s^i_{t,h'}, a^i_{t,h'}) -  \left[\evaluefunc[t,h'+1] -  \comvaluefunc[h'+1][\pi_t]\right](s^i_{t,h'+1})\right)\leq \sqrt{8 H^2 \cdot T\sepisode\nagent \cdot \beta(\delta) } \Bigg\}\,.
\end{align*}
We also introduce the intersection of these events, $\cG(\delta) := \cE^{\KL}(\delta) \cap \cE^{\common}(\delta) \cap \cE^\star(\delta) \cap \mathcal{E}^{\operatorname{Var}}(\delta) \cap \mathcal{E}^{\operatorname{count}}(\delta) \cap \cE(\delta)$. We prove that for the right choice of the functions $\beta^{\KL}, \beta^{\common}, \beta^\star, \beta^{\Var} $, and $ \beta $ the above events hold with high probability.
\begin{lemma}
\label{lem:proba_master_event}
For any $\delta \in (0,1)$ and for the following choices of functions $\beta,$
\begin{align*}
    \beta^{\KL}(\delta, n) &:= \log(6\nstates\nactions\sepisode/\delta) + \log\left(\rme(1+n) \right)\,,
    &
    \beta^{\common}(\delta, n) &:= \log(6\nstates\nactions\sepisode/\delta) + \log\left(6\rme (2n+1)\right) \,,\\ 
    \beta^\star(\delta) &:= \log(12\nstates\nactions\sepisode/\delta)\,, & \beta^{\Var}(\delta, t) &:= \log \left(24\rme(2 \nagent t+1)/\delta\right)\, , \\
     \beta(\delta) &:= \log(48\sepisode/\delta)\,. & &    
\end{align*}
it holds that
\begin{align*}
\P[\cE^{\KL}(\delta)]\geq 1-\delta/6, &\qquad \P[\cE^{\common}(\delta)] \geq 1-\delta/6, \qquad \P[\cE^\star(\delta)]\geq 1-\delta/6, \qquad \P[\mathcal{E}^{\operatorname{Var}}(\delta)] \geq  1-\delta/6 ,\\
&\ 
\P[\mathcal{E}^{\operatorname{count}}(\delta)] \geq 1-\delta/6 ,\qquad \P[\cE(\delta)]\geq 1-\delta/6 \,.
\end{align*}
In particular, $\P[\cG(\delta)] \geq 1-\delta$.
\end{lemma}
\begin{proof}

First, let us define an appropriate filtration for martingale and optional skipping-based arguments.
A natural federated online filtration is defined as
\begin{equation}\label{eq:online_filtration}
    \cF_{t,h}^i = \sigma\left(\{ s^{i'}_{t',h'} a^{i'}_{t',h'} \}_{(t',h',i')\preceq(t,h,i)}\right)\,, 
\end{equation}
where the order over triplets $(t',h',i')$ is lexicographic. With respect to this filtration, for any fixed state-action-step triplet $(s,a,h) \in \S \times \A \times [\sepisode]$ we define the partial global counters that form a sequence of excursion times on an extended time $(t,i) \in \nset \times [\nagent]$ and a first extended timestamp to reach a particular partial counter value $j \in [T\cdot \nagent]$
\begin{equation}\label{eq:partial_counters_definition}
     \parcounter^{\,i}_{t,h}(s,a) = \sum_{(t', i') \preceq (t,i)} \Ind_{(s,a)}(s^i_{t,h},a^i_{t,h})\,,\quad (t_{s,a,h,j}, i_{s,a,h,j}) := \min\{ (t,i) \in \nset \times [\nagent] \mid \parcounter^{\,i}_{t,h}(s,a) = j \}\,.
\end{equation}
For a given time $t$, we also define $\psi_t := T_{r_t}$ representing the number of episodes visited before $r_t$. In particular, we have $\counter_{(r),h}(s,a) = \parcounter^{\,\nagent}_{T_r,h}(s,a)$.
\paragraph{Event $\cE^{\KL}(\delta)$}
To analyze it, we need first to represent the virtual estimate of the common transition kernel as follows
\begin{align}
    \ecomkerMDP[(r),h](s'|s,a) &= \frac{1}{\counter_{(r),h}(s,a)} \sum_{t=1}^{T_r} \sum_{i=1}^{\nagent} \Ind_{(s,a)}(s^i_{t,h}, a^i_{t,h}) \Ind_{s'}( s^{\common,i}_{t,h+1})\notag \\
    &= \frac{1}{\counter_{(r),h}(s,a)}  \sum_{j=1}^{\counter_{(r),h}(s,a)} \Ind_{s'}( s^{\common,i_{s,a,h,j}}_{t_{s,a,h,j},h+1} )\,. \label{eq:estimate_common_kernel_optional_skipping}
\end{align}
By the optional skipping argument (see, \eg, \citealp[Chapter III, p. 145]{doob1953stochastic}), the sampled states $\{ \tilde{s}^{\,\common}_{s,a,h,j}\}_{j\in[T \nagent]} := \{ s^{\common,i_{s,a,h,j}}_{t_{s,a,h,j},h+1} \}_{j \in [T\nagent]}$, conditioned on the value of $\counter_{(r),h}(s,a)$, form an i.i.d. sequence of categorical random variables from the distribution $\comkerMDP[h](s,a)$. Thus, we have for any fixed $(s,a,h) \in \S \times \A \times [\sepisode]$ by \Cref{lem:kl_deviation_inequality}
\[
    \P\left[ \exists r \geq 1:  \KL\left(\ecomkerMDP[(r),h](s,a) \Big\Vert \comkerMDP[h](s,a)\right) \leq \frac{\log(6\nstates\nactions\sepisode/\delta) + \nstates\log\left(\rme(1+n) \right)}{N_{(r),h}(s,a)}  \right] \leq \frac{\delta}{6\nstates \nactions \sepisode}\,.
\]
By a union bound argument and noticing that $ \log(6\nstates\nactions\sepisode/\delta) + \nstates\log\left(\rme(1+n) \right) \leq \nstates \beta^{KL}(\delta, n)$, we conclude the first statement.

\paragraph{Event $\cE^{\common}(\delta)$}  By a union bound argument, it is enough to show that each of the following events 
\[
    \overline{\cE}^{\,\common}(\delta, s,a,h) = \left\{ \exists r \in [\troundsmax]: \left\Vert \ekerMDP[(r),h](\cdot | s,a) -  \ecomkerMDP[(r),h](\cdot | s,a)\right\Vert_1 \ge \frac{9}{8}\hgkernel + \frac{11\beta^{\common}(\delta, \counter_{(r),h}(s,a))}{\counter_{(r),h}(s,a)}\right\}
\]
holds with probability less or equal than $\delta' := \delta/(6\nstates \nactions \sepisode)$ for any $(s,a,h) \in \S \times \A \times [\sepisode]$. To do it, let us analyze the difference between kernels.   By the definitions \eqref{eq:virtial_estimate_common_kernel}-\eqref{eq:weighted_avg_emp_kernel}:
\[
    [\ekerMDP[(r),h] -  \ecomkerMDP[(r),h]](s'|s,a) = \frac{1}{\counter_{(r),h}(s,a)} \sum_{i=1}^\nagent \sum_{t=1}^{T_r} \left( \Ind_{(s,a,s')}(s^i_{t,h},a^i_{t,h},s^{i}_{t,h}) - \Ind_{(s,a,s')}(s^i_{t,h},a^i_{t,h},s^{\common,i}_{t,h}) \right)\,.
\]
Next, we notice that, using representation \eqref{eq:mixture_state_generation}, we can rewrite the first indicator as follows
\[
    \Ind_{(s,a,s')}(s^i_{t,h},a^i_{t,h},s^{i}_{t,h}) = (1 - \xi^i_{t,h}) \cdot \Ind_{ (s,a,s')}(s^i_{t,h},a^i_{t,h},s^{\common, i}_{t,h})  + \xi^i_{t,h} \Ind_{(s,a,s')}(s^i_{t,h},a^i_{t,h},s^{\indiv, i}_{t,h})\,,
\]
thus, we have the following expression for the difference between kernels
\[
    [\ekerMDP[(r),h] -  \ecomkerMDP[(r),h]](s'|s,a) = \frac{1}{N_{(r),h}(s,a)} \sum_{i=1}^\nagent \sum_{t=1}^{T_r} \xi^i_{t,h} \Ind_{(s,a)}(s^i_{t,h},a^i_{t,h})\cdot \left( \Ind_{s'}(s^{\indiv,i}_{t,h}) - \Ind_{s'}( s^{\common,i}_{t,h}) \right) ,
\]
and thus, using $\abs{ \Ind_{ s'}(s^{\indiv,i}_{t,h}) - \Ind_{s'}( s^{\common,i}_{t,h})  } \le 1$ and Definition \eqref{eq:partial_counters_definition}, we obtain
\begin{align}\label{eq:emp_heterogeneity_bound}
    \left\Vert [\ekerMDP[(r),h] -  \ecomkerMDP[(r),h]](s,a) \right\Vert_1 
    & \leq \frac{1}{N_{(r),h}(s,a)} \sum_{t=1}^{T_r} \sum_{i=1}^\nagent \xi^i_{t,h} \Ind_{(s,a)}(s^i_{t,h},a^i_{t,h})
     = \frac{1}{N_{(r),h}(s,a)}\sum_{j=1}^{N_{(r),h}(s,a)} \xi^{i_{s,a,h,j}}_{t_{s,a,h,j}}\,.
\end{align}
Again, by the optional skipping argument, conditioned on the event $N_{(r),h}(s,a) = N$ the sequence $\{\tilde{\xi}_{s,a,h,j} \}_{j \in [T\nagent]} := \{ \xi^{i_{s,a,h,j}}_{t_{s,a,h,j}} \}_{j \in [T\nagent]}$ is i.i.d., thus \Cref{cor:bernoulli-deviation} implies
\[
    \P[\overline{\cE}^{\,\common}(\delta, s,a,h)] \leq \P\left[\exists N \geq 1: \sum_{j=1}^N \tilde{\xi}_{s,a,h,j}  > \frac{9}{8}N \hgkernel + 11\log\left(\frac{24\nstates\nactions\sepisode \rme (2N+1)}{\delta}\right)\right] \leq \frac{\delta}{6\nstates\nactions\sepisode}\,.
\]

\paragraph{Event $\cE^\star(\delta)$}

To analyze this event, we use the representation of the kernel \eqref{eq:estimate_common_kernel_optional_skipping} and optional skipping argument conditioned on $\counter_{(r),h}(s,a) = N$ where $N \geq 2$
\[
     [\ecomkerMDP[(r),h] - \comkerMDP[h]] \comstarvaluefunc[h+1](s,a) = \frac{1}{N}\sum_{j=1}^N \comstarvaluefunc[h+1](s^{\common}_{s,a,h,j}) -  \comkerMDP[h] \comstarvaluefunc[h+1](s,a)\,.
\]
Thus, we have a sum of centered i.i.d. random variables, and thus we can apply \Cref{lem:empiricalbernstein}
\[
    \P\left[ \left| [\ecomkerMDP[(r),h] - \comkerMDP[h]] \comstarvaluefunc[h+1](s,a) \right| \geq \sqrt{\frac{2\comempvar[(r),h]{\expandafter{\comstarvaluefunc[h+1]}}(s,a) \beta^\star(\delta)}{N-1}} + \frac{7\beta^\star(\delta)}{N-1}  \Bigg| \counter_{(r),h}(s,a) = N \right] \leq \frac{\delta}{6\nstates\nactions\sepisode\nepisode},
\]
where $\beta^\star(\delta) = \log(12\nstates\nactions\sepisode/\delta)$.
We then conclude by a union bound over $(s,a,h,N) \in \S \times \A \times [\sepisode] \times \{2,\ldots,\nagent T\}$. If $\counter_{(r),h}(s,a) \le 1$, we have the trivial bound $ \left| [\ecomkerMDP[(r),h] - \comkerMDP[h]] \comstarvaluefunc[h+1](s,a) \right| \leq \sepisode$.

\paragraph{Event $\mathcal{E}^{\operatorname{Var}}(\delta)$} For any $t' \in [\nepisode]$, define
\begin{align*}
X_{t'}^i = \sum_{h=1}^\sepisode  \agentvar[h]{i}{\valuefunc[h+1][i,\pi_{t'}]}\left(s^i_{t',h}, a^i_{t',h}\right) - \sigma \valuefunc[1][i,\policy_{t'}]\left(s^i_{t',1}\right) \eqsp,
\end{align*}
where $\sigma \valuefunc[1][i,\policy_{t'}]$ is defined in \eqref{eq:definition_variance_bellman}. This sequence forms a martingale-difference sequence with respect to the following filtration
\begin{align*}
    \cF_{t}^i = \sigma\left(\{ s^{i'}_{t',h} a^{i'}_{t',h} \}_{(t',i')\preceq(t,i), h\in [\sepisode]}\right)\,, 
\end{align*}
where the order over the pairs $(t',i')$ is lexicographic. Applying  \Cref{th:bernstein} yields
\begin{align*}
\P\left[\exists t\geq 1, \sum_{(t'\geq1, i \geq 1)}^{(t,\nagent)} X_{t'}^i \leq \sqrt{2 \sum_{(t'\geq1, i \geq 1)}^{(t,\nagent)}\mathbb{E}_\pi[(X_{t'}^i)^2 \mid \cF_{t'_{\mathrm{prev}}}^{i_{\mathrm{prev}}}]\log\left(24\rme(2 \nagent t+1)/\delta\right)}+ 3\sepisode^3\log\left(24\rme(2\nagent t+1)/\delta\right) \right]\leq \frac{\delta}{6} \eqsp,
\end{align*}
as we have $|X_{t'}^i| \leq \sepisode^3 $ and where $(t'_{\mathrm{prev}},i_{\mathrm{prev}})$ is a previous element in a lexicographic order with respect to $(t',i)$. Now, we bound the conditional second-order moment of $X_{t'}^i$ as follows
\begin{align*}
\mathbb{E}_{\policy_{t'}}[(X_{t'}^i)^2 \mid \cF_{t'_{\mathrm{prev}}}^{i_{\mathrm{prev}}}] 
\leq 
\mathbb{E}_{\policy_{t'}}\left[\left(\sum_{h=1}^\sepisode  \agentvar[h]{i}{\valuefunc[h+1][i,\pi_{t'}]}\left(s^i_{t',h}, a^i_{t',h}\right) \right)^2 \middle\vert  \cF_{t}^{i-1}\right] 
\leq
\sepisode^3 \mathbb{E}_{\policy_{t'}}\left[\sum_{h=1}^\sepisode  \agentvar[h]{i}{\valuefunc[h+1][i,\pi_{t'}]}\left(s^i_{t',h}, a^i_{t',h}\right)\right] \eqsp .
\end{align*}
By \Cref{lem:bellman_variance}, we have
\begin{align*}
\mathbb{E}_{\policy_{t'}}\left[\sum_{h=1}^\sepisode  \agentvar[h]{i}{\valuefunc[h+1][i,\pi_{t'}]}\left(s^i_{t',h}, a^i_{t',h}\right)\right] = 
\mathbb{E}_{\policy_{t'}}\left[\left(\sum_{h=1}^\sepisode \rewardMDP[h][i] (s^i_{h}, a^i_{h})-  \valuefunc[1][i,\policy_{t'}](s_1^i)\right)^2 \right]
\leq
\mathbb{E}_{\policy_{t'}}\left[\left(\sum_{h=1}^\sepisode \rewardMDP[h][i] (s^i_{h}, a^i_{h})\right)^2 \right]
\leq \sepisode^2 \eqsp.
\end{align*}
By combining the previous inequalities, we obtain
\begin{align*}
\sum_{(t'\geq1, i \geq 1)}^{(t,\nagent)}  X_{t'}^i \leq \sqrt{2\sepisode^5 \nagent t \log \left(24\rme(2 \nagent t+1)/\delta\right)} + 3\sepisode^3  \log \left(24\rme(2 \nagent t+1)/\delta\right)
\end{align*}
Now using \Cref{lem:bellman_variance} again we get
\begin{align*}
\sum_{(t'\geq1, i \geq 1)}^{(t,\nagent)} \sum_{h=1}^\sepisode  \agentvar[h]{i}{\valuefunc[h+1][i,\pi_{t'}]}\left(s^i_{t',h}, a^i_{t',h}\right)  
&= \sum_{(t'\geq1, i \geq 1)}^{(t,\nagent)} X_{t'}^i +  \sigma \valuefunc[1][i,\policy_{t'}]\left(s^i_{t',1}\right)  
\\
&\leq
\sqrt{2\sepisode^5 \nagent t \log \left(24\rme(2 \nagent t+1)/\delta\right)}
+ 3\sepisode^3  \log \left(24\rme(2 \nagent t+1)/\delta\right)
+ \sepisode^2 \nagent t \eqsp.
\end{align*}
\paragraph{Event $\mathcal{E}^{\operatorname{count}}(\delta)$} For any fixed  $(s,a,h,i,t_1) \in \S \times \A \times [\sepisode] \times [\nagent] \times [\nepisode]  $, we have by \Cref{cor:bernoulli-deviation}
\begin{align*}    
\P\big[\exists t_2 \in \nset : \eqsp &\left| \sum_{t'=t_1}^{t_2} \Ind_{ (s,a)}(s^i_{t',h},a^i_{t',h}) -  \statdist[h]{i, \policy_{t'}}(s, a) \right| \\
&\ge \frac{1}{8} \sum_{t'=t_1}^{t_2} \statdist[h]{i, \policy_{t'}}(s, a) + 11\beta^{\common}(\delta, t_2 - t_1 +1)  \big] \leq \frac{\delta}{6\nstates \nactions \nagent \nepisode \sepisode}\,,
\end{align*}
holds with probability less or equal than $\delta':= \delta/(6\nstates \nactions \nagent \nepisode\sepisode)$. Thus, by a union bound argument, the following event 
\begin{align*}
\overline{\mathcal{E}}^{\,\operatorname{dev}}(\delta) := \bigg\{\forall (t_1, t_2) \in [\nepisode]^2, \forall h \in [\sepisode], \forall (s,a) \in \S \times \A, \forall i \in [\nagent]: \quad &\left| \sum_{t'=t_1}^{t_2} \Ind_{(s,a)}(s^i_{t',h},a^i_{t',h}) -  \statdist[h]{i, \policy_{t'}}(s, a) \right| \\
&\ge \frac{1}{8} \sum_{t'=t_1}^{t_2} \statdist[h]{i, \policy_{t'}}(s, a) + 11\beta^{\common}(\delta, t_2 - t_1 +1) \bigg\} \eqsp,
\end{align*}
holds with probability less or equal to $\delta/6$. Now, to conclude the proof, it is enough to show $\mathcal{E}^{\operatorname{count}}(\delta) \subset \mathcal{E}^{\,\operatorname{dev}}(\delta)$.
Let's recall the definition of the estimated counter by agent $i$ 
\begin{align*}
\hat{N}^{\,i}_{t,h}(s,a) = \counter_{(r_t),h}(s,a)  + \nagent \sum_{t'=\psi_t}^{t} \Ind_{(s,a)}(s^i_{t',h},a^i_{t',h}) \eqsp.
\end{align*}
Using \eqref{eq:partial_counters_definition}, the definition of $\hat{N}^{\,i}_{t,h}(s,a)$, and the triangular inequality, we have
 for any fixed $(s,a) \in \S \times \A$,
\begin{align*}
|\parcounter^{\,\nagent}_{t,h}(s,a) - \hat{N}^{\,i}_{t,h}(s,a) | & =  \left| \sum_{j\neq i}\sum_{t'=\psi_t}^{t}   \Ind_{(s,a)}(s^j_{t',h},a^j_{t',h}) - \Ind_{(s,a)}(s^i_{t',h},a^i_{t',h}) \right| \\
&\leq  \underbrace{\left|\sum_{j\neq i}  \sum_{t'=\psi_t}^{t}  \Ind_{(s,a)}(s^j_{t',h},a^j_{t',h}) - \statdist[h]{j, \policy_{t'}} (s,a)\right |}_{\term{1}} 
+  \underbrace{\left|\sum_{j\neq i}\sum_{t'=\psi_t}^{t}   \statdist[h]{j, \policy_{t'}} (s,a) -  \statdist[h]{i, \policy_{t'}} (s,a) \right|}_{\term{2}} \\
&+  \underbrace{(\nagent -1)\left|\sum_{t'=\psi_t}^{t}  \statdist[h]{i, \policy_{t'}} (s,a) - \Ind_{(s,a)}(s^i_{t',h},a^i_{t',h})\right|}_{\term{3}} \eqsp.
\end{align*}
\textbf{Term $\term{2}$: Heterogeneity error}
Using \Cref{lem:diff_stat_dist} combined with the triangular inequality, it holds that
\begin{align*}
\term{2} = \left|\sum_{j\neq i}\sum_{t'=\psi_t}^{t}   \statdist[h]{j, \policy_{t'}} (s,a) -  \statdist[h]{i, \policy_{t'}} (s,a) \right| \leq \hgkernel \sepisode \nagent \nepisode \eqsp.
\end{align*}
\textbf{Terms $\term{1}$ and $\term{3}$: concentration error} On the event $\mathcal{E}^{\,\operatorname{dev}}(\delta)$, we can bound $\term{1}$ as follows
\begin{align*}
\term{1} &\leq \sum_{j\neq i}  \sum_{t'=\psi_t}^{t} \left|  \Ind_{(s,a)}(s^j_{t',h},a^j_{t',h}) - \statdist[h]{j, \policy_{t'}} (s,a) \right| \\
&\leq \sum_{j\neq i} \frac{1}{8} \sum_{t'=\psi_t}^{t} \statdist[h]{j, \policy_{t'}}(s, a) + 11 \nagent\beta^{\common}(\delta, \nepisode) \\
&\leq  \underbrace{\frac{1}{8}\sum_{j\neq i}  \left|\sum_{t'=\psi_t}^{t} \statdist[h]{j, \policy_{t'}}(s, a) - \statdist[h]{i, \policy_{t'}}(s, a)\right|}_{\term{2}} + 11 \nagent\beta^{\common}(\delta, \nepisode)+  \frac{\nagent}{8} \sum_{t'=\psi_t}^{t}  \statdist[h]{i, \policy_{t'}}(s, a) \eqsp.
\end{align*}
Now using the latter bound on $\term{2}$ combined with the inequality $\frac{7}{8} \sum_{t'=\psi_t}^{t}  \statdist[h]{i, \policy_{t'}}(s, a) - 11\beta^{\common}(\delta, \nepisode) \leq  \sum_{t'=\psi_t}^{t} \Ind_{(s,a)}(s^i_{t',h},a^i_{t',h}) $ that follows from $\cE^{\mathrm{dev}}(\delta)$, we get
\begin{align*}
\term{1} \leq \frac{1}{8} \hgkernel \sepisode \nagent \nepisode + \frac{88 \nagent}{7}\beta^{\common}(\delta, \nepisode)+  \frac{1}{7} \hat{N}^{\,i}_{t,h}(s,a) \eqsp.
\end{align*}
We proceed similarly to bound $\term{3}$
\begin{align*}
\term{3} &= (\nagent -1)\left|\sum_{t'=\psi_t}^{t}  \statdist[h]{i, \policy_{t'}} (s,a) - \Ind_{(s,a)}(s^i_{t',h},a^i_{t',h}) \right| 
\\
&\leq   \frac{\nagent}{8} \sum_{t'=\psi_t}^{t} \statdist[h]{j, \policy_{t'}}(s, a) + 11\nagent\beta^{\common}(\delta, \nepisode)  \leq
 \frac{1}{7} \hat{N}^{\,i}_{t,h}(s,a) + \frac{88 \nagent}{7} \beta^{\common}(\delta, \nepisode) \eqsp.
\end{align*}
Finally combining the bounds on $\term{1}, \term{2}$ and $\term{3}$ yields the desired result.
\paragraph{Event $\cE(\delta)$} Notice that the two following sequences
\begin{align*}
X_{t,h}^i &:= \left(1 + \frac{1}{H}\right)^{H-h'-1} \left(\kerMDP[h][i]\left[ \evaluefunc[t,h+1] -  \comvaluefunc[h+1][\pi_t]\right](s^i_{t,h}, a^i_{t,h}) -  \left[\evaluefunc[t,h+1] -  \comvaluefunc[h+1][\pi_t]\right](s^i_{t,h+1})\right) \eqsp, \\
Y_{t,h}^i &:= \kerMDP[h][i]\left[ \evaluefunc[t,h+1] -  \comvaluefunc[h+1][\pi_t]\right](s^i_{t,h}, a^i_{t,h}) -  \left[\evaluefunc[t,h+1] -  \comvaluefunc[h+1][\pi_t]\right](s^i_{t,h+1}) \eqsp,
\end{align*}
forms a martingale-difference sequence with respect to filtration $\cF^i_{t,h}$ defined in \eqref{eq:online_filtration}. Thus, applying Azuma-Hoeffding inequality with a union bound over $h$ and over the two events allows us to conclude the statement.
\end{proof}

\begin{lemma}
\label{lem:bernstein_kl_corr}
Conditioned on $\cE^{\KL}(\delta)$, for any function $ f: \S \mapsto[0, \sepisode], h \in [\sepisode], (s,a) \in \S \times \A$, and any $r \in [\troundsmax]$, we have
\begin{align*}
(\ecomkerMDP[(r),h] - \comkerMDP[h])f(s,a) 
\leq \frac{1}{\sepisode} \comkerMDP[h]f(s,a) 
+ \frac{2 \sepisode^2 \nstates \beta^{\KL}(\delta, N_{(r),h}(s,a))}{N_{(r),h}(s,a)} \eqsp , \\
\Vert \ecomkerMDP[(r),h] - \comkerMDP[h] \Vert_1 \leq \sqrt{\frac{2 \nstates \beta^{\KL}(\delta, N_{(r),h}(s,a))}{N_{(r),h}(s,a)}} \eqsp.
\end{align*}
\end{lemma}
\begin{proof}
    Using \Cref{lem:talebi} with $\mathsf{P} = \ecomkerMDP[(r),h](\cdot | s,a )$ and $\mathsf{Q} = \comkerMDP[h](\cdot|s,a)$ it holds that
    \begin{align}
        \label{eq:bound-diff-phat-p-f}
        \!(\ecomkerMDP[(r),h] - \comkerMDP[h]) f (\cdot | s,a )
        \leq \sqrt{2 \operatorname{Var}_{\comkerMDP[h](\cdot|s,a)}(f) \mathrm{KL}\left(\ecomkerMDP[(r),h](\cdot | s,a ) \big\Vert \comkerMDP[h](\cdot|s,a)\right)}
        \!+\!
        \frac{2}{3} \sepisode \mathrm{KL}\left(\ecomkerMDP[(r),h](\cdot | s,a ) \big\Vert \comkerMDP[h](\cdot|s,a)\right) \eqsp. 
    \end{align}
    Now, since $f$'s values are in $[0, \sepisode]$, we can write
    \begin{align}
        \operatorname{Var}_{\comkerMDP[h](\cdot|s,a)}(f) 
        \le
        \comkerMDP[h] (f^2) (s,a)
        \le
        \sepisode \comkerMDP[h] (f) (s,a)
        \eqsp.
    \end{align}
    Combining the latter inequality with the fact that for all $a, b \ge 0$, $\sqrt{2 a b} \le a + b$, we obtain
    \begin{align}
        \sqrt{2 \operatorname{Var}_{\comkerMDP[h](\cdot|s,a)}(f) \mathrm{KL}\left(\ecomkerMDP[(r),h](\cdot | s,a ) \big\Vert \comkerMDP[h](\cdot|s,a)\right)}
        & =
        \sqrt{\frac{2}{\sepisode} \comkerMDP[h] (f) (s,a) \cdot \sepisode^2 \mathrm{KL}\left(\ecomkerMDP[(r),h](\cdot | s,a ) \big\Vert \comkerMDP[h](\cdot|s,a)\right)}
        \nonumber
        \\
        \label{eq:bound-square-variance-phat-p}
        & \le
        \frac{1}{\sepisode} \comkerMDP[h] (f) (s,a) 
        + \sepisode^2 \mathrm{KL}\left(\ecomkerMDP[(r),h](\cdot | s,a ) \big\Vert \comkerMDP[h](\cdot|s,a)\right)
        \eqsp.
    \end{align}
    Furthermore, since $\cE^{\KL}(\delta)$ holds, we have the inequality $\KL\left(\ecomkerMDP[(r),h](s,a) \Big\Vert \comkerMDP[h](s,a)\right) \leq \frac{\nstates \beta^{\KL}(\delta, N_{(r),h}(s,a))}{N_{(r),h}(s,a)}$.
    Plugging this bound in \eqref{eq:bound-square-variance-phat-p}, we can upper bound \eqref{eq:bound-diff-phat-p-f} as
    \begin{align*}
        \!\!(\ecomkerMDP[(r),h] - \comkerMDP[h]) f (\cdot | s,a )
        \leq 
         \frac{1}{\sepisode} \comkerMDP[h] (f) (s,a) 
        + \sepisode^2 \frac{\nstates \beta^{\KL}(\delta, N_{(r),h}(s,a))}{N_{(r),h}(s,a)}
        + \frac{2\sepisode }{3} \frac{\nstates \beta^{\KL}(\delta, N_{(r),h}(s,a))}{N_{(r),h}(s,a)} \eqsp,
    \end{align*}
    which gives the result. The second inequality follows from the combination of Pinsker inequality and the definition of $\cE^{\KL}(\delta)$.
\end{proof}

%% file: 2025-AISTATS/appendix/regret_new.tex
\section{REGRET ANALYSIS}
\label{sec:regretanalysis}
We define the common MDP $\cM^{c}$ as
\begin{align}
\cM^{c} := (\S, \A, \sepisode, \{\rewardMDP[h][c] := \frac{1}{\nagent}\sum_{i=1}^\nagent \rewardMDP[h][i]\}_{h } , \{\comkerMDP[h]\}_{h} ) \eqsp.
\end{align}
We denote by $ \comstarvaluefunc[h]$ and $\comstarqfunc[h]$ the value function and the Q-function at step $h$ in the common environment $\cM^{c}$. In particular, these functions satisfy Bellman's equations and Bellman's optimality equations \citep{sutton2018}
\begin{align}
\label{def:bellmanequation_common}
    \comqfunc[h][\policy](s,a) &=  \rewardMDP[h][c](s,a) + \comkerMDP[h]\comvaluefunc[h+1][\policy](s,a)\eqsp, \quad
    &\comvaluefunc[h][\policy](s) = \comqfunc[h][\policy](s, \policy_h(s))\eqsp \\
\label{def:bellmanoptimalequation_common}
    \comstarqfunc[h](s,a) &=  \rewardMDP[h][c](s,a) +\comkerMDP[h]\comstarvaluefunc[h+1](s,a)\eqsp,\quad 
    &\comstarvaluefunc[h](s) = \max_{a \in \cA} \comstarqfunc[h](s,a)\eqsp,
\end{align}

\subsection{Optimism }
Let us define the following event
\begin{align*}
    \cE^{\mathrm{optimism}} = \Bigg\{ \forall r \in [\troundsmax], \forall (s,a,h) \in \S \times \A \times [\sepisode]:  \evaluefunc[(r),h](s) &\geq \comstarvaluefunc[h](s) - (2\hgreward + 3\hgkernel\sepisode)(\sepisode +1 -h), \\
    \eqfunc[(r),h](s,a) &\geq \comstarqfunc[h](s,a)- (2\hgreward + 3\hgkernel\sepisode)(\sepisode +1-h) \Bigg\}\,.
\end{align*}
Then, we will show that this event holds on event $\cG(\delta)$. To prove the optimism of our estimates, we use the same monotonicity arguments as in \cite{zhang2021reinforcement}, see also \cite{zhang2024settling}. Define
\begin{align}
\label{eq:monotonic_function}
g(\mathsf{P},f,\alpha) = \mathsf{P}f + \max (  \sqrt{ \alpha \Var_{\mathsf{P}}(f) }, \alpha \sepisode ) \eqsp,
\end{align}
where $\mathsf{P}$ be is a probability measure on $\S$, $f\in \rset^{\nstates}$ is a non negative vector satisfying $\Vert f \Vert_\infty \leq \sepisode$, and $\alpha$ is a positive real number.
\begin{lemma}[Lemma 14 by \citealt{zhang2021reinforcement}]
\label{lem:monotonicity}
The function $g$ is non-decreasing in each entry of $f$.
\end{lemma}
For completeness, we provide the proof below.
\begin{proof}
To justify this claim, consider any $ s \in \S$, and let us fix $\mathsf{P}, \alpha$ and all but the $s$-th entries of $f$. It then suffices to observe that (i) $g$ is a differentiable almost anywhere function, and (ii) except for at most two possible choices of $f(s)$ that obey $\sqrt{ \alpha \Var_{\mathsf{P}}(f) } =  \alpha \sepisode $, one can use the properties of $\mathsf{P}$ and $f$ to calculate
\begin{align}
\nonumber
\frac{\partial g(\mathsf{P}, f, \alpha)}{\partial f(s)} & =\mathsf{P}(s)+ \sqrt{\alpha} \mathbf{1}\left\{ \sqrt{ \alpha \Var_{\mathsf{P}}(f) }\ge \alpha \sepisode \right\} \frac{\mathsf{P}(s)(f(s)- \PE_{s' \sim \mathsf{P}}[f(s')])}{\sqrt{\Var_{\mathsf{P}}(f)}} \\ \nonumber
& =\mathsf{P}(s)+\mathbf{1} \left\{ \sqrt{ \alpha \Var_{\mathsf{P}}(f) }\ge \alpha \sepisode \right\} \frac{ \alpha \sepisode}{\sqrt{ \alpha \Var_{\mathsf{P}}(f) }} \cdot \frac{\mathsf{P}(s)(f(s)- \PE_{s' \sim \mathsf{P}}[f(s')])}{\sepisode} 
\\ \nonumber
& \geq \min \left\{\mathsf{P}(s)+\mathsf{P}(s) \frac{(f(s)- \PE_{s' \sim \mathsf{P}}[f(s')])}{\sepisode}, \mathsf{P}(s)\right\}
\\ \nonumber
& \geq \mathsf{P}(s) \min \left\{\frac{\sepisode+f(s)- \PE_{s' \sim \mathsf{P}}[f(s')]}{\sepisode}, 1\right\} \geq 0 \eqsp,
\end{align}
where in the end we used the fact that $\norm{f}[\infty] \leq \sepisode$.
\end{proof}
We define the bonus function as 
\begin{align}
\label{eq:bonus_function}
\bonus_{(r),h}(s,a) 
:= \begin{cases}
     \frac{28\beta^\star(\delta)\sepisode + 11\beta^{\common}(\delta, N)}{N} + \sqrt{ \frac{8 \beta^\star(\delta)}{N}\cdot \empvar[(r)]{\expandafter{\evaluefunc[(r),h+1]}}(s,a)}\eqsp, & N \geq 2 \\ 
    H\,, & N  \leq 1
\end{cases}
\end{align}
for $N= \counter_{(r),h}(s,a)$ and where $\beta^\star$ and $\beta^{\common}$ are defined in \Cref{lem:proba_master_event}.
\begin{lemma}\label{lem:optimism}
    Under conditions of \Cref{lem:proba_master_event}, it holds $\cE^{\mathrm{optimism}} \subseteq \cG(\delta)$ for any $\delta \in (0,1)$.
\end{lemma}
\begin{proof}
We process the proof by backward induction over $h$.
\paragraph{Base case} For $h = H+1$ and for all $(s,a,r) \in \S \times \A \times [\troundsmax]$, we have  
\begin{align}
\nonumber
\evaluefunc[(r),h](s)
= 0 \geq \comstarvaluefunc[h](s) - 0  = 0  \eqsp \text{ and }
\eqfunc[(r),h](s,a)  = 0 \geq \comstarqfunc[h](s,a) -  0 = 0\eqsp,
\end{align}
which gives the desired result.

\paragraph{Induction} Let $ h \in [\sepisode] $ such that for all $(s,a,r) \in \S\times \A\times[\troundsmax]$ and $h' \geq h$
\begin{align}
\label{eq:induction_assumption}
 \evaluefunc[(r),h'](s) &\geq \comstarvaluefunc[h'](s)  - (2\hgreward + 3\hgkernel\sepisode)(\sepisode+1 -h)\eqsp, \text{ and }\\
 \eqfunc[(r),h'](s,a) &\geq \comstarqfunc[h'](s,a) - (2\hgreward + 3\hgkernel\sepisode)(\sepisode+1 -h) \eqsp.
\end{align}
First, let us consider a trivial case  $\eqfunc[(r),h](s,a) = \sepisode$. The result is trivial since $\sepisode \geq \comstarqfunc[h](s,a) $.

Next, we assume that $\eqfunc[(r),h](s,a) < \sepisode $. In particular, by the definition of bonuses, it automatically follows that $N_{(r),h}(s,a) \ge 
    2$. In this case, according to the update rule \eqref{eq:q_agregation}, we have
\begin{align}
\nonumber
\eqfunc[(r),h](s,a) &\ge \sum_{i=1}^\nagent \frac{n^{i}_{(r),h}(s,a)}{ N_{(r),h}(s,a)} \eqfunc[(r),h][i](s,a) + \bonus_{(r),h}(s,a) 
\\ \nonumber
& = \sum_{i=1}^\nagent \frac{n^{i}_{(r),h}(s,a)}{ N_{(r),h}(s,a)} \erewardMDP[h]^{\,i}(s,a) + \ekerMDP[(r),h]  \evaluefunc[(r),h+1](s,a) + \bonus_{(r),h}(s,a) 
\\ \nonumber
& = \frac{1}{\nagent} \sum_{i=1}^{\nagent} \rewardMDP[h][i](s,a) + \comkerMDP[h]\comstarvaluefunc[h+1](s,a) + \bonus_{(r),h}(s,a) + \underbrace{\sum_{i=1}^\nagent \frac{n^{i}_{(r),h}(s,a)}{ N_{(r),h}(s,a)} \erewardMDP[h]^{\,i}(s,a) - \frac{1}{\nagent} \sum_{i=1}^{\nagent} \rewardMDP[h][i](s,a)}_{\term{I}}
\\ \label{eq:optimism_decomposition}
&+ \underbrace{\ekerMDP[(r),h]  (\evaluefunc[(r),h+1](s,a) - \comstarvaluefunc[h+1](s,a))}_{\term{II}}
+ \underbrace{(\ekerMDP[(r),h]-\ecomkerMDP[(r),h])\comstarvaluefunc[h+1](s,a)}_{\term{III}} +  \underbrace{( \ecomkerMDP[(r),h] -\comkerMDP[h])\comstarvaluefunc[h+1](s,a)}_{\term{IV}} \eqsp.
\end{align}
\paragraph{Terms $\term{I}$ and $\term{III}$: heterogeneity errors} First, let us handle the terms that come from the presence of heterogeneity between agents. To analyse $\term{I}$, recall that since for all $(s, a, i, h, r) \in \S \times \A \times [\nagent] \times [\sepisode] \times [\troundsmax]$, either 1) $n^{i}_{(r),h}(s,a) = 0$ and the value of $\erewardMDP[h]^{\,i}(s,a)$ is ignored in the weighted sum, or 2) $n^{i}_{(r),h}(s,a) > 0$ and  $\erewardMDP[h]^{\,i}(s,a) = \rewardMDP[h][i](s,a)$.
Thus, $\sum_{i=1}^\nagent \frac{n^{i}_{(r),h}(s,a)}{ N_{(r),h}(s,a)} \erewardMDP[h]^{\,i}(s,a)$ is a convex combination of the true rewards over $i$, which ensures that 
\begin{align}
\label{eq:bound_on_A_optimism}
 \term{I} = \sum_{i=1}^\nagent \frac{n^{i}_{(r),h}(s,a)}{ N_{(r),h}(s,a)} \erewardMDP[h]^{\,i}(s,a) - \frac{1}{\nagent} \sum_{i=1}^{\nagent} \rewardMDP[h][i](s,a) \geq - 2 \hgreward \eqsp .
\end{align}
Conditioned on $\cE^{\common}(\delta)$, Hölder's inequality yields the following bound on $\term{III}$
\begin{equation}
\label{eq:bound_on_C_optimism}
\term{III} = (\ekerMDP[(r),h]-\ecomkerMDP[(r),h])\comstarvaluefunc[h+1](s,a) 
\ge -\Vert \ekerMDP[(r),h] - \ecomkerMDP[(r),h] \Vert_1 \cdot \Vert \comstarvaluefunc[h+1] \Vert_\infty
\ge - 2\hgkernel \sepisode- \frac{11\beta^{\common}(\delta, \counter_{(r),h}(s,a))}{\counter_{(r),h}(s,a)} \sepisode \eqsp.
\end{equation}
\paragraph{Term $\term{II}$: correction error} To control this term, we aim to apply \Cref{lem:monotonicity}. We first define the \textit{shifted estimator} $\sevaluefunc[(r),h+1]$ as
\begin{align}
\label{def:shifted_version}
\sevaluefunc[(r),h+1](s) := \evaluefunc[(r),h+1](s) + (2 \hgreward + 3 \hgkernel\sepisode) (\sepisode-h)\eqsp. 
\end{align}
By the induction hypothesis \eqref{eq:induction_assumption}, we know that $\sevaluefunc[(r),h+1](s) \geq \comstarvaluefunc[h+1](s,a) $.
We decompose further $\term{II}$ as 
\begin{align}
\nonumber
 \term{II} &= \ekerMDP[(r),h]  \evaluefunc[(r),h+1](s,a) - \ekerMDP[(r),h] \comstarvaluefunc[h+1](s,a)
\\ \nonumber
&\geq \ekerMDP[(r),h]  \sevaluefunc[(r),h+1](s,a) + \max\bigg(\sqrt{\frac{4\beta^\star(\delta)\empvar[(r)]{\expandafter{\sevaluefunc[(r),h+1](s)}}(s,a)}{N_{(r),h}(s,a)}},\frac{4\beta^\star(\delta)\sepisode)}{N_{(r),h}(s,a)}\bigg) - \ekerMDP[(r),h] \comstarvaluefunc[h+1](s,a) 
\\ \nonumber
&- \sqrt{\frac{4\beta^\star(\delta)\empvar[(r)]{\expandafter{\sevaluefunc[(r),h+1](s)}}(s,a)}{N_{(r),h}(s,a)}} -\frac{4\beta^\star(\delta)\sepisode}{N_{(r),h}(s,a)} - (2 \hgreward + 3  \hgkernel \sepisode)(H-h) \eqsp, 
\end{align}
where we used in the last inequality that for any $a, b \in \rset_{+}, \max(a,b) \le a+b$ and the fact that $\ekerMDP[(r),h]  \evaluefunc[(r),h+1](s,a) - \ekerMDP[(r),h]  \sevaluefunc[(r),h+1](s,a) = - (2 \hgreward + 3  \hgkernel \sepisode) (\sepisode - h)$. Now by applying \Cref{lem:monotonicity}, we get
\begin{align}
\nonumber
 \term{II} &\geq   \max\bigg(\sqrt{\frac{4\beta^\star(\delta)\empvar[(r)]{\expandafter{\comstarvaluefunc[(r),h+1]}}(s,a)}{N_{(r),h}(s,a)}},\frac{4\beta^\star(\delta)\sepisode}{N_{(r),h}(s,a)}\bigg) - \sqrt{\frac{4\beta^\star(\delta)\empvar[(r)]{\expandafter{\sevaluefunc[(r),h+1]}}(s,a)}{N_{(r),h}(s,a)}} -\frac{4\beta^\star(\delta)\sepisode}{N_{(r),h}(s,a)} 
 \\ \label{eq:bound_on_B_optimism_1}
 &\geq   \underbrace{\sqrt{\frac{4\beta^\star(\delta)\empvar[(r)]{\expandafter{\comstarvaluefunc[(r),h+1]}}(s,a)}{N_{(r),h}(s,a)}}}_{\term{1}} - \underbrace{\sqrt{\frac{4\beta^\star(\delta)\empvar[(r)]{\expandafter{\sevaluefunc[(r),h+1]}}(s,a)}{N_{(r),h}(s,a)}}}_{\term{2}} -\frac{4\beta^\star(\delta)\sepisode}{N_{(r),h}(s,a)} \eqsp.
\end{align}
We want now to control the variance terms that appear in $\term{1}$ and $\term{2}$. Using inequalities \eqref{eq:distribution_transfert} and \eqref{eq:function_transfert} of \Cref{lem:transportation_measure}, we have 
\begin{align*}
\empvar[(r)]{\expandafter{\comstarvaluefunc[h+1]}}(s,a) &\ge \comempvar[(r),h]{\expandafter{\comstarvaluefunc[h+1]}}(s,a) - 3\sepisode^2 \hgkernel \eqsp,\\
\empvar[(r)]{\expandafter{\sevaluefunc[(r),h+1]}}(s,a) &\leq 2 \empvar[(r)]{\expandafter{\evaluefunc[(r),h+1]}}(s,a) + 2\ecomkerMDP[(r),h] |\sevaluefunc[(r),h+1] -\evaluefunc[(r),h+1] | \\
&\leq 2 \empvar[(r)]{\expandafter{\evaluefunc[(r),h+1]}}(s,a) + 2(2\hgreward + 3 \hgkernel \sepisode)(\sepisode - h) \eqsp,
\end{align*}
where in the last inequality we used the induction hypothesis.
 Besides, as for any $a, b ,c \in \rset_{+}$, we have $ a \ge b-c \implies \sqrt{a}\ge \sqrt{b} -\sqrt{c}$, and also for any $d,f\in \rset_{+}$ we have $\sqrt{d+f} \leq \sqrt{d} + \sqrt{f}$, we get
\begin{align}
\label{eq:bound_on_B_optimism_2}
\term{1} &:= \sqrt{\frac{4\beta^\star(\delta)\empvar[(r)]{\expandafter{\comstarvaluefunc[(r),h+1]}}(s,a)}{N_{(r),h}(s,a)}} \geq \sqrt{\frac{4\beta^\star(\delta)\comempvar[(r)]{\expandafter{\comstarvaluefunc[(r),h+1]}}(s,a)}{N_{(r),h}(s,a)}} -\sqrt{\frac{12 \hgkernel \sepisode^2\beta^\star(\delta)}{N_{(r),h}(s,a)}} \eqsp, \text{ and }\\
\label{eq:bound_on_B_optimism_3}
\term{2} &:= \sqrt{\frac{4\beta^\star(\delta)\empvar[(r)]{\expandafter{\sevaluefunc[(r),h+1]}}(s,a)}{N_{(r),h}(s,a)}} \leq \sqrt{\frac{8\beta^\star(\delta)\empvar[(r)]{\expandafter{\evaluefunc[(r),h+1]}}(s,a)}{N_{(r),h}(s,a)}} + \sqrt{\frac{8\beta^\star(\delta)(3\hgkernel \sepisode +2 \hgreward)(\sepisode+1 -h)}{N_{(r),h}(s,a)}}
\end{align}
Plugging the inequalities \eqref{eq:bound_on_B_optimism_2} and \eqref{eq:bound_on_B_optimism_3} in \eqref{eq:bound_on_B_optimism_1}, we obtain
\begin{align}
\nonumber
\term{II} \geq &\sqrt{\frac{4\beta^\star(\delta)\comempvar[(r)]{\expandafter{\comstarvaluefunc[(r),h+1]}}(s,a)}{N_{(r),h}(s,a)}} -  \sqrt{\frac{8\beta^\star(\delta)\empvar[(r)]{\expandafter{\evaluefunc[(r),h+1]}}(s,a)}{N_{(r),h}(s,a)}} -\sqrt{\frac{12 \hgkernel \sepisode^2\beta^\star(\delta)}{N_{(r),h}(s,a)}}  -\frac{4\beta^\star(\delta)\sepisode}{N_{(r),h}(s,a)} \\
&- \sqrt{\frac{8\beta^\star(\delta)(3\hgkernel \sepisode +2 \hgreward)(\sepisode+1 -h)}{N_{(r),h}(s,a)}} \eqsp.
\end{align}
Finally, as for any $a,b\in \rset{+}$ we have $\sqrt{2ab} \leq a +b$ then
\begin{align}
\nonumber
\term{II} &\geq \sqrt{\frac{4\beta^\star(\delta)\comempvar[(r)]{\expandafter{\comstarvaluefunc[(r),h+1]}}(s,a)}{N_{(r),h}(s,a)}} -  \sqrt{\frac{8\beta^\star(\delta)\empvar[(r)]{\expandafter{\evaluefunc[(r),h+1]}}(s,a)}{N_{(r),h}(s,a)}}  -  \hgkernel \sepisode - \frac{10\beta^\star(\delta)\sepisode}{N_{(r),h}(s,a)} \\
\label{eq:bound_on_B_optimism}
&-(3\hgkernel \sepisode +2 \hgreward)(\sepisode -h) -\frac{4\beta^\star(\delta)\sepisode}{N_{(r),h}(s,a)}
\end{align}
\paragraph{Term $\term{IV}$: concentration error} Conditioned on $ \cE^\star(\delta)$, we have
\begin{align}
\nonumber
\term{IV} &= ( \ecomkerMDP[(r),h] -\comkerMDP[h])\comstarvaluefunc[h+1](s,a) \geq -\left| [\ecomkerMDP[(r),h] - \comkerMDP[h]] \comstarvaluefunc[h+1](s,a)   \right|
\\ \label{eq:bound_on_D_optimism}
&\geq -\sqrt{\frac{2\comempvar[(r)]{\expandafter{\comstarvaluefunc[h+1]}}(s,a) \beta^\star(\delta)}{{\counter_{(r),h}(s,a)}-1}} - \frac{7\beta^\star(\delta)}{{\counter_{(r),h}(s,a)}-1} \geq -\sqrt{\frac{4\comempvar[(r)]{\expandafter{\comstarvaluefunc[h+1]}}(s,a) \beta^\star(\delta)}{{\counter_{(r),h}(s,a)}}} -\frac{14\beta^\star(\delta)}{{\counter_{(r),h}(s,a)}} \eqsp, 
\end{align} 
as for $n \geq 2$, we have $\frac{2}{n} \geq \frac{1}{n-1}$.
\paragraph{Combine everything together}By plugging in the bounds on $\term{I}$, $\term{II}$, $\term{III}$, and $\term{IV}$ in \eqref{eq:optimism_decomposition}, we get
\begin{align}
\nonumber
&\eqfunc[(r),h](s,a) \ge \frac{1}{\nagent} \sum_{i=1}^{\nagent} \rewardMDP[h][i](s,a) + \comkerMDP[h]\comstarvaluefunc[h+1](s,a) + \bonus_{(r),h}(s,a) - \frac{11\beta^{\common}(\delta, \counter_{(r),h}(s,a))}{\counter_{(r),h}(s,a)} \sepisode - \frac{28\beta^\star(\delta)\sepisode}{N_{(r),h}(s,a)}
\\ \nonumber
&  -\sqrt{\frac{8\beta^\star(\delta)\empvar[(r)]{\expandafter{\evaluefunc[(r),h+1]}}(s,a)}{N_{(r),h}(s,a)}}   -(3\hgkernel \sepisode +2 \hgreward)(\sepisode +1 -h) = \comstarqfunc[h](s,a) -  (2\hgreward + 3\hgkernel\sepisode)(\sepisode+1 -h) \eqsp,
\end{align}
where the last inequality is a consequence of the definition of the bonus \eqref{eq:bonus_function} and optimal Bellman equations \eqref{def:bellmanoptimalequation_common}. 
\end{proof}

\subsection{Regret decomposition}

We will start by writing down a regret decomposition. Let us define the essential technical quantities, such as common regret and partial common upper regret
\[
    \regret^{\common}(T) := \frac{1}{\nagent} \sum_{t=1}^T \sum_{i=1}^\nagent \comstarvaluefunc[1](s_{t,1}^i) - \comvaluefunc[1][\pi_{t}](s_{t,1}^i), \qquad \overline{\regret}^{\common}_{h}(T):= \frac{1}{\nagent} \sum_{t=1}^T \sum_{i=1}^\nagent \evaluefunc[t,h](s_{t,h}^i) - \comvaluefunc[h][\pi_{t}](s_{t,h}^i)
\]

\begin{lemma}
\label{lem:rht_bound}
    Assume conditions of \Cref{lem:proba_master_event}. Then, on the event $\cG(\delta)$, the following inequality for any partial upper common regret holds
    \[
        \overline{\regret}^{\common}_h(T) \leq U^T_h := A^T_h + B^T_h + C^T_h + 7\rme T\sepisode^2 \hgkernel + 2\rme T\sepisode \hgreward +
       \sqrt{8\sepisode^2 \cdot T\sepisode \cdot \beta(\delta) / \nagent} + \frac{1}{\nagent}\sum_{i=1}^{\nagent}\sum_{t=1}^T \sum_{h'=h}^\sepisode 2\rme \sepisode \Ind_{[\![0;1]\!]}( \bar{\counter}^i_{t,h'}) ,
    \]
    where
    \begin{align*}
        \bar{\counter}^i_{t,h} &:= \counter_{(r_t),h}(s^i_{t,h}, a^i_{t,h})\eqsp, \\
        A^T_h &:= \frac{\rme}{\nagent}\sum_{i=1}^{\nagent}\sum_{t=1}^T \sum_{h'=h}^\sepisode \sqrt{\frac{4\beta^\star(\delta) \comempvar[t,h']{\expandafter{\comstarvaluefunc[h+1]}}(s^i_{t,h'}, a^i_{t,h'}) }{{\bar{\counter}^i_{t,h'}}}} \Ind_{[\![2;+\infty]\!]}( \bar{\counter}^i_{t,h'} )  \eqsp, \\
        B^T_h &:= \frac{\rme}{\nagent}\sum_{i=1}^{\nagent}\sum_{t=1}^T \sum_{h'=h}^\sepisode \sqrt{\frac{8\beta^\star(\delta)\comempvar[t,h]{\expandafter{\evaluefunc[t,h'+1]}}(s^i_{t,h'}, a^i_{t,h'})}{\bar{\counter}^i_{t,h'}}} \Ind_{[\![2;+\infty]\!]}( \bar{\counter}^i_{t,h'} )  \eqsp, \\
        C^T_h &:=\frac{\rme}{\nagent}\sum_{i=1}^{\nagent}\sum_{t=1}^T \sum_{h'=h}^\sepisode \frac{22\beta^{\common}(\delta, \bar{\counter}^i_{t,h})+ 46\sepisode \beta^\star(\delta) + 2\sepisode^2 \nstates \beta^{\KL}(\delta, \bar{\counter}^i_{t,h}) }{{\bar{\counter}^i_{t,h}}}   \Ind_{[\![2;+\infty]\!]}( \bar{\counter}^i_{t,h'} ) \eqsp.
    \end{align*}
\end{lemma}
\begin{proof}    
    Let us define $\delta^i_{t,h} = \evaluefunc[t,h](s_{t,h}^i) - \comvaluefunc[1][\pi_{t}](s_{t,h}^i)$ and let us study this term separately. Since the policy is deterministic, i.e., $a^i_{t,h} = \pi_{t,h}(s_{t,h}^i)$, and satisfies $\evaluefunc[t,h](s^i_{t,h}) = \eqfunc[t,h](s^i_{t,h}, a^i_{t,h})$, we have
    \begin{align*}
        \delta^i_{t,h} &= \eqfunc[t,h](s_{t,h}^i, a_{t,h}^i) -  \comstarqfunc[h](s_{t,h}^i, a_{t,h}^i) + \comstarqfunc[h](s_{t,h}^i, a_{t,h}^i) -  \comqfunc[h][\pi_t](s_{t,h}^i, a_{t,h}^i).
    \end{align*}
    Next, for empirical Q-values, we have the following bound due to the clipping mechanism and \Cref{assum:lhg1}
    \begin{align*}
        \eqfunc[t,h](s,a) &\leq \sum_{i=1}^N \frac{n^i_{t,h}(s,a)}{\bar{\counter}^i_{t,h}} \erewardMDP[h]^{\,i}(s,a) + \ekerMDP[t,h] \evaluefunc[t,h+1](s,a) + b_{t,h}(s,a) \\
        &\leq \comrewardMDP[h](s,a) + \ekerMDP[t,h] \evaluefunc[t,h+1](s,a) + b_{t,h}(s,a) + 2\hgreward \,,
    \end{align*}
    thus, applying Bellman equations \eqref{def:bellmanequation_common} and optimal Bellman equations \eqref{def:bellmanoptimalequation_common}, we have after a simple rearranging
    \begin{align*}
        \delta^i_{t,h} &\leq \ekerMDP[t,h] \evaluefunc[t,h+1](s^i_{t,h}, a^i_{t,h}) - \comkerMDP[h] \comstarvaluefunc[h+1](s^i_{t,h}, a^i_{t,h}) + b_{t,h}(s^i_{t,h},a^i_{t,h}) + \comkerMDP[h] \left[ \comstarvaluefunc[h+1] - \comvaluefunc[h+1][\pi_t] \right](s^i_{t,h},a^i_{t,h}) + 2 \hgreward \\
        &= [\ekerMDP[t,h] - \comkerMDP[h]] \evaluefunc[t,h+1](s^i_{t,h}, a^i_{t,h}) + \comkerMDP[h]\left[ \evaluefunc[t,h+1] -  \comvaluefunc[h+1][\pi_t]\right](s^i_{t,h}, a^i_{t,h}) + \bonus_{t,h}(s^i_{t,h},a^i_{t,h}) + 2 \hgreward\,.
    \end{align*}
    In the decomposition above, we further rearrange it, using a virtual estimate of $\ecomkerMDP[t,h]$ defined in \eqref{eq:virtial_estimate_common_kernel} and re-introducing again the kernel for $i$-th agent $\kerMDP[h][i]$
    \begin{align}
        \delta^i_{t,h} &\leq \underbrace{[\ekerMDP[t,h] - \ecomkerMDP[t,h]] \evaluefunc[t,h+1](s^i_{t,h}, a^i_{t,h})}_{\term{A}} + \underbrace{[\ecomkerMDP[t,h] - \comkerMDP[h]]\left[\evaluefunc[t,h+1] - \comstarvaluefunc[h+1]\right](s^i_{t,h}, a^i_{t,h})}_{\term{B}} + \underbrace{[\ecomkerMDP[t,h] - \comkerMDP[h]]\comstarvaluefunc[h+1](s^i_{t,h}, a^i_{t,h})}_{\term{C}} \\
        &+ \underbrace{[\comkerMDP[h] - \kerMDP[h][i]]\left[ \evaluefunc[t,h+1] -  \comvaluefunc[h+1][\pi_t]\right](s^i_{t,h}, a^i_{t,h})}_{\term{D}} + \underbrace{ \kerMDP[h][i]\left[ \evaluefunc[t,h+1] -  \comvaluefunc[h+1][\pi_t]\right](s^i_{t,h}, a^i_{t,h}) -  \left[\evaluefunc[t,h+1] -  \comvaluefunc[h+1][\pi_t]\right](s^i_{t,h+1})}_{=: \zeta^i_{t,h}} \\
        \label{def:zeta_t_h}
        &+ \underbrace{\left[\evaluefunc[t,h+1] -  \comvaluefunc[h+1][\pi_t]\right](s^i_{t,h+1})}_{\delta^i_{t,h+1}} + b_{t,h}(s^i_{t,h},a^i_{t,h}) + 2\hgreward\,.
    \end{align}

    Next, we analyze each term separately. With a slight abuse of notation, let us define $\bar{\counter}^i_{t,h} = \counter_{(r_t),h}(s^i_{t,h}, a^i_{t,h})$. In the sequel, we analyze only such $(t,i,h) \in [T] \times [\nagent] \times [\sepisode]$ such that $\bar{\counter}^i_{t,h} \geq 2$. In the case where $\bar{\counter}^i_{t,h} \leq 1$, we have the trivial bound $\delta^i_{t,h} \leq \sepisode$.

    \paragraph{Terms $\term{A}$ and $\term{D}$: heterogeneity errors}

    First, let us handle the terms that come from the presence of heterogeneity between agents. To analyze $\term{A}$, let us apply the definition of the event $\cE^{\common}(\delta) \subseteq \cG(\delta)$ combined with Holder's inequality 
    \[
        \term{A}  \leq \sepisode \norm{\ekerMDP[t,h](s^i_{t,h}, a^i_{t,h}) - \ecomkerMDP[t,h](s^i_{t,h}, a^i_{t,h})}[1] \leq 2\sepisode \hgkernel + \frac{11\sepisode \beta^{\common}(\delta, \bar{\counter}^i_{t,h})}{\bar{\counter}^i_{t,h}}\,.
    \]

    For $\term{D}$ we apply Holder's inequality, \Cref{assum:lhg1} and \Cref{cor:performance-difference-tv}
    \[
       \term{D} \leq 2\sepisode \norm{\comkerMDP[h](s^i_{t,h}, a^i_{t,h}) - \kerMDP[h][i](s^i_{t,h}, a^i_{t,h})}[1] \leq 2\sepisode \hgkernel\,.
    \]

    \paragraph{Term $\term{B}$: correction error} To analyze this term, we apply \Cref{lem:bernstein_kl_corr} with $f(s') := [\evaluefunc[t,h+1] - \comstarvaluefunc[h+1]](s')$ and get
    \begin{align*}
        \term{B} &\leq \frac{1}{\sepisode} \comkerMDP[h] \left[\evaluefunc[t,h+1] - \comstarvaluefunc[h+1]\right](s^i_{t,h}, a^i_{t,h}) + \frac{2\sepisode^2 \nstates \beta^{\KL}(\delta, \bar{\counter}^i_{t,h})}{\bar{\counter}^i_{t,h}} \\
        & \overset{(1)}{\leq} \frac{1}{\sepisode} \comkerMDP[h] \left[\evaluefunc[t,h+1] - \comvaluefunc[h+1][\pi_t]\right](s^i_{t,h}, a^i_{t,h}) + \frac{2\sepisode^2 \nstates\beta^{\KL}(\delta,\bar{\counter}^i_{t,h})}{\bar{\counter}^i_{t,h}} \\
        & \overset{(2)}{\leq} \frac{1}{\sepisode} \term{D}  + \frac{1}{\sepisode} \delta^{i}_{t,h+1} + \frac{1}{\sepisode} \zeta^i_{t,h+1} + \frac{2\sepisode^2 \nstates \beta^{\KL}(\delta, \bar{\counter}^i_{t,h})}{\bar{\counter}^i_{t,h}}\,.
    \end{align*}
    where $(1)$ follows from the definition of optimal policy, and $(2)$ follows from a simple rearranging of terms, similar to the decomposition of $\delta^i_{t,h}$. Additional term $\term{D}$ appeared compared to a standard decomposition.

    \paragraph{Term $\term{C}$: concentration error} 
    From the definition of the event $\cE^\star(\delta) \subseteq \cG(\delta)$ defined in \Cref{lem:proba_master_event}, and from the analysis of the case $\bar{\counter}^i_{t,h} \geq 2$  it follows that
    \[
        \term{C} \leq \sqrt{\frac{2\comempvar[t,h]{\expandafter{\comstarvaluefunc[h+1]}}(s^i_{t,h}, a^i_{t,h}) \beta^\star(\delta)}{{\bar{\counter}^i_{t,h}}-1}} + \frac{7\beta^\star(\delta)}{{\bar{\counter}^i_{t,h}}-1} \leq \sqrt{\frac{4\comempvar[t,h]{\expandafter{\comstarvaluefunc[h+1]}}(s^i_{t,h}, a^i_{t,h}) \beta^\star(\delta)}{{\bar{\counter}^i_{t,h}}}} + \frac{14\beta^\star(\delta)}{{\bar{\counter}^i_{t,h}}} \,.
    \]
    
    \paragraph{Bounding the bonus} From the definition of the bonus \eqref{eq:bonus_function}, we have for all $(t,i,h) \in [T] \times [\nagent] \times [\sepisode]$ such that $\bar{\counter}^i_{t,h} \geq 2$
\begin{align}
\nonumber
\bonus_{t,h}(s^i_{t,h}, a^i_{t,h}) =\frac{28\beta^\star(\delta)\sepisode + 11\beta^{\common}(\delta, \bar{\counter}^i_{t,h})}{\bar{\counter}^i_{t,h}} + \sqrt{ \frac{8 \beta^\star(\delta)}{\bar{\counter}^i_{t,h}}\cdot \empvar[(r)]{\expandafter{\evaluefunc[(r),h+1]}}(s,a)} \eqsp.
\end{align}
Using the inequality \eqref{eq:distribution_transfert} of \Cref{lem:transportation_measure}, we have $ \empvar[t,h]{\expandafter{\evaluefunc[t,h+1]}}(s^i_{t,h}, a^i_{t,h}) \le \comempvar[t,h]{\expandafter{\evaluefunc[t,h+1]}}(s^i_{t,h}, a^i_{t,h}) + 3\sepisode^2 \hgkernel  $. Besides, as for any $a, b  \in \rset_{+}$, we have $\sqrt{a +b} \le \sqrt{a} +\sqrt{b}$, we get
\begin{align}
\nonumber
\sqrt{\frac{8\beta^\star(\delta)\empvar[t,h]{\expandafter{\evaluefunc[t,h+1]}}(s^i_{t,h}, a^i_{t,h})}{\bar{\counter}^i_{t,h}}} \leq \sqrt{\frac{8\beta^\star(\delta)\comempvar[t,h]{\expandafter{\evaluefunc[t,h+1]}}(s^i_{t,h}, a^i_{t,h})}{\bar{\counter}^i_{t,h}}} + \sqrt{\frac{24 \hgkernel \sepisode^2\beta^\star(\delta)}{\bar{\counter}^i_{t,h}}} \eqsp.
\end{align}
Now as for any $a, b  \in \rset_{+}$, we have $\sqrt{2ab} \le a + b$, we get
\begin{align}
\nonumber
\sqrt{\frac{8\beta^\star(\delta)\empvar[t,h]{\expandafter{\evaluefunc[t,h+1]}}(s^i_{t,h}, a^i_{t,h})}{\bar{\counter}^i_{t,h}}} \leq \sqrt{\frac{8\beta^\star(\delta)\comempvar[t,h]{\expandafter{\evaluefunc[t,h+1]}}(s^i_{t,h}, a^i_{t,h})}{\bar{\counter}^i_{t,h}}} + 3 \hgkernel \sepisode + \frac{4  \sepisode\beta^\star(\delta)}{\bar{\counter}^i_{t,h}} \eqsp.
\end{align}
\paragraph{Combine everything together} After combining all the terms, we have for all $(t,i,h) \in [T] \times [\nagent] \times [\sepisode]$ such that $\bar{\counter}^i_{t,h} \geq 2$
\begin{align*}
    \delta^i_{t,h} &\leq 7\sepisode^2\hgkernel
    + 2\hgreward  + \left(1 + \frac{1}{\sepisode} \right) \delta^i_{t,h+1} + \left( 1+\frac{1}{\sepisode} \right) \zeta^i_{t,h+1} + \frac{22\beta^{\common}(\delta, \bar{\counter}^i_{t,h})+ 46\sepisode \beta^\star(\delta) + 2\sepisode^2 \nstates \beta^{\KL}(\delta, \bar{\counter}^i_{t,h}) }{{\bar{\counter}^i_{t,h}}}\\
    &+\sqrt{\frac{4\comempvar[t,h]{\expandafter{\comstarvaluefunc[h+1]}}(s^i_{t,h}, a^i_{t,h}) \beta^\star(\delta)}{{\bar{\counter}^i_{t,h}}}}  + \sqrt{\frac{8\beta^\star(\delta)\comempvar[t,h]{\expandafter{\evaluefunc[t,h+1]}}(s^i_{t,h}, a^i_{t,h})}{\bar{\counter}^i_{t,h}}} \,.
\end{align*}
Let us define $\gamma_h = (1+1/\sepisode)^{\sepisode-h}$. Notice that for any $h \geq 0$ it holds $\gamma_h \leq \rme$. Then by summing and expanding over $h \in [\sepisode]$ we have
\begin{align*}
    \overline{\regret}^{\common}_h(T) &\leq 7\rme T\sepisode^2 \hgkernel + 2\rme T\sepisode \hgreward +
    \frac{1}{\nagent}\sum_{i=1}^{\nagent}\sum_{t=1}^T \sum_{h'=h}^\sepisode \gamma_{h'-1} \zeta^i_{t,h'+1} + \frac{1}{\nagent}\sum_{i=1}^{\nagent}\sum_{t=1}^T \sum_{h'=h}^\sepisode 2\rme \sepisode \Ind_{[\![0;1]\!]}( \bar{\counter}^i_{t,h'}) \\
    & + \frac{\rme}{\nagent}\sum_{i=1}^{\nagent}\sum_{t=1}^T \sum_{h'=h}^\sepisode \sqrt{\frac{4\beta^\star(\delta) \comempvar[t,h']{\expandafter{\comstarvaluefunc[h+1]}}(s^i_{t,h'}, a^i_{t,h'}) }{{\bar{\counter}^i_{t,h'}}}} \Ind_{[\![2;+\infty]\!]}( \bar{\counter}^i_{t,h'} )   &=: A^T_h\\
    &+ \frac{\rme}{\nagent}\sum_{i=1}^{\nagent}\sum_{t=1}^T \sum_{h'=h}^\sepisode \sqrt{\frac{8\beta^\star(\delta)\comempvar[t,h]{\expandafter{\evaluefunc[t,h'+1]}}(s^i_{t,h'}, a^i_{t,h'})}{\bar{\counter}^i_{t,h'}}} \Ind_{[\![2;+\infty]\!]}( \bar{\counter}^i_{t,h'} )  &=: B^T_h \\
    &+ \frac{\rme}{\nagent}\sum_{i=1}^{\nagent}\sum_{t=1}^T \sum_{h'=h}^\sepisode \frac{22\beta^{\common}(\delta, \bar{\counter}^i_{t,h})+ 46\sepisode \beta^\star(\delta) + 2\sepisode^2 \nstates \beta^{\KL}(\delta, \bar{\counter}^i_{t,h}) }{{\bar{\counter}^i_{t,h}}} \Ind_{[\![2;+\infty]\!]}( \bar{\counter}^i_{t,h'})\,. &=: C^T_h
\end{align*}
To conclude the statement, we apply a definition of the event $\cE(\delta)$ to the third term in the decomposition above. 
\end{proof}
\begin{lemma}
\label{lem:pigeon_hole_principle}
Define $  \bar{\counter}^i_{t,h} = \counter_{(r_t),h}(s^i_{t,h}, a^i_{t,h})$.  Assume conditions of \Cref{lem:proba_master_event}. Then, on the event $\cG(\delta)$, the following inequalities holds:
\begin{align*}
\sum_{i=1}^\nagent \sum_{t=1}^\nepisode \sum_{h=1}^\sepisode
\frac{\Ind_{[\![2;+\infty]\!]}( \bar{\counter}^i_{t,h})}{ \bar{\counter}^i_{t,h}}  &\leq  4\nstates \nactions \sepisode\log\left({\frac{\rme \nagent\nepisode \sepisode}{\nstates \nactions}}\right)  \eqsp,\\
\sum_{i=1}^\nagent \sum_{t=1}^\nepisode \sum_{h=1}^\sepisode
\sqrt{\frac{\Ind_{[\![2;+\infty]\!]}( \bar{\counter}^i_{t,h})}{N_{h,i}^{(r,l)}}} &\leq  8 \sepisode \sqrt{\nstates \nactions \nagent \nepisode} \eqsp, \\
\sum_{i=1}^{\nagent}\sum_{t=1}^T \sum_{h=1}^\sepisode \Ind_{[\![0;1]\!]}(\bar{\counter}^i_{t,h})
&\leq 4\sepisode \nstates \nactions \eqsp.
\end{align*}
\end{lemma}
\begin{proof}
The quantity $\parcounter^{\,i}_{t,h}(s,a)$ represents the exact number of visits of the pair $(s,a)$ at step $h$ until episode $t$, and after the first $i$ agents executed the $h$-step. We want to bound $ \bar{\counter}^i_{t,h}$ using $ \parcounter^{\,i}_{t,h}(s^i_{t,h}, a^i_{t,h}) $ so that we can compute the latter sums by applying the pigeon-hole principle. To derive such a bound, we distinguish two cases:

\paragraph{Case 1: $N_{r_t,h}(s,a) < \nu(\delta, \nepisode)$}
In this case, by the synchronization rule described in Algorithm \ref{algo:server-UCBVI}, we have $ n_{t,h}^{i}(s, a) < 2 n_{(r_t),h}^{i}(s, a) $. If we sum the latter inequality over all the agents, we obtain $\parcounter^{\,\nagent}_{t,h}(s, a)\leq 2 N_{(r_t),h}(s, a)  $. Now using definition of $\parcounter^{\,i}_{t,h}(s, a)$ yields
\begin{align*}
\counter_{(r_t),h}(s, a) \leq \parcounter^{\,i}_{t,h}(s, a) \leq 2 \counter_{(r_t),h}(s, a) \eqsp.
\end{align*}
\paragraph{Case 2: $\counter_{r_t,h}(s,a) \geq \nu(\delta, \nepisode)$} In this case, the synchronization rule ensures $\hat{\counter}^{\,i}_{t,h}(s,a) \leq 2 N_{r_t,h}(s,a)$. Conditioned on $\mathcal{E}^{\operatorname{count}}(\delta)$, we have $\parcounter^{\,\nagent}_{t,h}(s, a) \leq \frac{10}{7} \hat{N}^{\,i}_{t,h}(s,a)$. Combining the two latter inequalities gives
\begin{align*}
\counter_{(r_t),h}(s, a) \leq \parcounter^{\,i}_{t,h}(s, a) \leq 4 \counter_{(r_t),h}(s, a) \eqsp,
\end{align*}
where the lower bound follows directly from the definition of $\parcounter^{\,i}_{t,h}$.
Using the two previous inequalities, we derive the following bound
\begin{align*}
 \bar{\counter}^i_{t,h} \leq \parcounter^{\,i}_{t,h}(s^i_{t,h}, a^i_{t,h}) \leq 4\bar{\counter}^i_{t,h} \eqsp.
\end{align*}
Applying the latter inequality in the first sum of the lemma yields
\begin{align*}
\sum_{i=1}^\nagent \sum_{t=1}^\nepisode \sum_{h=1}^\sepisode
\frac{\Ind_{[\![2;+\infty]\!]}( \bar{\counter}^i_{t,h})}{ \bar{\counter}^i_{t,h}} \leq \sum_{i=1}^\nagent \sum_{t=1}^\nepisode \sum_{h=1}^\sepisode
\frac{4 \cdot \Ind_{[\![1;+\infty]\!]}( \tilde{\counter}^i_{t,h})}{ \tilde{\counter}^i_{t,h}} \eqsp.
\end{align*}
By construction, this counter is thus incremented by at most $1$ every time and we can apply the pigeon-hole principle on this counter which yields
\begin{align*}
\sum_{i=1}^\nagent \sum_{t=1}^\nepisode \sum_{h=1}^\sepisode
\frac{\Ind_{[\![2;+\infty]\!]}( \bar{\counter}^i_{t,h})}{ \bar{\counter}^i_{t,h}}
&
\leq\sum_{i=1}^\nagent \sum_{t=1}^\nepisode \sum_{h=1}^\sepisode
\frac{4 \cdot \Ind_{[\![1;+\infty]\!]}( \tilde{\counter}^i_{t,h})}{ \tilde{\counter}^i_{t,h}}  
\leq  4\sum_{h=1}^\sepisode \sum_{s\in \S}\sum_{a \in \A} \sum_{n=1}^{N_{\nepisode, h}(s,a)} \frac{1}{n} \\
&
\leq  4\sum_{h=1}^\sepisode\sum_{s\in \S}\sum_{a \in \A} (\log({N_{\nepisode, h}(s,a)}) + 1) \leq 4\nstates \nactions \sepisode\log\left({\frac{\rme \nagent\nepisode \sepisode}{\nstates \nactions}}\right) \eqsp,
\end{align*}
where we used the concavity of the logarithm in the last inequality. Similarly, we have
\begin{align*}
\sum_{i=1}^\nagent \sum_{t=1}^\nepisode \sum_{h=1}^\sepisode
\sqrt{\frac{\Ind_{[\![2;+\infty]\!]}( \bar{\counter}^i_{t,h})}{ \bar{\counter}^i_{t,h}}}
&
\leq\sum_{i=1}^\nagent \sum_{t=1}^\nepisode \sum_{h=1}^\sepisode
\sqrt{\frac{4 \cdot \Ind_{[\![1;+\infty]\!]}( \tilde{\counter}^i_{t,h})}{ \tilde{\counter}^i_{t,h}}} \\
& 
\leq  2\sum_{h=1}^\sepisode \sum_{s\in \S}\sum_{a \in \A} \sum_{n=1}^{N_{\nepisode, h}(s,a)} \sqrt{\frac{1}{n}} \leq 8\sum_{h=1}^\sepisode\sum_{s\in \S}\sum_{a \in \A} \sqrt{{N_{\nepisode, h}(s,a)}} \leq   8 \sepisode \sqrt{\nstates  \nactions \nagent \nepisode} \eqsp,
\end{align*}
where we used the concavity of the square root in the last inequality. Now as $ \parcounter^{\,i}_{t,h}(s^i_{t,h}, a^i_{t,h}) \leq 4\bar{\counter}^i_{t,h} $, then we have $ \Ind_{[\![0;1]\!]}( \bar{\counter}^i_{t,h}) \leq \Ind_{[\![0;4]\!]}( \tilde{\counter}^i_{t,h})$. Plugging in the latter inequality in the last sum of the lemma yields
\begin{align*}
\sum_{i=1}^{\nagent}\sum_{t=1}^T \sum_{h=1}^\sepisode \Ind_{[\![0;1]\!]}( \bar{\counter}^i_{t,h}) &\leq \sum_{i=1}^{\nagent}\sum_{t=1}^T \sum_{h=1}^\sepisode\Ind_{[\![0;4]\!]}( \tilde{\counter}^i_{t,h}) \leq \sum_{h=1}^\sepisode \sum_{s\in \S}\sum_{a \in \A} \sum_{n=1}^{4} 1 \leq 4\sepisode \nstates \nactions \eqsp.
\end{align*}
\end{proof} 
For ease of reading, we define $\beta^{\text{max}}(\delta)$ as 
\begin{align}
\label{eq:def-beta-max}
\beta^{\text{max}}(\delta) := \max\left( \beta^{\KL}(\delta, \nagent \nepisode),\beta^{\common}(\delta, \nagent \nepisode), \beta^\star(\delta), \beta(\delta),\beta^{\Var}(\delta, \nepisode), \log\left({\frac{\rme \nagent\nepisode \sepisode}{\nstates \nactions }}\right) \right)\,.
\end{align}
\begin{lemma}
\label{lem:bound_sum_variance}
Assume conditions of \Cref{lem:proba_master_event}. Then, on the event $\cG(\delta)$, the following inequality holds
\begin{align*}
 \frac{1}{\nagent}\sum_{i=1}^{\nagent}\sum_{t=1}^T \sum_{h=1}^\sepisode \comempvar[t,h]{\expandafter{\comstarvaluefunc[h+1]}}(s^i_{t,h}, a^i_{t,h}) \Ind_{[\![2;+\infty]\!]}( \bar{\counter}^i_{t,h}) &\leq 2 \sepisode^2 \nepisode + 2 \sepisode^2 U^T_1 +   11\sepisode^3 \nepisode \hgkernel + 6 \sepisode \nepisode \hgreward   \\
& +30 \sepisode^3 \beta^{\max}(\delta) \nstates \nactions^{1/2} \nepisode^{1/2} \nagent^{-1/ 2}\eqsp,
\end{align*}
and we also have
\begin{align*}
 \frac{1}{\nagent}\sum_{i=1}^{\nagent}\sum_{t=1}^T \sum_{h=1}^\sepisode \comempvar[t,h]{\evaluefunc[t,h+1]}(s^i_{t,h}, a^i_{t,h}) \Ind_{[\![2;+\infty]\!]}( \bar{\counter}^i_{t,h}) &\leq  2 \sepisode^2 \nepisode + 2 \sepisode^2 U^T_1 +   17\sepisode^3 \nepisode \hgkernel + 10 \sepisode \nepisode \hgreward   \\
& +30 \sepisode^3 \beta^{\max}(\delta) \nstates \nactions^{1/2} \nepisode^{1/2} \nagent^{-1/ 2}\eqsp,
\end{align*}
where $\beta^{\max}(\delta)$ is defined in \Cref{lem:proba_master_event} as a worst-case concentration logarithmic factor.
\end{lemma}
\begin{proof}
Using inequality \eqref{eq:distribution_transfert} of \Cref{lem:transportation_measure}, we have    
\begin{align*}
 \frac{1}{\nagent}\sum_{i=1}^{\nagent}\sum_{t=1}^T \sum_{h=1}^\sepisode \comempvar[t,h]{\expandafter{\comstarvaluefunc[h+1]}}(s^i_{t,h}, a^i_{t,h}) \Ind_{[\![2;+\infty]\!]}( \bar{\counter}^i_{t,h}) &\leq    \underbrace{\frac{1}{\nagent}\sum_{i=1}^{\nagent}\sum_{t=1}^T \sum_{h=1}^\sepisode \agentvar[h]{i}{\expandafter{\comstarvaluefunc[h+1]}}(s^i_{t,h}, a^i_{t,h}) \Ind_{[\![2;+\infty]\!]}( \bar{\counter}^i_{t,h})}_{\term{W}} \\
 & +  \underbrace{3 \sepisode^2 \frac{1}{\nagent}\sum_{i=1}^{\nagent}\sum_{t=1}^T \sum_{h
=1}^\sepisode \Vert (\kerMDP[h][i] - \ecomkerMDP[t,h])(s^i_{t,h}, a^i_{t,h}) \Vert_1 \Ind_{[\![2;+\infty]\!]}( \bar{\counter}^i_{t,h})}_{\term{X}} \eqsp.
\end{align*}
\textbf{Term $\term{X}$:} We have for any $(s,a) \in \S \times \A$
\begin{align}
\label{eq:defbetamax}
\Vert  \kerMDP[h][i](s,a) -\ecomkerMDP[t,h](s,a) \Vert_1 \leq \Vert \kerMDP[h][i](s,a) - \comkerMDP[h](s,a) \Vert_1 + \Vert \comkerMDP[h](s,a) - \ecomkerMDP[t,h](s,a) \Vert_1 \leq \hgkernel + \sqrt{\frac{2 \nstates\beta^{\KL}(\delta,  \bar{\counter}^i_{t,h})}{ \bar{\counter}^i_{t,h}}} \eqsp,
\end{align}
where the bound on $\Vert \kerMDP[h][i](s,a) - \comkerMDP[h](s,a) \Vert_1$ is provided by \Cref{lem:tvbound} and the bound on $ \Vert \comkerMDP[h](s,a) - \ecomkerMDP[t,h](s,a) \Vert_1$ is provided by the second inequality of \Cref{lem:bernstein_kl_corr}. Thus we get
\begin{align*}
3 \sepisode^2 \frac{1}{\nagent}\sum_{i=1}^{\nagent}\sum_{t=1}^T \sum_{h
=1}^\sepisode \Vert (\comkerMDP[h] - \ecomkerMDP[t,h])(s^i_{t,h}, a^i_{t,h}) \Vert_1 \Ind_{[\![2;+\infty]\!]}( \bar{\counter}^i_{t,h}) \leq \frac{ 3 \sepisode^2}{\nagent}\sum_{i=1}^{\nagent}\sum_{t=1}^T \sum_{h
=1}^\sepisode  \hgkernel +\sqrt{\frac{2 \nstates\beta^{\KL}(\delta, \bar{\counter}^i_{t,h})}{\bar{\counter}^i_{t,h}}} \Ind_{[\![2;+\infty]\!]}( \bar{\counter}^i_{t,h}) \eqsp.
\end{align*}
Finally applying \Cref{lem:pigeon_hole_principle} yields
\begin{align*}
\term{X} \leq 3\sepisode^3 \nepisode \hgkernel +    24\sqrt{\frac{2 \sepisode^6 \beta^{\KL}(\delta, \nagent \nepisode)\nstates^2  \nactions \nepisode}{\nagent} } \eqsp.
\end{align*}
\textbf{Term $\term{W}$:} Using inequality \eqref{eq:function_transfert} of \Cref{lem:transportation_measure}, we have
\begin{align*}
\frac{1}{\nagent}\sum_{i=1}^{\nagent}\sum_{t=1}^\nepisode \sum_{h=1}^\sepisode \agentvar[h]{i}{\expandafter{\comstarvaluefunc[h+1]}}(s^i_{t,h}, a^i_{t,h}) \Ind_{[\![2;+\infty]\!]}( \bar{\counter}^i_{t,h})
&\leq
\underbrace{\frac{1}{\nagent}\sum_{i=1}^{\nagent}\sum_{t=1}^\nepisode \sum_{h=1}^\sepisode 2\agentvar[h]{i}{\valuefunc[h+1][i,\pi_{t}]}(s^i_{t,h}, a^i_{t,h}) \Ind_{[\![2;+\infty]\!]}( \bar{\counter}^i_{t,h})}_{\term{Y}}
\\
&\leq
\underbrace{\frac{1}{\nagent}\sum_{i=1}^{\nagent}\sum_{t=1}^\nepisode \sum_{h=1}^\sepisode 2 \sepisode \kerMDP[h][i] | \comstarvaluefunc[h+1] - \valuefunc[h+1][i,\pi_{t}] |(s^i_{t,h}, a^i_{t,h})\Ind_{[\![2;+\infty]\!]}( \bar{\counter}^i_{t,h})}_{\term{Z}} \eqsp,
\end{align*}
where we recall that $\valuefunc[h+1][i,\pi_{t}]$ is the value function of the policy $\pi_{t}$ in the environment of the $i$-th agent \eqref{def:value_function_for_agent}.
Conditioned on $\mathcal{E}^{\operatorname{Var}}(\delta)$, we have
\begin{align*}
\term{Y} \leq  \sqrt{\frac{8\sepisode^5  \nepisode \beta^{\Var}(\delta, \nepisode)}{\nagent}}
+ \frac{6\sepisode^3  \beta^{\Var}(\delta, \nepisode)}{\nagent}
+ 2\sepisode^2 \nepisode \eqsp.
\end{align*}
Now by \Cref{cor:performance-difference-tv}, we have conditioned on $\cE^{\mathrm{optimism}} $ for all $s\in \S$
\begin{align*}
|\comstarvaluefunc[h+1](s) - \valuefunc[h+1][i,\pi_{t}](s) |&\leq |\comvaluefunc[h+1][\pi_{t}](s) - \valuefunc[h+1][i,\pi_{t}](s) |  + \comstarvaluefunc[h+1](s) - \comvaluefunc[h+1][\pi_{t}](s)   \\
&\leq \hgkernel \sepisode^2 + \hgreward \sepisode + \evaluefunc[t,h+1](s) - \comvaluefunc[h+1][\pi_{t}](s) +  (2\hgreward + 3\hgkernel\sepisode)(\sepisode -h) \\
&\leq 4\hgkernel \sepisode^2 + 3\hgreward \sepisode + \evaluefunc[t,h+1](s) - \comvaluefunc[h+1][\pi_{t}](s) \eqsp.
\end{align*}
Using the definition of $\delta^i_{t,h}$ and $\zeta^i_{t,h+1}$ introduced in \Cref{lem:rht_bound}, we have
\begin{align*}
\term{Z} &\leq \frac{1}{\nagent}\sum_{i=1}^{\nagent}\sum_{t=1}^\nepisode \sum_{h=1}^\sepisode 2 \sepisode ( 4\hgkernel \sepisode^2 + 3\hgreward + \delta^i_{t,h} + \zeta^i_{t,h+1} )
\\
&\leq 8 \sepisode^3 \nepisode \hgkernel + 6 \sepisode \nepisode \hgreward + \sqrt{\frac{8 \sepisode^5 \nepisode \beta(\delta)}{\nagent} } + 2 \sepisode \sum_{h=1}^\sepisode  \overline{\regret}^{\common}_{h+1}(\nepisode) \leq 8 \sepisode^3 \nepisode \hgkernel + 6 \sepisode \nepisode \hgreward + \sqrt{\frac{8 \sepisode^5 \nepisode \beta(\delta)}{\nagent} } + 2 \sepisode^2 U^T_1 \eqsp,
\end{align*}
where the second inequality holds conditioned on $ \cE(\delta) $. Combining everything yields
\begin{align*}
\frac{1}{\nagent}\sum_{i=1}^{\nagent}\sum_{t=1}^\nepisode \sum_{h=1}^\sepisode &\agentvar[h]{i}{\expandafter{\comstarvaluefunc[h+1]}}(s^i_{t,h}, a^i_{t,h}) \Ind_{[\![2;+\infty]\!]}( \bar{\counter}^i_{t,h}) \leq \term{X} + \term{Y} + \term{Z} 
\\ &\leq  11\sepisode^3 \nepisode \hgkernel +    24\sqrt{\frac{2 \sepisode^6 \beta^{\KL}(\delta, \nagent \nepisode)\nstates^2  \nactions \nepisode}{\nagent} }  +  \sqrt{\frac{8\sepisode^5  \nepisode \beta^{\Var}(\delta, \nepisode)}{\nagent}}
+ \frac{6\sepisode^3  \beta^{\Var}(\delta, \nepisode)}{\nagent}
+ 2\sepisode^2 \nepisode  \\
&+ 6 \sepisode \nepisode \hgreward + \sqrt{\frac{8 \sepisode^5 \nepisode \beta(\delta)}{\nagent} } + 2 \sepisode^2 U^T_1
\end{align*}
Now let's move to the second inequality of this lemma. Again by using inequality \eqref{eq:distribution_transfert} of \Cref{lem:transportation_measure}, we have    
\begin{align*}
 \frac{1}{\nagent}\sum_{i=1}^{\nagent}\sum_{t=1}^T \sum_{h=1}^\sepisode  \comempvar[t,h]{\evaluefunc[t,h+1]}(s^i_{t,h}, a^i_{t,h}) \Ind_{[\![2;+\infty]\!]}( \bar{\counter}^i_{t,h}) &\leq    \underbrace{\frac{1}{\nagent}\sum_{i=1}^{\nagent}\sum_{t=1}^T \sum_{h=1}^\sepisode \agentvar[h]{i}{\evaluefunc[t,h+1]}(s^i_{t,h}, a^i_{t,h}) \Ind_{[\![2;+\infty]\!]}( \bar{\counter}^i_{t,h})}_{\term{W'}} \\
 & +  \underbrace{3 \sepisode^2 \frac{1}{\nagent}\sum_{i=1}^{\nagent}\sum_{t=1}^T \sum_{h
=1}^\sepisode \Vert (\kerMDP[h][i] - \ecomkerMDP[t,h])(s^i_{t,h}, a^i_{t,h}) \Vert_1 \Ind_{[\![2;+\infty]\!]}( \bar{\counter}^i_{t,h})}_{\term{X}} \eqsp.
\end{align*}
\textbf{Term $\term{W'}$:}  Using inequality \eqref{eq:function_transfert} of \Cref{lem:transportation_measure}, we have
\begin{align*}
\frac{1}{\nagent}\sum_{i=1}^{\nagent}\sum_{t=1}^\nepisode \sum_{h=1}^\sepisode\agentvar[h]{i}{\evaluefunc[t,h+1]}(s^i_{t,h}, a^i_{t,h}) &\Ind_{[\![2;+\infty]\!]}( \bar{\counter}^i_{t,h})
\leq
\underbrace{\frac{1}{\nagent}\sum_{i=1}^{\nagent}\sum_{t=1}^\nepisode \sum_{h=1}^\sepisode 2\agentvar[h]{i}{\valuefunc[h+1][i,\pi_{t}]}(s^i_{t,h}, a^i_{t,h}) \Ind_{[\![2;+\infty]\!]}( \bar{\counter}^i_{t,h})}_{\term{Y}}
\\
&\leq
\underbrace{\frac{1}{\nagent}\sum_{i=1}^{\nagent}\sum_{t=1}^\nepisode \sum_{h=1}^\sepisode 2 \sepisode \kerMDP[h][i] | \evaluefunc[t,h+1] - \valuefunc[h+1][i,\pi_{t}] |(s^i_{t,h}, a^i_{t,h})\Ind_{[\![2;+\infty]\!]}( \bar{\counter}^i_{t,h})}_{\term{Z'}} \eqsp. 
\end{align*}
Now again by \Cref{cor:performance-difference-tv}, we have conditioned on $\cE^{\mathrm{optimism}} $ for all $s\in \S$
\begin{align*}
|\evaluefunc[t,h+1](s) - \valuefunc[h+1][i,\pi_{t}](s) |&\leq |\comvaluefunc[h+1][\pi_{t}](s) - \valuefunc[h+1][i,\pi_{t}](s) |  + \evaluefunc[t,h+1](s) +  (2\hgreward + 3\hgkernel\sepisode)(\sepisode -h) - \comvaluefunc[h+1][\pi_{t}] +  (2\hgreward + 3\hgkernel\sepisode)(\sepisode -h)    \\
&\leq \hgkernel \sepisode^2 + \hgreward \sepisode + \evaluefunc[t,h+1](s) - \comvaluefunc[h+1][\pi_{t}](s) +  2(2\hgreward + 3\hgkernel\sepisode)(\sepisode -h) \\
&\leq 7\hgkernel \sepisode^2 + 5\hgreward \sepisode + \evaluefunc[t,h+1](s) - \comvaluefunc[h+1][\pi_{t}](s) \eqsp.
\end{align*}
By combining the bounds that we have on $\term{X}, \term{Y}$, and $\term{Z'}$, we derive the following bound
\begin{align*}
\frac{1}{\nagent}\sum_{i=1}^{\nagent}\sum_{t=1}^T \sum_{h=1}^\sepisode &\comempvar[t,h]{\evaluefunc[t,h+1]}(s^i_{t,h}, a^i_{t,h}) \Ind_{[\![2;+\infty]\!]}( \bar{\counter}^i_{t,h}) \leq \term{X} + \term{Y} + \term{Z'} 
\\
&\leq  17\sepisode^3 \nepisode \hgkernel +    24\sqrt{\frac{2 \sepisode^6 \beta^{\KL}(\delta, \nagent \nepisode)\nstates^2  \nactions \nepisode}{\nagent} }  +  \sqrt{\frac{8\sepisode^5  \nepisode \beta^{\Var}(\delta, \nepisode)}{\nagent}}
+ \frac{6\sepisode^3  \beta^{\Var}(\delta, \nepisode)}{\nagent}
+ 10\sepisode^2 \nepisode  \\
&+ 10 \sepisode \nepisode \hgreward + \sqrt{\frac{8 \sepisode^5 \nepisode \beta(\delta)}{\nagent} } + 2 \sepisode^2 U^T_1 \eqsp.
\end{align*}
Finally, as we have
\begin{align*}
24\sqrt{\frac{2 \sepisode^6 \beta^{\KL}(\delta, \nagent \nepisode)\nstates^2  \nactions \nepisode}{\nagent} }  &\leq 48 \sepisode^3 \beta^{\max}(\delta) \nstates \nactions^{1/2} \nepisode^{1/2} \nagent^{-1/ 2} \\
\sqrt{\frac{8 \sepisode^5 \nepisode \beta(\delta)}{\nagent} } &\leq 3 \sepisode^3 \beta^{\max}(\delta) \nstates \nactions^{1/2} \nepisode^{1/2} \nagent^{-1/ 2} \\
 \sqrt{\frac{8\sepisode^5  \nepisode \beta^{\Var}(\delta, \nepisode)}{\nagent}} &\leq  3 \sepisode^3 \beta^{\max}(\delta) \nstates \nactions^{1/2} \nepisode^{1/2} \nagent^{-1/ 2} \\
 \frac{6\sepisode^3  \beta^{\Var}(\delta, \nepisode)}{\nagent} &\leq  6 \sepisode^3 \beta^{\max}(\delta) \nstates \nactions^{1/2} \nepisode^{1/2}  \nagent^{-1/ 2} \eqsp,
\end{align*}
then
\begin{align*}
\frac{1}{\nagent}\sum_{i=1}^{\nagent}\sum_{t=1}^T \sum_{h=1}^\sepisode \comempvar[t,h]{\evaluefunc[t,h+1]}(s^i_{t,h}, a^i_{t,h}) \Ind_{[\![2;+\infty]\!]}( \bar{\counter}^i_{t,h}) &\leq   17\sepisode^3 \nepisode \hgkernel + 10 \sepisode \nepisode \hgreward + 2 \sepisode^2 \nepisode + 2 \sepisode^2 U^T_1 \\
&\ + 60 \sepisode^3 \beta^{\max}(\delta) \nstates \nactions^{1/2} \nepisode^{1/2} \nagent^{-1/ 2}\eqsp.
\end{align*}
\end{proof}
\begin{lemma}\label{lem:bound_abc_terms}
Assume conditions of \Cref{lem:proba_master_event}. Then, on the event $\cG(\delta)$, the following inequality holds
\begin{align*}
A^T_1 &\leq 23\rme \cdot \beta^{\max}(\delta) \cdot \sqrt{\sepisode^3 \nstates \nactions \nepisode \nagent^{-1}} + 23 \rme  \beta^{\max}(\delta) \sqrt{\sepisode^3 \nstates \nactions U^{\nepisode}_1 \nagent^{-1}} \\
    &+ 16 \rme \beta^{\max}(\delta) \nepisode (6\sepisode^2 \hgkernel + 3 \sepisode \hgreward) + 96 \rme \sepisode^{3} \nstates^{3/2} \nactions \nagent^{-1/2} (\beta^{\max}(\delta))^2\eqsp,\\
B^{\nepisode}_1 &\leq 46\rme \cdot \beta^{\max}(\delta) \cdot \sqrt{\sepisode^3 \nstates \nactions \nepisode \nagent^{-1}} + 46 \rme  \beta^{\max}(\delta) \sqrt{\sepisode^3 \nstates \nactions U^{\nepisode}_1 \nagent^{-1}} \\
    &+ 32 \rme \beta^{\max}(\delta) \nepisode (6\sepisode^2 \hgkernel + 3 \sepisode \hgreward) + 192 \rme \sepisode^{3} \nstates^{3/2} \nactions \nagent^{-1/2} (\beta^{\max}(\delta))^2\eqsp, \\
C^T_1 &\leq   272 \nstates^2 \nactions \nagent^{-1}\sepisode^3 (\beta^{\max}(\delta))^{2}  \eqsp.
\end{align*}
\end{lemma}
\begin{proof}
\textbf{Term $A^T_1$.} To bound the term $A^T_1$, we start by applying Cauchy-Schwartz inequality
\begin{align*}
A^T_1 &= \frac{\rme}{\nagent}\sum_{i=1}^{\nagent}\sum_{t=1}^T \sum_{h'=h}^\sepisode \sqrt{\frac{4\beta^\star(\delta) \comempvar[t,h']{\expandafter{\comstarvaluefunc[h+1]}}(s^i_{t,h'}, a^i_{t,h'}) }{{\bar{\counter}^i_{t,h'}}}}\Ind_{[\![2;+\infty]\!]}( \bar{\counter}^i_{t,h'} )  \\
&\leq \frac{\rme}{\sqrt{\nagent}} \sqrt{\frac{1}{\nagent}\sum_{i=1}^{\nagent}\sum_{t=1}^T \sum_{h'=h}^\sepisode \comempvar[t,h']{\expandafter{\comstarvaluefunc[h+1]}}(s^i_{t,h'}, a^i_{t,h'})\Ind_{[\![2;+\infty]\!]}( \bar{\counter}^i_{t,h'} ) }  \sqrt{\sum_{i=1}^{\nagent}\sum_{t=1}^T \sum_{h'=h}^\sepisode \frac{4\beta^\star(\delta)\Ind_{[\![2;+\infty]\!]}( \bar{\counter}^i_{t,h'})}{\bar{\counter}^i_{t,h'}}} \eqsp,
\end{align*}
Now, applying \Cref{lem:bound_sum_variance}, \Cref{lem:pigeon_hole_principle}, and the subadditivity of the square root, we obtain
\begin{align*}
A_1^T \leq \frac{16  \beta^{\max}(\delta) \rme}{\sqrt{\nagent}} 
 \bigg( &\sqrt{2 \sepisode^3 \nstates \nactions \nepisode} + \sqrt{2 \sepisode^3\nstates \nactions U^{\nepisode}_1} + \sqrt{\sepisode^2 \nstates\nactions \nepisode \cdot (11\sepisode^2 \hgkernel + 6 \sepisode\hgreward)}  \\
&+
\sqrt{60 \sepisode^4 \nstates^2 \nactions^{3/2} \nepisode^{1/2} \nagent^{-1/2} \cdot \beta^{\max}(\delta)  }\bigg)  \eqsp.
\end{align*}
Next, we analyze the last two terms in the upper bound above. For the third one, by a standard inequality $\sqrt{2ab} \leq a+b$ it holds that
\[
    \sqrt{\sepisode^2 \nstates\nactions \nepisode \cdot (11\sepisode^2 \hgkernel + 6 \sepisode\hgreward)} \leq 3 \sepisode\nepisode \hgreward + 6 \sepisode^2 \nepisode \hgkernel + \sepisode^2 \nstates\nactions \eqsp.
\]
Notably, the first two terms already appeared in the regret decomposition; see \Cref{lem:rht_bound}. For the last term, the decomposition is more standard
\[
    \sqrt{60 \sepisode^4 \nstates^2 \nactions^{3/2} \nepisode^{1/2}\nagent^{-1/2} \beta^{\max}(\delta) } \leq  6\sqrt{\sepisode^3 \nstates \nactions \nepisode} + 5 \sepisode^{5/2} \nstates^{3/2} \nactions \nagent^{-1/2} \beta^{\max}(\delta) \eqsp.
\]
Thus, by a simple rearranging of the terms and applying inequalities $M \geq 1, H \geq 1$, we have
\begin{align*}
    A^{\nepisode}_1 &\leq 23\rme \cdot \beta^{\max}(\delta) \cdot \sqrt{\sepisode^3 \nstates \nactions \nepisode \nagent^{-1}} + 23 \rme  \beta^{\max}(\delta) \sqrt{\sepisode^3 \nstates \nactions U^{\nepisode}_1 \nagent^{-1}} \\
    &+ 16 \rme \beta^{\max}(\delta) \nepisode (6\sepisode^2 \hgkernel + 3 \sepisode \hgreward) + 96 \rme \sepisode^{3} \nstates^{3/2} \nactions \nagent^{-1/2} (\beta^{\max}(\delta))^2\eqsp.
\end{align*}

\textbf{Term $B^T_1$.} Similarly, the bound for the term $B^T_1$ is derived using a combination of Cauchy-Scwartz, \Cref{lem:bound_sum_variance}, \Cref{lem:pigeon_hole_principle}, and the subadditivity of the square root
\begin{align*}
    B^{\nepisode}_1 &\leq  46\rme \cdot \beta^{\max}(\delta) \cdot \sqrt{\sepisode^3 \nstates \nactions \nepisode \nagent^{-1}} + 46 \rme  \beta^{\max}(\delta) \sqrt{\sepisode^3 \nstates \nactions U^{\nepisode}_1 \nagent^{-1}} \\
    &+ 32 \rme \beta^{\max}(\delta) \nepisode (6\sepisode^2 \hgkernel + 3 \sepisode\hgreward) + 192 \rme \sepisode^{3} \nstates^{3/2} \nactions \nagent^{-1/2} (\beta^{\max}(\delta))^2\eqsp.
\end{align*}
\textbf{Term $C^T_1$.} Finally to estimate $C^T_1$, we apply \Cref{lem:pigeon_hole_principle}
\begin{align*}
 \frac{\rme}{\nagent}\sum_{i=1}^{\nagent}\sum_{t=1}^T \sum_{h'=h}^\sepisode &\frac{22\beta^{\common}(\delta, \bar{\counter}^i_{t,h})+ 46\sepisode \beta^\star(\delta) + 2\sepisode^2 \nstates \beta^{\KL}(\delta, \bar{\counter}^i_{t,h}) }{{\bar{\counter}^i_{t,h}}} \Ind_{[\![2;+\infty]\!]}( \bar{\counter}^i_{t,h'} )  \\
 & \leq \frac{68 \sepisode^2 \nstates\beta^{\max}(\delta)}{\nagent} \sum_{i=1}^{\nagent}\sum_{t=1}^T \sum_{h=1}^\sepisode \frac{1}{\bar{\counter}^i_{t,h}} \Ind_{[\![2;+\infty]\!]}( \bar{\counter}^i_{t,h}) \leq
 272 \nstates^2 \nactions \nagent^{-1}\sepisode^3 (\beta^{\max}(\delta))^{2} \eqsp.
\end{align*}

\end{proof}

\subsection{Proof of \Cref{thm:regret}}
Hereafter, we establish the following bound on the regret
\begin{align*}
\regret(T) \leq  138\rme  \sqrt{\sepisode^3 \nstates \nactions \nepisode \nagent^{-1} (\beta^{\max}(\delta) )^2} + 6020 \rme^2 \sepisode^3 \nstates^2 \nactions (\beta^{\max}(\delta))^{2} +\rme \beta^{\max}(\delta) \nepisode \sepisode (595 \sepisode \hgkernel + 148 \sepisode \hgreward) \eqsp.
\end{align*}
\begin{proof}
    Let us start by moving from our regret to a regret connected to a common kernel, using a combination of \Cref{cor:performance-difference-tv} and \Cref{assum:lhg1}
    \[
        \regret(T) =  \max_{\pi} \frac{1}{\nagent} \sum_{t=1}^T \sum_{i=1}^{\nagent} \valuefunc[1][i,\pi](s_{t,1}^i) - \valuefunc[1][i, \pi_{t}](s_{t,1}^i) \leq \underbrace{\frac{1}{\nagent} \sum_{t=1}^T \sum_{i=1}^\nagent \comstarvaluefunc[1](s_{t,1}^i) - \comvaluefunc[1][\pi_{t}](s_{t,1}^i)}_{\regret^{\common}(T)} + 2 T \hgkernel \sepisode^2 + 2 T\hgreward \sepisode\,.
    \]
    Next, we assume that the event $\cG(\delta)$, defined in \Cref{lem:proba_master_event} holds. Then \Cref{lem:optimism} implies
    \[
        \regret^{\common}(T) \leq \overline{\regret}^{\common}_1(T) + 3\nepisode \hgkernel \sepisode^2 +2  \nepisode \hgreward \sepisode  = \frac{1}{\nagent} \sum_{t=1}^T \sum_{i=1}^\nagent \evaluefunc[t,1](s_{t,1}^i) - \comvaluefunc[1][\pi_{t}](s_{t,1}^i) + 3\nepisode \hgkernel \sepisode^2 +2  \nepisode \hgreward \sepisode \eqsp.
    \]
By \Cref{lem:rht_bound} we have
\begin{align*}
\overline{\regret}^{\common}_1(\nepisode) \leq U^\nepisode_1 = A^\nepisode_1 + B^\nepisode_1 + C^\nepisode_1 +7\rme \nepisode\sepisode^2 \hgkernel + 2\rme \nepisode\sepisode \hgreward +
       \sqrt{8\sepisode^3 \nepisode \cdot \beta(\delta) / \nagent} + 4\rme \sepisode^2 \nstates \nactions\eqsp,
\end{align*}
and, applying \Cref{lem:bound_abc_terms}, we have the following quadratic inequality on $U^T_1$
\begin{align*}
    U^{\nepisode}_1 &\leq 69\rme \cdot \beta^{\max}(\delta) \cdot \sqrt{\sepisode^3 \nstates \nactions \nepisode \nagent^{-1}} + 69 \rme  \beta^{\max}(\delta) \sqrt{\sepisode^3 \nstates \nactions U^{\nepisode}_1 \nagent^{-1}} \\
    &+ 144 \rme \beta^{\max}(\delta) \nepisode (2\sepisode^2 \hgkernel + \hgreward \sepisode) + 288 \rme \sepisode^{3} \nstates^{3/2} \nactions \nagent^{-1/2} (\beta^{\max}(\delta))^2 \\
    &+ 272 \nstates^2 \nactions \nagent^{-1}\sepisode^3 (\beta^{\max}(\delta))^{2} + \rme \nepisode (7\sepisode^2 \hgkernel + 2\sepisode \hgreward ) + \sqrt{8 \sepisode^3 \nepisode \nagent^{-1} \beta^{\max}(\delta) } + 4 \rme \sepisode^2 \nstates \nactions\eqsp.
\end{align*}
After some rearranging of the terms, we have the following simplified version
\begin{align*}
    U^{\nepisode}_1 &\leq 69 \rme \cdot  \beta^{\max}(\delta) \sqrt{\sepisode^3 \nstates \nactions U^{\nepisode}_1 \nagent^{-1}} + 71\rme \cdot \beta^{\max}(\delta) \cdot \sqrt{\sepisode^3 \nstates \nactions \nepisode \nagent^{-1}}  \\
    &+ \rme \beta^{\max}(\delta) \nepisode (295\sepisode^2 \hgkernel + 146 \sepisode \hgreward) + 560 \sepisode^3 \nstates^2 \nactions (\beta^{\max}(\delta))^{2}\,.
\end{align*}
Finally, using inequality $2ab \leq a^2 + b^2$, we have
\[
    69 \rme \cdot  \beta^{\max}(\delta) \sqrt{\sepisode^3 \nstates \nactions U^{\nepisode}_1 \nagent^{-1}}  \leq \frac{1}{2} U^{\nepisode}_1 + 2450 \rme^2 \sepisode^3 \nstates \nactions \nagent^{-1}(\beta^{\max}(\delta))^2\eqsp,
\]
thus
\begin{align*}
    U^{\nepisode}_1 &\leq  138\rme  \sqrt{\sepisode^3 \nstates \nactions \nepisode \nagent^{-1} (\beta^{\max}(\delta) )^2} + 6020 \rme^2 \sepisode^3 \nstates^2 \nactions (\beta^{\max}(\delta))^{2} +\rme \beta^{\max}(\delta) \nepisode \sepisode (590 \sepisode \hgkernel + 144 \sepisode \hgreward)\,.
\end{align*}
\end{proof}

%% file: 2025-AISTATS/appendix/communication.tex
\section{COMMUNICATION COMPLEXITY}
\label{app:communicationanalysis}

In the sequel, we prove the bound on the communication complexity of \algo\, stated in \Cref{lem:communication}.

\comcomplexity*

\begin{proof}
Let us fix $(s,a,h) \in \nstates \times \A \times [\sepisode]$ and bound the maximum number of abortion signals triggered by this triplet. We define $R$ as the value of the variable $r$ that indicates the current round of the communication, defined in \algo, during iteration $\nepisode$. Let us also define $k_{s,a,h,i}$ as the number of times agent $i$ triggered the synchronization rule because of the triplet $(s,a,h)$ and $k_{s,a,h}$ the number of times the synchronization rule was triggered because of the triplet $(s,a,h)$. Recall that
\begin{align}
\label{def:sdeltat}
\nu(\delta, \nepisode) =  14 \hgkernel \nepisode \sepisode \nagent + 182 \nagent \beta^{\common}(\delta, \nepisode) \eqsp.
\end{align}
We distinguish two cases:

 (1) $\counter_{(R),h}(s,a) \leq \nu(\delta, \nepisode)$: Thus, it holds that $2^{k_{s,a,h,i}} \leq n_{(R),h}^i \leq \nu(\delta, \nepisode)$. Thereby $k_{s,a,h,i} \leq \log\left(\nu(\delta, \nepisode)\right)$. Hence, we have $k_{s,a,h} \leq \nagent \log\left(\nu(\delta, \nepisode)\right)$.

(2) $\counter_{(R),h}(s,a) > \nu(\delta, \nepisode)$: In this case, we can define
    \begin{align*}
    k_{s,a,h}^{\min} = \min\{r \in [R]: \counter_{(r),h}(s,a) \leq  \nu(\delta, \nepisode) \text{ and } \counter_{(r+1),h}(s,a) >  \nu(\delta, \nepisode) \} \eqsp.
    \end{align*}
    By the precedent case, we have $ k_{s,a,h}^{\min} \leq \nagent \log\left(\nu(\delta, \nepisode)\right)$. Now let us denote by $ r_1, \ldots, r_{p}$, where $p = k_{s,a,h} -    k_{s,a,h}^{\min}$, the indices of the rounds where the synchronization rule was triggered because of the triplet $(s,a,h)$ starting from round $k_{s,a,h}^{\min}$. Thus, for a certain $i\in [\nagent]$, we have
    \begin{align*}
    \hat{\counter}_{( r_{t+1}),h}^i(s,a) \geq 2\counter_{( r_{t+1}-1),h}(s,a) \eqsp.
    \end{align*}
    Under the event $\mathcal{E}^{\operatorname{count}}(\delta)$, we have for any $t\in [1;p]$,
    \begin{align*}
    \parcounter^{\,\nagent}_{(r_{t+1}),h}(s,a) \geq \frac{3}{7} \hat{N}^{\,i}_{(r_{t+1}),h}(s,a) + \frac{1}{7} \nu(\delta, \nepisode) \geq \frac{4}{7} \hat{N}^{\,i}_{(r_{t+1}),h}(s,a),   
    \end{align*}
    where $\parcounter^{\,\nagent}_{(r_{t+1}),h}(s,a)$ is defined in \eqref{eq:partial_counters_definition}. Combining the two previous inequalities, it gives
    \begin{align*}
    \parcounter^{\,\nagent}_{(r_{t+1}),h}(s,a)
    \geq 
    2 \cdot (4/7)\counter_{( r_{t+1}-1),h}(s,a) \geq (8/7)\counter_{( r_{t}),h}(s,a) = (8/7)\parcounter^{\,\nagent}_{(r_{t}),h}(s,a) \eqsp,
    \end{align*}
    where the second inequality comes from $r_{t+1} > \r_t$ and monotonicity of the counters.
    Unrolling the last recursion yields 
    \begin{align*}
    \nepisode \nagent \geq \parcounter^{\,\nagent}_{(r_{p}),h}(s,a) \geq (8/7)^{ k_{s,a,h} -   k_{s,a,h}^{\min}}\nu(\delta, \nepisode) \eqsp.
    \end{align*}
    Thus, we obtain
    \begin{align*}
     k_{s,a,h} \leq k_{s,a,h}^{\min} + \frac{\log(\nepisode \nagent/\nu(\delta, \nepisode))}{\log(8/7)}  \leq \nagent \log\left(\nu(\delta, \nepisode)\right) +  \frac{\log(\nepisode \nagent/\nu(\delta, \nepisode))}{\log(8/7)} \eqsp,
    \end{align*}
    which yields
    \begin{align}
    \label{def:roundsmax}
    \mathfrak{C}(T) \leq \troundsmax := \nagent \nstates \nactions  \sepisode\log\left(\nu(\delta, \nepisode)\right) +   \nstates \nactions  \sepisode \frac{\log(\nepisode \nagent/\nu(\delta, \nepisode))}{\log(8/7)} \eqsp.
    \end{align}
\end{proof}

%% file: 2025-AISTATS/appendix/technical.tex
\section{TECHNICAL LEMMAS}
\label{sec:technicallemmas}
\begin{lemma}[$\ell_1$-norm Bound]
\label{lem:tvbound}
Assume \Cref{assum:lhg1}, then
\begin{align}
\nonumber 
\max_{(s,a,h) \in \S \times \A \times [\sepisode]}\norm{\comkerMDP[h](\cdot | s,a) - \kerMDP[h][i](\cdot | s,a)}[1] \leq \hgkernel \eqsp.
\end{align}
\end{lemma}
\begin{proof}
Let $(s,a,h)\in \S \times \A \times [\sepisode] $. Using \Cref{assum:lhg1}, we get
\begin{align}
\nonumber \textstyle
\norm{\comkerMDP[h](\cdot|s,a) - \kerMDP[h][i](\cdot|s,a) }[1] = \sum_{s' \in \S} |\comkerMDP[h](s'|s,a) - \kerMDP[h][i](s'|s,a) | =   \sum_{s' \in \S} \hgkernel |\comkerMDP[h](s'|s,a) - \kerMDP[h][\indiv, i](s'|s,a) | \leq \hgkernel \eqsp.
\end{align}    
\end{proof}

\begin{lemma}
\label{lem:diff_stat_dist}
For any policy $\policy$, for any $(s,a,h) \in \S \times \A \times [\sepisode]$, and for any $(i, j) \in [\nagent]^2 $, we have
\begin{align*}
 |\statdist[s_0, h]{i, \policy} (s,a) -  \statdist[s_0,h]{j, \policy} (s,a) | \leq \hgkernel \sepisode \eqsp .
\end{align*}
\end{lemma}
\begin{proof}
Let us consider two following MDPs $\cM_1 =( \S,  \A,\sepisode, \{\kerMDP[h',i]\}_{1 \leq h' \leq \sepisode}, \{  \Ind_{(s,a)}(\cdot) \Ind_{h}(h')  \}_{1 \leq h' \leq\sepisode})$ and  $\cM_2 =( \S,  \A,\sepisode, \{\kerMDP[h',j]\}_{1 \leq h' \leq \sepisode},  \{  \Ind_{(s,a)}( \cdot) \Ind_{h}(h') \}_{1 \leq h' \leq\sepisode})$. Let's denote by $\tilde{V}_{h}^{i, \policy}$ and  $\tilde{V}_{h}^{j, \policy}$ the values function associated with the policy $\policy$ in these two respective environments. We have
\begin{align*}
\tilde{V}_{h}^{i, \policy}(s_0) = \PE_{\policy} \left[\sum_{h'=1}^\sepisode  \Ind_{(s,a)}(s_{h}^i , a_{h}^i) \Ind_{h}(h') \right] = \PE_{\policy} \left[\Ind_{(s,a)} (s_{h}^i , a_{h}^i) \right]  = \statdist[s_0, h]{i, \policy} (s,a)  \eqsp.
\end{align*}
Similarly, we have $ \tilde{V}_{h}^{j, \policy}(s_0) = \statdist[s_0, h]{j, \policy} (s,a)  $.  Finally applying \Cref{lem:performance-difference-russo} combined with Holder's inequality, and the fact that $\Vert \tilde{V}_{h}^{j, \policy}\Vert_\infty \leq 1$ yields
\begin{align*}
 |\statdist[s_0,h]{i, \policy} (s,a) -  \statdist[s_0,h]{j, \policy} (s,a) | \leq  \hgkernel \sepisode \eqsp.
\end{align*}
\end{proof}
\subsection{Bellman type equations for the variance}

For a deterministic policy $\pi$ and an agent $i$, we recall the following definitions of the Bellman-type equations for the variances as follows
\begin{align}
\nonumber
\sigma \qfunc[h][i,\policy](s, a) & := \agentvar[h]{i}{ \valuefunc[h+1][i,\policy]}(s, a)+ \kerMDP[h][i] \sigma \valuefunc[h+1][i,\policy](s, a)
\\ \nonumber
\sigma \valuefunc[h][i,\policy](s) & := \sigma \qfunc[h][i,\policy](s, \pi(s)) 
\\ \label{eq:definition_variance_bellman}
\sigma\valuefunc[\sepisode+1][i,\policy](s) & := 0 \eqsp,
\end{align}
where $\agentvar[h]{i}{f}(s, a) := \mathbb{E}_{s^{\prime} \sim\kerMDP[h][i](\cdot \mid s, a)}\left[\left(f\left(s^{\prime}\right)-\kerMDP[h][i]f(s, a)\right)^2\right]$ denotes the variance operator. Unrolling the precedent relation yields
\begin{align*}
\sigma \valuefunc[1][i,\policy](s)=\sum_{h=1}^\sepisode \sum_{s', a'} \statdist[s,h]{i, \policy}(s', a') \agentvar[h]{i}{ \valuefunc[h+1][i,\policy]}(s', a') \eqsp,
\end{align*}
where $\statdist[s,h]{i, \policy}(s', a')$ is the probability of visiting a pair $(s',a')$ in the $i$-th environment while following the policy $\policy$ and starting from a state $s$. Next, we state the well-known Bellman equation for variances (see, e.g., \citealt{sobel1982variance,azar2017minimax}).
\begin{lemma}
\label{lem:bellman_variance}
For any deterministic policy $\policy$, for all $h \in[\sepisode]$, and for all $i \in [\nagent]$,
\begin{align}
\label{eq:bellman_var}
\mathbb{E}_\pi\left[\left(\sum_{h'=h}^\sepisode \rewardMDP[h'][i] (s^i_{h'}, a^i_{h'})- \qfunc[h][i,\policy] (s^i_{h}, a^i_{h})\right)^2 \middle\vert (s^i_{h}, a^i_{h})=(s, a)\right]=\qfunc[h][i,\policy](s, a) \eqsp.
\end{align}
In particular,
\begin{align*}
\mathbb{E}_\pi\left[\left(\sum_{h=1}^\sepisode \rewardMDP[h][i] (s^i_{h}, a^i_{h})-  \valuefunc[1][i,\policy](s_1^i)\right)^2\right]=\sigma \valuefunc[1][i,\policy](s_1^i)=\sum_{h=1}^\sepisode \sum_{s, a} \statdist[h]{i, \policy}(s, a) \agentvar[h]{i}{ \valuefunc[h+1][i,\policy]}(s, a) \eqsp.
\end{align*}
\end{lemma}

\subsection{Concentration inequalities}

\begin{lemma}[Deviation inequality for categorical distribution, \citealt{jonsson2020planning}]
\label{lem:kl_deviation_inequality}
Let $\left(X_t\right)_{t \in \mathbb{N}^*}$ be i.i.d. samples from a probability measure $\kerMDP$ supported on $\{1, \ldots, m\}$. We denote by $\widehat{\kerMDP}_n$ the empirical vector of probabilities, i.e., for all $k \in\{1, \ldots, m\}$,
$$
\widehat{\kerMDP}_n(k) := \frac{1}{n} \sum_{\ell=1}^n \mathbf{1}_{\{k\}}(X_{\ell}) .
$$
For all $\kerMDP$ and for all $\delta \in(0,1)$,
$$
\mathbb{P}\left(\exists n \in \mathbb{N}^{\star}, n \mathrm{KL}\big(\widehat{\kerMDP}_n \big\Vert \kerMDP\big)>\log (1 / \delta)+(m-1) \log (e(1+n /(m-1)))\right) \leq \delta \eqsp.
$$

\end{lemma}

\begin{lemma} [Corollary 11 by \citealt{talebi2018variance}]
\label{lem:talebi}
Let $\kerMDP, \mathsf{Q}$ two probability distributions on $\S$. For all functions $f: \mathcal{S} \mapsto[0, \sepisode]$,
$$
\kerMDP f-\mathsf{Q} f \leq \sqrt{2 \Var_{\mathsf{Q}}(f) \mathrm{KL}(\kerMDP \Vert \mathsf{Q})}+\frac{2}{3} \sepisode \mathrm{KL}(\kerMDP, \mathsf{Q}) \eqsp . 
$$
where we have defined $\kerMDP f:= \PE_{s \sim \kerMDP}[f(s)]$.
\end{lemma}

\begin{lemma}[Lemma H.9 by \citealt{tiapkin2023fast}]
\label{lem:transportation_measure}
For  any two probability measures $\mathsf{P},\mathsf{Q}$ on $\S$, for $f, g: \S \mapsto[0, b]$ two functions defined on $\S$, we have that
\begin{align}
& \label{eq:function_transfert}
\Var_{\mathsf{P}}(f)\leq 2 \Var_{\mathsf{P}}(g)+2 b \mathsf{P}|f-g| \quad \text { and } \\
& \label{eq:distribution_transfert}
\Var_{\mathsf{Q}}(f) \leq \Var_{\mathsf{P}}(f)+3 b^2 \Vert \mathsf{P}-\mathsf{Q} \Vert_1
\end{align}
where we denote the absolute operator by $|f|(s)=|f(s)|$ for all $s \in \S$.
\end{lemma}

\begin{lemma}[Theorem 4 by \citealt{maurer2009empirical}]
\label{lem:empiricalbernstein}
. Consider any $\delta>0$ and any integer $n \geq 2$. Let $Y, Y_1, \ldots, Y_n$ be a collection of i.i.d. random variables falling within $[0,1]$. Define the empirical mean $\bar{Y}:=$ $\frac{1}{n} \sum_{i=1}^n Y_i$ and empirical variance $\widehat{Y}_n:=\frac{1}{n} \sum_{i=1}^n\left(Y_i-\bar{Y}\right)^2$. Then we have
$$
\mathbb{P}\left[\left|\mathbb{E}[Y]-\frac{1}{n} \sum_{i=1}^n Y_i\right|>\sqrt{\frac{2 \widehat{Y}_n \log (2 / \delta)}{n-1}}+\frac{7 \log (2 / \delta)}{3(n-1)}\right] \leq \delta
$$
\end{lemma}

Below, we state the self-normalized Bernstein-type inequality by \cite{domingues2021kernel}. Let $(Y_t)_{t\in\nset^\star}$, $(w_t)_{t\in\nset^\star}$ be two sequences of random variables adapted to a filtration $(\cF_t)_{t\in\nset}$. We assume that the weights are in the unit interval $w_t\in[0,1]$ and predictable, i.e. $\cF_{t-1}$ measurable. We also assume that the random variables $Y_t$  are bounded $|Y_t|\leq b$ and centered $\PE[Y_t | \cF_{t-1}] = 0$.
Consider the following quantities
\begin{align*}
		S_t := \sum_{s=1}^t w_s Y_s\,, \qquad V_t := \sum_{s=1}^t w_s^2\cdot\PE[Y_s^2|\cF_{s-1}]\,,
\end{align*}
and let $h(x) \triangleq (x+1) \log(x+1)-x$ be the Cramér transform of a Poisson distribution of parameter~1.

\begin{theorem}[Anytime Bernstein-type concentration inequality for martingales]
  \label{th:bernstein}
	For all $\delta >0$,
	\begin{align*}
		\P\left[\exists t\geq 1,   (V_t/b^2+1)h\left(\!\frac{b |S_t|}{V_t+b^2}\right) \geq \log(1/\delta) + \log\left(4e(2t+1)\!\right)\right]\leq \delta.
	\end{align*}
  The previous inequality can be weakened to obtain a more explicit bound: if $b\geq 1$ with probability at least $1-\delta$, for all $t\geq 1$,
 \[
 |S_t|\leq \sqrt{2V_t \log\left(4\rme(2t+1)/\delta\right)}+ 3b\log\left(4\rme(2t+1)/\delta\right)\,.
 \]
\end{theorem}

Next, we apply this Bernstein inequality to a particular distribution.
Let $\mathcal F_t$ for $t\in\nset$ be a filtration and $(X_t)_{t\in\nset^\star}$ be a sequence of Bernoulli random variables with $\P(X_t = 1 | \mathcal F_{t-1}) = P_t$ with $P_t$ being $\mathcal F_{t-1}$-measurable and $X_t$ being $\mathcal F_{t}$-measurable.

\begin{corollary}\label{cor:bernoulli-deviation}
    For all $\delta>0$,
	\begin{align*}
	\P \left(\exists n : \,\,  \left|\sum_{t=1}^n X_t - P_t\right| > \frac{1}{8} \sum_{t=1}^n P_t +       11\log\left(\frac{4\rme(2n+1)}{\delta}\right)  \right) \leq \delta.
	\end{align*}
\end{corollary}
\begin{proof}
    Given a simplified version, we have with probability at least $1-\delta$ by applying inequality $2ab \leq a^2 + b^2$ for $a,b\geq 0$
    \[
        \left|\sum_{t=1}^n X_t - P_t\right|\leq \sqrt{2 \cdot \frac{V_n}{8} \cdot  8\log\left(4\rme(2n+1)/\delta\right)}+ 3\log\left(4\rme(2n+1)/\delta\right) \leq \frac{1}{8}\sum_{t=1}^n P_t + 11\log\left(4\rme(2n+1)/\delta\right)\,.
    \]
\end{proof}

\subsection{Performance-difference Lemma}

\begin{lemma}[Lemma 3 of \citealt{russo2019worst}]\label{lem:performance-difference-russo}
    Let us consider two MDPs $\cM_1 = (\S, \A, \sepisode, \r^{(1)}, \kerMDP[][(1)])$ and $\cM_2 = (\S, \A, \sepisode, \r^{(2)}, \kerMDP[][(2)]$. Let $\valuefunc[1][(1),\policy](s)$ and $\valuefunc[1][(2),\policy](s)$ are values of a fixed policy $\pi$ in MDP $\cM_1$ and $\cM_2$ respectively. Then it holds
    \[
        \valuefunc[1][(1),\policy](s) - \valuefunc[1][(2),\policy](s) = \PE_{\pi, \cM_1}\left[ \sum_{h=1}^{\sepisode} \left(\r^{(1)}_h - \r^{(2)}_h\right)(s_h,a_h) + \left(\kerMDP[h][(1)] - \kerMDP[h][(2)]\right)\valuefunc[h+1][(2),\policy](s_h,a_h)   \right],
    \]
    where expectation is taken over the trajectories $(s_1,a_1,\ldots,s_{\sepisode},a_{\sepisode})$ generated by policy $\pi$ in an MDP $\cM_1$.
\end{lemma}

\begin{corollary}\label{cor:performance-difference-tv}
    Let us consider two MDPs $\cM_1 = (\S, \A, \sepisode, \r^{(1)}, \kerMDP[][(1)])$ and $\cM_2 = (\S, \A, \sepisode, \r^{(2)}, \kerMDP[][(2)]$, such that $\forall (s,a,h) \in \S \times \A \times [\sepisode]: |\r^{(1)}_h(s,a) - \r^{(2)}(s,a)| \leq \varepsilon_{\r}$, $|\valuefunc[h][(1),\policy](s)| \leq c$, $|\valuefunc[h][(2),\policy](s) |\leq c$, and $\norm{\kerMDP[h][(1)](s,a) - \kerMDP[h][(2)](s,a)}[1] \leq \varepsilon_{\kerMDP}$ where $c > 0$ is a positive constant and $\valuefunc[1][(1),\policy](s)$ and $\valuefunc[1][(2),\policy](s)$ are values of a fixed policy $\pi$ in MDP $\cM_1$ and $\cM_2$ respectively. Then it holds
    \[
        \valuefunc[1][(1),\policy](s_1) - \valuefunc[1][(2),\policy](s_1) \leq \varepsilon_{\kerMDP} c \sepisode + \varepsilon_{\r} \sepisode\,.
    \]
\end{corollary}
\begin{proof}
    Follows directly from combination of \Cref{lem:performance-difference-russo}, Holder's inequality and a fact that $\norm{\valuefunc[h][(2),\policy]}[1] \leq c$.
\end{proof}

Inspired by a construction of \cite{ross2010efficient}, we can show that dependence $H^2$ in terms of $\ell_1$-distance between two models in non-improvable.

\begin{lemma}\label{lem:performance-difference-lb}
    There exist two MDPs $\cM_1 = (\S, \A, \sepisode, \r, \kerMDP[][1])$ and $\cM_2 = (\S, \A, \sepisode, \r, \kerMDP[][2])$ with the same reward function and different kernels, $H \geq 2$ such that $\forall (s,a,h) \in \S \times \A \times [\sepisode]: \norm{\kerMDP[h][1](s,a) - \kerMDP[h][2](s,a)}[1] \leq \varepsilon_{\kerMDP}$ for $ 0 < \hgkernel < 2/ \sepisode$.
    Then there is a policy $\pi$ such that values $\valuefunc[1][1,\policy](s)$ and $\valuefunc[1][2,\policy](s)$ in MDPs $\cM_1$ and $\cM_2$ satisfy
    \[
        \valuefunc[1][1,\policy](s_1) - \valuefunc[1][2,\policy](s_1) = \Omega(\hgkernel H^2)\,.
    \]
\end{lemma}
\begin{proof}
    Consider the problem with 2 states $\{s_1,s_2\}$ and 1 action $\{a\}$, the agent always starts at $s_1$. The reward function satisfies $\r_h(s_1,a) = 1, \r_h(s_2,a) = 0$ for all $h \in \sepisode$. Finally, the transition kernels are the same for all $h$ and are defined as
    \[
        \kerMDP[h][i](s_1 | s_1, a) = 1 - p_i, \quad  \kerMDP[h][i](s_2 | s_1, a) = p_i, \quad \kerMDP[h][i](s_1 | s_2, a) = 0, \quad \kerMDP[h][i](s_2 | s_2, a) = 1\,,
    \]
    for $i \in \{ 1,2\}$. In other words, the state $s_2$ is a sink with zero reward. Since there is only one action, the value is the same for any policy $\pi$. Let us take $p_1 = 0$ and $p_2 = \hgkernel$, then under the kernel $\kerMDP[][1]$ the value $\valuefunc[1][1, \pi](s_1)$ is equal to $H$, whereas under the kernel $\kerMDP[][2]$, the value function $\valuefunc[1][1, \pi](s_1)$ it is equal to
    \[
        \valuefunc[1][2, \pi](s_1) = 1 + (1-\hgkernel) + (1-\hgkernel)^2 + \ldots + (1-\hgkernel)^{H-1} = \frac{1-(1-\hgkernel)^{H} }{\hgkernel}.
    \]
    Then we have
    \[
        \valuefunc[1][1,\policy](s_1) - \valuefunc[1][2,\policy](s_1) = \frac{H\hgkernel - 1 + (1-\hgkernel)^{H}}{\hgkernel} \eqsp.
    \]
Now as $0<\hgkernel <2/\sepisode$, Bernoulli's inequality yields
\begin{align*}
\valuefunc[1][1,\policy](s_1) - \valuefunc[1][2,\policy](s_1)  =  \frac{H\hgkernel - 1 + (1-\hgkernel)^{\sepisode/2}(1-\hgkernel)^{\sepisode/2}}{\hgkernel} \geq \frac{H\hgkernel - 1 + (1-\sepisode\hgkernel/2)(1-\sepisode\hgkernel/2)}{\hgkernel} =  \frac{\hgkernel \sepisode^2}{4}  \eqsp,
\end{align*}
where the first inequality comes from $(1 - x)^{r} \ge 1 - rx $ for $0 \le x \le 1$ and $r > 1$.
\end{proof}